\documentclass{article}

\usepackage[preprint]{neurips_2022}

\usepackage[utf8]{inputenc} 
\usepackage[T1]{fontenc}    
\usepackage{hyperref}       
\usepackage{url}            
\usepackage{booktabs}       
\usepackage{amsfonts}       
\usepackage{nicefrac}       
\usepackage{microtype}      
\usepackage{xcolor}         

\usepackage{graphicx}
\usepackage{booktabs} 

\usepackage{makecell}

\usepackage{amsmath}
\usepackage{amssymb}
\usepackage{mathtools}
\usepackage{amsthm}
\usepackage{breqn}
\usepackage{pgfplots}

\usepackage[capitalize,noabbrev]{cleveref}

\theoremstyle{plain}

\theoremstyle{definition}

\theoremstyle{remark}

\usepackage[textsize=tiny]{todonotes}

\usepackage[utf8]{inputenc}
\usepackage{microtype}
\usepackage{graphicx}
\usepackage{booktabs} 
\usepackage{hyperref}

\usepackage{pifont}
\usepackage[ruled,vlined]{algorithm2e}
\usepackage{float}
\newfloat{eqn}{htbp}{loe}
\floatname{eqn}{Equation}
\usepackage{tabularx}
\usepackage{amssymb}
\usepackage{breqn}
\usepackage{amsthm}
\usepackage[nopar]{lipsum}
\usepackage{changepage}
\usepackage{multirow}
\usepackage{algorithmic,multicol}

\usepackage{caption}
\usepackage{subcaption}

\usepackage{tikz}
\definecolor{smallcnn}{RGB}{85,163,205}
\definecolor{resnet9}{RGB}{73,84,176}
\definecolor{resnet18}{RGB}{59,33,39}
\definecolor{resnet34}{RGB}{156,47,69}
\definecolor{resnet50}{RGB}{233,111,54}
\definecolor{wide_resnet50_2}{RGB}{27,97,69}
\definecolor{resnet101}{RGB}{105,123,48}
\definecolor{wide_resnet101_2}{RGB}{200,123,124}
\definecolor{resnet152}{RGB}{205,162,224}
\definecolor{vgg}{RGB}{198,225,241}

\definecolor{good_ASR}{RGB}{233,141,107}
\definecolor{bad_ASR}{RGB}{108,43,109}
\usepackage{multirow}

\usepackage{tikz}

\usepackage{cases}
\usepackage{enumitem}

\usepackage{amsmath}

\usepackage{wrapfig}
\usepackage{multirow}
\usepackage{needspace}
\usepackage{blindtext}

\usepackage[scaled=.7]{beramono}
\usepackage{comment}

\usepackage{collectbox}
\makeatletter

\usepackage{centernot}

\usepackage{nicefrac}       
\usepackage{thm-restate}
\usepackage{amsmath,amsfonts,bm}
\usepackage{amssymb,amsmath}
\usepackage{amsthm}
\usepackage[scr=boondoxo]{mathalfa}

\usepackage{xcolor}
\newcommand*{\colorboxed}{}
\def\colorboxed#1#{%
  \colorboxedAux{#1}%
}
\newcommand*{\colorboxedAux}[3]{%
  \begingroup
    \colorlet{cb@saved}{.}%
    \color#1{#2}%
    \boxed{%
      \color{cb@saved}%
      #3%
    }%
  \endgroup
}

\title{Interpolating Compressed Parameter Subspaces}

\author{%
  Siddhartha ~Datta \\
  Department of Computer Science\\
  University of Oxford\\
  \texttt{siddhartha.datta@cs.ox.ac.uk} \\
  \And
  Nigel ~Shadbolt \\
  Department of Computer Science\\
  University of Oxford\\
  \texttt{nigel.shadbolt@cs.ox.ac.uk} \\
}

\begin{document}

\maketitle

\begin{abstract}
Inspired by recent work on neural subspaces and mode connectivity, 
we revisit parameter subspace sampling for shifted and/or interpolatable input distributions (instead of a single, unshifted distribution). 
We enforce a compressed geometric structure upon a set of trained parameters mapped to a set of train-time distributions, denoting the resulting subspaces as \textit{Compressed Parameter Subspaces (CPS)}. 
We show the success and failure modes of the types of shifted distributions whose optimal parameters reside in the CPS. 
We find that ensembling point-estimates within a CPS can yield a high average accuracy across a range of test-time distributions, including backdoor, adversarial, permutation, stylization and rotation perturbations. 
We also find that the CPS can contain low-loss point-estimates for various task shifts (albeit interpolated, perturbed, unseen or non-identical coarse labels).
We further demonstrate this property in a continual learning setting with CIFAR100.
\end{abstract}

\section{Introduction}

The implicit construction of a parameter subspace is a commonality across methods in 
meta/continual learning, domain adaptation/generalization, differential privacy, personalized federated learning, machine unlearning, etc.
Improved interpolatability of the parameter subspace and consistently mapping low-loss parameters back to inputs
translates into increased method performance.
This has motivated task augmentation techniques for meta learning, such as MLTI \citep{yao2021metalearning} which interpolates between different tasks in label-sharing and non-label-sharing settings.
This also motivates tools for input-parameter space diagnostics, such as datamodels \citep{https://doi.org/10.48550/arxiv.2202.00622} which maps parameters trained on different subsets of CIFAR10 in order to compute per-instance statistics.

Motivated by recent work studying the geometry of the loss landscape,
such as neural subspaces \citep{wortsman2021learning} and mode connectivity \citep{fort2019large, draxler2019essentially, garipov2018loss}, 
extending on the current approach of constructing a geometric structure (specifically a parameter subspace) optimized to a single, relatively-unperturbed/unshifted source distribution,
we investigate the construction of subspaces that can return low-loss parameter point-estimates mapped back to a wide range of shifted distributions. 

\noindent\textbf{Contributions.}
This work proposes a method to construct a compressed parameter space such that the likelihood of a subspace-sampled parameter that can be mapped back to an input in the input space, and vice versa, is higher. 
We demonstrate the utility of CPS for single and multiple test-time distribution settings, 
with improved mappings between the two spaces with higher accuracy,
improved robustness performance across perturbation types, 
reduced catastrophic forgetting on Split-CIFAR10/100,
strong capacity for multi-task solutions and unseen/distant tasks,
and storage-efficient inference (ensembling, hypernetworks).

\begin{figure*}[t]
    \centering
    \includegraphics[width=0.28\textwidth]{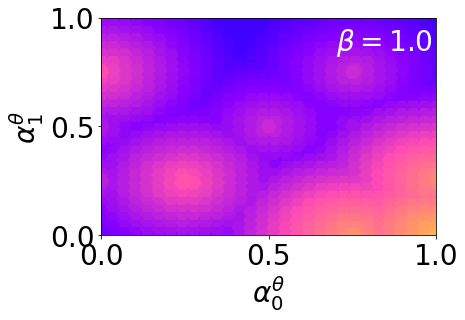}
    \hspace{0.25cm}
    \includegraphics[width=0.28\textwidth]{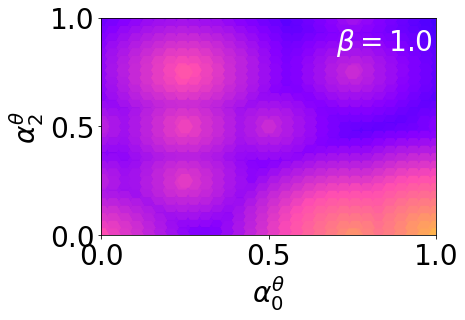}
    \hspace{0.25cm}
    \includegraphics[width=0.28\textwidth]{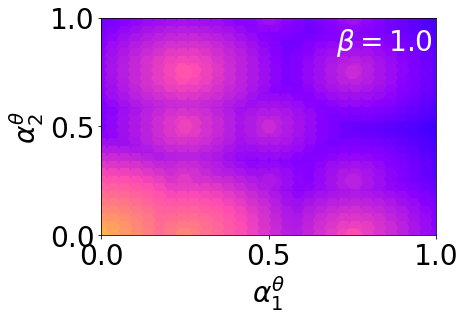}
    \includegraphics[width=0.5\textwidth]{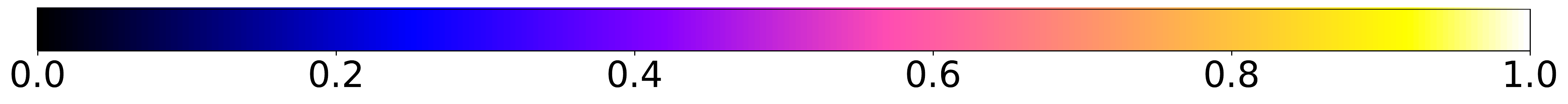}
    \caption{
    \textit{Change in parameter subspace dynamics: }
    Landscape of 
    unique lowest-loss parameters mapped to interpolated inputs. 
    Refer to 
    Appendix A.3.3
    for supplementary visualization.
    }
    \label{fig:loss_landscape}
    \vspace{-0.25cm}
\end{figure*}

\section{Compressed Parameter Spaces}

\subsection{Problem}

Let $\mathcal{X}$, $\mathcal{Y}$, $\mathcal{K}$, $\Theta$ be denoted as the input space, coarse label space, fine label space, and parameter space respectively, such that $\mathcal{X} \mapsto \mathcal{Y}$, $\mathcal{X} \mapsto \mathcal{K}$, $\mathcal{K} \mapsto \mathcal{Y}$, $\mathcal{X} \mapsto \Theta$.
Coarse labels are the higher-order labels of fine labels.
$\mathscr{f}$ is a base learner function
that accepts inputs $\mathbf{x}$, $\theta$ to return predicted labels $\bar{\mathbf{y}} = \mathscr{f}(\theta; \mathbf{x})$. 
The model parameters $\theta \sim \Theta$ are sampled from the parameter space
such that it minimizes the loss between the ground-truth and predicted labels: 
$\mathcal{L}(\theta; \mathbf{x}, \mathbf{y}) = \frac{1}{|\mathbf{x}|} \sum_{i}^{|\mathbf{x}|} (\mathscr{f}(\theta; \mathbf{x})-\mathbf{y})^2$.

\noindent\textbf{Definition 1. }
\textit{
A \textbf{distribution shift} is the divergence between a train-time distribution $\mathbf{x_0}, \mathbf{y_0}$ and a test-time distribution $\hat{\mathbf{x}}, \hat{\mathbf{y}}$, where
$\hat{\mathbf{x}}, \hat{\mathbf{y}}$ is an interpolated distribution between $\mathbf{x_0}, \mathbf{y_0}$ and a target distribution $\mathbf{x_i^{\Delta}}, \mathbf{y_i^{\Delta}}$ such that
$\hat{\mathbf{x}} = \sum_{i}^{N} \alpha_i \mathbf{x_i^{\Delta}}$
and $\hat{\mathbf{y}} = \mathbf{y_i} | i := \arg\max_{i} \alpha_i$.
A \textbf{disjoint distribution shift} is a distribution shift of one target distribution $|\{ \alpha_i \}| = 2$ s.t. $\hat{\mathbf{x}} = \alpha \mathbf{x^{\Delta}} + (1-\alpha) \mathbf{x_0}$.
A \textbf{joint distribution shift} is a distribution shift of multiple target distributions $|\{ \alpha_i \}| > 2$ s.t. $\hat{\mathbf{x}} = \sum_{i}^{N} \alpha_i \mathbf{x_i^{\Delta}}$.
}

For a set of distributions $\{ \mathbf{x_0} \mapsto \mathbf{y_0},  \mathbf{x_1^{\Delta}} \mapsto \mathbf{y_1^{\Delta}}, ..., \mathbf{x_N^{\Delta}} \mapsto \mathbf{y_N^{\Delta}}\}$ containing $(N-1)$ target distributions, we sample interpolation coefficients $\alpha_i \sim [0, 1]$  s.t. $\alpha_i \mapsto \mathbf{x_i}$ and $\sum_{i}^{N} \alpha_i \leq N$. 
$N$ is the number of distributions being used; in CPS training, it is the number of train-time distributions used in training the subspace; in task interpolation, it is the number of train/test-time distributions used in interpolating between task sets.
Further details on the scoping of distribution shifts are discussed in Appendix A.1. 
Given a set of valid interpolated input sets (manifesting disjoint/joint distribution shifts) and their corresponding interpolation coefficients $\{ \{ \alpha_i \} \mapsto \hat{\mathbf{x}} \}$, the primary objective of this work is to minimize the cumulative loss $\mathcal{L}(\Theta; \mathcal{X}) = \sum_{\{\alpha_i\}}^{\{\{\alpha_i\}\}} \mathcal{L}(\theta \sim \Theta; \sum_i^{|\{\alpha_i\}|} \alpha_i \mathbf{x_i})$.

\noindent\textbf{(Methodology) Test-time distributions. }
We evaluate on CIFAR10 and CIFAR100 \citep{krizhevsky2009learning} for single (non-task-shift) and multiple test-time distributions (task-shift) respectively.
For label shift w.r.t. train-time perturbations,
we implement backdoor attacks with Random-BadNet \citep{datta2022backdoors}, a variation of 
BadNet \citep{8685687} to allow for multiple trigger patterns.
For label shift w.r.t. test-time perturbations, 
we implement adversarial attacks for targeted label shift with Projected Gradient Descent \citep{madry2018towards}, 
and implement random permutations for untargeted label shift with Random-BadNet (in-line with PermuteCIFAR10).
We implement stylization with Adaptive Instance Normalization \citep{huang2017adain}.
We rotate images in-line with RotateCIFAR10.
For task shift, where the tasks are pre-defined in the dataset with shared or non-shared label sets, 
we evaluate on both the end-point tasks as well as interpolated tasks.
Defined tasks and implementation details are discussed in Appendix A.2.

\noindent\textbf{(Methodology) Task interpolation. }
Task interpolation is an existing task augmentation technique. For example, MLTI \citep{yao2021metalearning} linearly interpolates the hidden state representations at the $\ell$th layer of a model when a set of tasks are passed through it. 
We evaluate interpolated tasks by linearly interpolating a set of tasks: 
for a set of tasks $\{X\}^N$ where each task $X$ is a set of inputs $x$ (and the number of inputs per task is identical, i.e. $\{\{x\}^M\}^N$),
we linearly-interpolate between indexed inputs across a task set to return an interpolated task: $\hat{\mathbf{x}} = \sum_i^N \sum_j^M \alpha_i x_{i,j}$.
In this work, we only interpolate between task sets with identical labels (label-shared). 
We provide further details on the interpolation coefficient sampling and sample interpolated images in Appendix A.2.

\subsection{Implementation (Algorithm \ref{alg:train})}

\noindent\textbf{Definition 2. }
\textit{
The \textbf{parameter space} $\Theta \in \mathbb{R}^M$ is a topological space, where $M$-dimensional parameter point-estimates $\theta \sim \Theta$ are sampled and loaded into a base learner function $\mathscr{f}(\theta; \cdot)$.
A \textbf{parameter subspace} $\vartheta$ is a bounded subspace contained in $\Theta$ such that: 
{\small
$$\vartheta = \Big\{ \sum_i^N \alpha_i \theta_i \in \Theta \Big| 0 \leq \alpha_i \leq 1, 0 \leq \sum_i^N \alpha_i \leq N, \{ \min \mathcal{L}(\theta_i; X_i, Y_i) \}^{i \in N} \Big\}$$
}
A \textbf{compressed parameter subspace} is a parameter subspace with the added constraint that the distance between the set of end-point parameters 
should be minimized (Algorithm \ref{alg:train}) such that:
{\small
$$
\vartheta = \Big\{ \sum_i^N \alpha_i \theta_i \in \Theta \Big| 0 \leq \alpha_i \leq 1, 0 \leq \sum_i^N \alpha_i \leq N, \{ \min \mathcal{L}(\theta_i; X_i, Y_i) \}^{i \in N}, 
\colorboxed[rgb]{0.827, 0.827, 0.827}{\min \textnormal{\texttt{dist}}(\{\theta_i\}^N)}
\Big\}
$$
}
}

\noindent\textbf{Theorem 1. }
\textit{
For $N+1$ parameters trained in parallel,
any individual parameter $\theta_i$ is a function of its mapped data $X_i \mapsto Y_i$ as well as the remaining parameters $\{ \theta_n \}^{n \in N}$ (and by extension, $X_n \mapsto Y_n$).
Hence, a linearly-interpolated parameter between $\theta_i$ and $\{ \theta_n \}^{n \in N}$ is a function of $\{ \theta_n \}^{n \in N}$ and $\theta_i | \{ \theta_n \}^{n \in N}$.
}

Denoting $\mathcal{C} = \sum_n^N \theta_n$, computing $\mathcal{L}(\theta_i) = \mathcal{L}(\theta_i, X_i, Y_i) + \texttt{dist}(\theta_i, \{ \theta_n \}^{n \in N})$ such that $\mathcal{L}(\theta_i) = 0$ leads to the property that: 
{\small
$$\theta_i := \frac{2\mathcal{C}-X_i \pm \sqrt{-4 X_i \mathcal{C} + 4N Y_i + X_i^2}}{2N}$$
}
This and subsequent results find that any end-point and interpolated parameter $\theta_i$ will be a function of a nested $\{ \theta_n \}^{n \in N}$.
Refer to proof in Appendix A.1.

\noindent\textbf{Theorem 2. }
\textit{
For $N+1$ parameters trained in parallel,
for an individual parameter $\theta_i$ and another parameter $\theta_n$ where $n \in N$,
the change in loss per iteration of SGD with respect to each parameter $\frac{\partial \mathcal{L}(\theta_i; X_i)}{\partial \theta_i}$, $\frac{\partial \mathcal{L}(\theta_n; X_n)}{\partial \theta_n}$ will both be in the same direction (i.e. identical signs), and will tend towards a local region of the parameter space upon convergence of loss.
}

By computing the incremental loss attributed to the distance regularization term upon convergence $\Delta \mathcal{L}(\theta_i)_e = 0$, 
we find that: 
{\small
$$\frac{\partial \mathcal{L}(\theta_i; X_i)}{\partial \theta_i} \cdot \frac{\partial \mathcal{L}(\theta_n; X_n)}{\partial \theta_n} \equiv \frac{1}{2}\bigg[\bigg(\frac{\partial \mathcal{L}(\theta_n; X_n)}{\partial \theta_n}\bigg)^2 + \bigg(\frac{\partial \mathcal{L}(\theta_i; X_i)}{\partial \theta_i}\bigg)^2\bigg]
$$
}
Subsequently,
we find 
the respective cumulative loss of (and consequently distance between) $\theta_i$ and $\theta_n$ from iterations $e=0$ to $e=E-1$: 
{\small
$$\bigg|\sum_{e}^{E-1} \frac{\partial \mathcal{L}(\theta_{i, t=e}; X_i)}{\partial \theta_{i, t=e}}\bigg|-\bigg|\sum_{e}^{E-1} \frac{\partial \mathcal{L}(\theta_{n, t=e}; X_n)}{\partial \theta_{n, t=e}}\bigg|$$
}
to be lower when $\frac{\partial \mathcal{L}(\theta_i; X_i)}{\partial \theta_i} \cdot \frac{\partial \mathcal{L}(\theta_n; X_n)}{\partial \theta_n} > 0$ (i.e. active distance regularization) than when $\frac{\partial \mathcal{L}(\theta_i; X_i)}{\partial \theta_i} \cdot \frac{\partial \mathcal{L}(\theta_n; X_n)}{\partial \theta_n} \leq 0$.
Refer to proof in Appendix A.1.

\noindent\textbf{Theorem 3. }
\textit{
For $N+1$ parameters trained in parallel,
the expected distance between the optimal parameter of
an interpolated input $\hat{\mathbf{x}} \mapsto \hat{\theta}$ 
would be smaller with respect to 
sampled points in a 
distance-regularized subspace
than 
non-distance-regularized subspace.
}

For the the ground-truth parameter of a sampled interpolated input distribution, 
in order for the loss to converge $\mathcal{L}(\hat{\theta}) \rightarrow 0$, 
the product of the loss terms converges in the same direction $\frac{\partial \mathcal{L}(\hat{\theta}; X_i)}{\partial \hat{\theta}} \cdot \frac{\partial \mathcal{L}(\hat{\theta}; X_n)}{\partial \hat{\theta}} > 0$.
For iterations $e=0$ to $e=E-1$,
the cumulative loss (and consequently distance) between the ground-truth parameter and parameter end-points 
{\small
$$
\mathbb{E}
\Bigg[
\Bigg|
2 \bigg[
\alpha_i \frac{\partial \mathcal{L}(\hat{\theta}_{t=e}; \mathbf{x_i})}{\partial \hat{\theta}_{t=e}}
+ \alpha_n \frac{\partial \mathcal{L}(\hat{\theta}_{t=e}; \mathbf{x_n})}{\partial \hat{\theta}_{t=e}}
\bigg]
+
\bigg[
\gamma_n \frac{\partial \mathcal{L}(\theta_{n, t=e}; \mathbf{x_n})}{\partial \theta_{n, t=e}}
- 
\gamma_i \frac{\partial \mathcal{L}(\theta_{i, t=e}; \mathbf{x_i})}{\partial \theta_{i, t=e}}
\bigg]
\Bigg|
\Bigg]
$$
}
is closer in the distance-regularized space than non-distance-regularized space. Refer to proof in Appendix A.1.

From Theorem 2, we show that the end-point parameters converge towards a local region in the parameter space; as the average distance between any end-point and the centre of these points is lower upon loss convergence, this space is referred to as being \textit{compressed}.
From Theorem 1, we show that the end-point parameters (and subsequently interpolated parameters within the subspace) contain real-valued artifacts of opposing end-point parameters. 
Further, from Theorem 3, we show that an interpolated input set $\hat{\mathbf{x}}$ has a lower expected distance towards any interpolated (or end-point) parameter point-estimate in a compressed space than an uncompressed space.
From these properties, 
we empirically validate that interpolated input sets in single and multiple test-time distributions can be mapped back to interpolated parameters in the CPS.

\cite{garipov2018loss}, \cite{fort2020deep}, \cite{mirzadeh2020linear} and \cite{wortsman2021learning}
construct paths and/or subspaces in the parameter space, and interpolate points within these subspaces $\widehat{\theta_i \theta_{\neg i}} = \frac{1}{N} \sum_i^N \alpha_i \theta_i$ given $N$ end-point parameters.
Given one train-time distribution $\mathbf{x}$ but $N$ different random initializations, 
they show interpolated points within these subspaces to construct diverse ensembles perform robustly
against end-point train-time distributions and label shift.
We would be among the first to construct a single subspace on non-identical train-time distributions as well as evaluate the subspace on non-identical / interpolatable test-time distributions.

Instead of a single distribution with multiple random initializations, 
we begin with a set of shifted distributions $\{ \mathbf{x_i} \}^N$ and a single constant (randomized) initialization. 
For a single test-time distribution (CIFAR10), we slice subsets of the train set and insert unique perturbations of a given type (backdoor / stylization / rotation) into each subset.
For multiple test-time distributions (CIFAR100), we use different task sets as each distribution.
Each model is a $\ell$-layer CNN trained with early-stopping at loss 1.0 (to accommodate computational load for training, hence limiting accuracy).
We load the $N$ sets in parallel.
At the end of each epoch, we compute the loss of each model's parameters with respect to their train-time sets.
We then compute the average cosine distance between each model's weight against all the other model's weights.
We add this distance multiplied by a distance coefficient $\beta=1.0$ to the total loss, and update each model's weights with respect to this total loss.

\begin{algorithm}[t]
  \footnotesize 
  \caption{CPS: Train}
  \SetKwInOut{Input}{Input}
  \SetKwInOut{Output}{Output}
  \SetKwProg{trainAgentSubspace}{trainSpace}{}{}
  \trainAgentSubspace{$(f, \{\theta_i\}^{N}, \{D_i\}^{N}, \beta, E$)}{
    \Input{Model $f$, Parameters $\{\theta_i\}^{N}$, Train set $\{D_i\}^{N}$,  Distance coefficient $\beta$, Epochs $E$}
    \Output{Trained parameters $\{\theta_i\}^{N}$}
    Constant initialization $\{\theta_{\textnormal{init}} \gets \theta_i\}^{N}$
    \smallbreak
    Train each parameter against their indexed train set in parallel per epoch \\
    \ForEach{\textnormal{epoch} $e \in E$}{
        \ForEach{$\theta_i, D_i \in  \{\theta_i\}^{N}, \{D_i\}^{N}$}{
            \ForEach{\textnormal{batch} $x, y \in D_i$}{
                $\hat{y} \gets f(x, \theta_i)$\\
                $\mathcal{L} \gets \ell(\hat{y}, y) + \beta \texttt{cos}(\{\theta_i\}^{N})$\\
                Backprop $\theta_i$ w.r.t. $\mathcal{L}$
            }
        }
    }
    \KwRet{$\{\theta_i\}^{N}$}\; 
  }
  \label{alg:train}
\end{algorithm}

\section{Evaluation}

We summarize below the experiments evaluated (detailed configurations in Appendix A.2).
\begin{itemize}[leftmargin=*]
    \item 
    \textbf{Change in parameter subspace dynamics: }
    We plot the changes in loss and cosine distance for training CPS with 3 train-time distributions \textit{(Figure \ref{fig:beta1_timeseries}, 
    Appendix A.3.1
    )}.
    We linearly-interpolate between tasks and end-point parameters \textit{(Figure \ref{fig:loss_landscape})}. 
    We plot the individual tasks along interpolation coefficients $\{\alpha_i^{X}\}^N$, and unique lowest-loss parameters that can correspond to a task along interpolation coefficients $\{\alpha_i^{\theta}\}^N$. 
    We also render an alternative continuous cross-sectional plot
    \textit{(Appendix A.3.3)}.
    The task end-points in these plots are: 3 tasks, 3 $\times$ same coarse label, train-time task set.
    
    \item 
    \textbf{CPS vs single test-time distributions: }
    We evaluate baselines and CPS trained on 3 subsets of perturbed CIFAR10
    against a range of test-time perturbations \textit{(Table \ref{tab:source_baselines})}. 
    We evaluate the relationship between CPS, number of perturbed sets trained on, and model capacity \textit{(Appendix A.3.4)}.

    \item 
    \textbf{CPS vs multiple test-time distributions: }
    We evaluate the insertion of a CPS-regularization term in training hypernetworks for continual learning \textit{(Table 3)}.
    We evaluate the improvement in mapping low-loss parameters in CPS
    \textit{(Table 2)}.
    We evaluate CPS (varying for over-parameterization w.r.t. number of tasks and model capacity) against seen/unseen tasks (varying for additional perturbation types) \textit{(Appendix A.3.8)}.
    We also evaluate varying model width against train-time distribution label set diversity 
    \textit{(Appendix A.3.6)}.
    We also evaluate varying model depth against varying distinct coarse label sets in train-time distributions \textit{(Appendix A.3.5)}.
    We also evaluate varying model depth against varying task label set diversity (different fine and coarse labels)  \textit{(Appendix A.3.7)}.

\end{itemize}

\subsection{Single Test-Time Distribution}

\noindent\textbf{(Methodology) Baselines. }
We baseline with CutMix data augmentation \citep{Yun_2019_ICCV}, 
adversarial training with PGD \citep{goodfellow2015explaining}, and 
backdoor adversarial training \citep{geiping2021doesnt}. 
We additionally evaluate a subspace and ensembling method, Fast Geometric Ensembles \citep{garipov2018loss}.
In a continual learning setting, we evaluate with hypernetworks \citep{vonoswald2020continual}.
We list all implementation details in Appendix A.2.

\noindent\textbf{(Methodology) Subspace Inference (Algorithm \ref{alg:inf}). }
Inferencing w.r.t. the \textbf{centre} of the subspace is denoted as computing a prediction $\bar{y} = \mathscr{f}(\theta^{*}; \hat{\mathbf{x}})$ w.r.t. a given input and the centre parameter point-estimate $\theta^{*} = \sum_i^N \frac{1}{N} \theta_i$ (the average of the end-point parameters). 
Inferencing w.r.t. an \textbf{ensemble} in the subspace is denoted as computing a mean prediction $\bar{y} = \frac{1}{M} \sum_j^M \mathscr{f}(\theta_j^{*}; \hat{\mathbf{x}})$ w.r.t. a given input and $M$ randomly-sampled parameter point-estimates in the subspace; we randomly sample $M=1000$ interpolation coefficients $\{ \alpha_{j, i}\}^{M \times N}$ to return an ensemble set of parameter point estimates $\{ \theta_j^{*} \}^M = \{ \alpha_{j, i}\}^{M \times N} \cdot \{\theta_i\}^N$.
Inferencing w.r.t. the maximum-accuracy (lowest-loss) \textbf{interpolated} point-estimate in the subspace is denoted as 
computing a prediction $\bar{y} = \mathscr{f}(\theta_{j^{*}}; \hat{\mathbf{x}})$ w.r.t. a given input and the lowest-loss parameter point-estimate 
where $j^{*} := \arg\min_{j \sim M} \mathcal{L}(\theta_j; \mathbf{x}, \mathbf{y})$; this is specifically a unique-task solution, where a low-loss parameter maps back to one task/distribution $\hat{\mathbf{x}} \mapsto \theta_{j^{*}}$.
For a multi-task solution, a low-loss parameter is mapped back to a set of $T$ task/distributions $\{\hat{\mathbf{x_t}}\}^T \mapsto \theta_{j^{*}}$, where $j^{*} := \arg\min_{j \sim M} \sum_t^T \mathcal{L}(\theta_j; \mathbf{x_t}, \mathbf{y_t})$.
Inferencing w.r.t. the lowest-loss \textbf{boundary} parameter in the subspace is denoted as computing a prediction $\bar{y} = \mathscr{f}(\theta_{i^*}; \hat{\mathbf{x}})$ w.r.t. a given input and lowest-loss boundary parameter,
where the latter is the parameter from a set of $N$ boundary or end-point parameters that returns the lowest loss 
 $i^{*} := \arg\min_{i \sim N} \mathcal{L}(\theta_i; \mathbf{x}, \mathbf{y})$.

\noindent\textbf{(Observation 1)}
\textit{
(Figure \ref{fig:beta1_timeseries}, Appendix A.3.1)
A parameter subspace can be found where the distance between end-point parameters to the centre is reduced.
}
The intended result from cosine distance minimization is not to converge cosine distance to 0 (e.g. using a larger capacity model can result in lower cosine distance between parameters); the intended result is to minimize the distance between parameters compared to without regularization, and through this process find an interplotable subspace in the parameter space.
Comparing the distance w.r.t. opposing end-point parameters and the centre in Figure \ref{fig:beta1_timeseries} and 
Appendix A.3.1, we find that the distances have comparatively converged when $\beta=1$ than $\beta=0$ for single test-time distributions, but the convergence in distance is not as apparent for multiple test-time distributions; the latter may suggest a minimum distance (interference) is required between certain tasks particularly if there is limited transferable features between them.

From Figure \ref{fig:beta1_timeseries} and Appendix A.3.1,
when training w.r.t.  a single test-time distribution, we observe that the cosine distance (w.r.t. subspace centre) first increases substantially before decreasing and continuing to stay at a low level.
As opposed to SGD performing point-estimate optimization, this phenomenon can be interpreted as SGD performing subspace optimization, in-line with Theorem 2.
Without distance regularization, the trajectory taken by SGD tends to depend more on the initialization point \citep{fort2020deep}.
In-line with \cite{NEURIPS2020_0607f4c7}, 
the parameters at peak cosine distance may be interpreted as a second, albeit unshared, pre-trained weight initialization 
that propagate parameters towards a same low-loss basin.

\noindent\textbf{(Observation 2)}
\textit{
(Table \ref{tab:source_baselines}, Appendix A.3.8)
Sampling parameter point-estimates in the CPS can attain a high average robustness accuracy across different perturbations or shifts.
}
Evaluated against the clean test-time distribution, we find that sampling point-estimates in the CPS can retain its accuracy across varying perturbation types (Table \ref{tab:source_baselines},
Appendix A.3.8).
For a single test-time distribution (CIFAR10) with varying perturbation types, 
we find that inference with the centre or ensemble
can yield a higher average accuracy across perturbed test-time distributions than wide-spectrum and/or niche defenses, particularly CPS trained on backdoored subsets. 
We find that the increased low-loss parameter mappings to the input space enables robustness for multiple perturbation types, in-line with Theorem 3. 
Though outperforming on average, CPS underperforms data augmentation with respect to test-time stylization and rotation.
Though a lowest-loss point-estimate can be found for most perturbation types, we note that a comparably-accurate one cannot be found for test-time stylization.
Though CPS trained on rotated sets can contain lowest-loss point-estimates for rotational perturbations, an averaging strategy (neither averaging the parameters to compute the centre, nor ensembling the predictions of 1000 parameters) does not work as well as other perturbation types. 
We achieve above-chance (CIFAR10 10\%, CIFAR100 20\%) performance on stylization and rotation, and show that a lowest-loss parameter can be interpolated within the CPS for these cases.
Other than a backdoor attack, training a CPS on any perturbed type of subset can attain similar performance.

\noindent\textbf{(Observation 3)}
\textit{
There are efficient methods to query parameters from the CPS without explicitly/separately storing large subsets of the CPS. 
}
For single test-time distribution,
storing a single set of parameters (capacity $\mathcal{C}$) is feasible (subspace centre) and yields comparable average accuracy across perturbation types. 
Given ensembling outperforms the centre in certain cases, storage for constructing ensembling is $N \times$ storage of each end-point parameter ($N \times \mathcal{C}$).
For multiple test-time distributions,
a hypernetwork is found to be an efficient generalization function of the CPS, yielding better performance than without compression, and retaining comparable performance to manually interpolating point-estimates. 
Storage for a hypernetwork is also lower than manually storing subsets of the CPS, though higher than storing the parameters of a single base learner parameter ($\frac{\mathcal{C}_{\textnormal{h}}}{\mathcal{C}} > 1.0$).

\begin{table*}[t]
\centering
\parbox{\textwidth}{
\centering
\resizebox{\textwidth}{!}{
\begin{tabular}{|lrcccccc}
\multicolumn{2}{c}{} & \multicolumn{2}{c}{Acc w.r.t. seen distributions} & \multicolumn{4}{c}{Acc w.r.t. unseen shift distributions} \\
\multicolumn{2}{c}{} & Clean Test Set & Backdoor Attack & Adversarial Attack & Random Permutations & Stylization & Rotation \\ \hline \hline
\multicolumn{2}{|l}{Data Augmentation 
} & \multicolumn{1}{|c}{$55.6$} & \multicolumn{1}{c|}{$49.0 \pm 25.2$} & \multicolumn{1}{c}{$40.4$} & \multicolumn{1}{c}{$49.6 \pm 23.2$} & \multicolumn{1}{c}{$49.0 \pm 21.3$} & \multicolumn{1}{c|}{$51.4 \pm 23.7$} \\ \hline
\multicolumn{2}{|l}{Adversarial Training 
} & \multicolumn{1}{|c}{$55.1$} & \multicolumn{1}{c|}{$37.2 \pm 22.4$} & \multicolumn{1}{c}{$50.0$} & \multicolumn{1}{c}{$41.8 \pm 22.4$} & \multicolumn{1}{c}{$40.0 \pm 19.9$} & \multicolumn{1}{c|}{$40.8 \pm 23.3$} \\ \hline
\multicolumn{2}{|l}{Backdoor Adversarial Training 
} & \multicolumn{1}{|c}{$52.3$} & \multicolumn{1}{c|}{$45.0 \pm 23.5$} & \multicolumn{1}{c}{$31.2$} & \multicolumn{1}{c}{$36.2 \pm 21.0$} & \multicolumn{1}{c}{$37.6 \pm 21.5$} & \multicolumn{1}{c|}{$39.2 \pm 20.5$} \\ \hline
\multicolumn{2}{|l}{Fast Geometric Ensembling 
} & \multicolumn{1}{|c}{$55.4$} & \multicolumn{1}{c|}{$32.0 \pm 11.9$} & \multicolumn{1}{c}{$32.5$} & \multicolumn{1}{c}{$29.2 \pm 18.7$} & \multicolumn{1}{c}{$25.2 \pm 18.6$} & \multicolumn{1}{c|}{$31.4 \pm 18.5$} \\ \hline \hline
\multicolumn{1}{|r}{\multirow{3}{*}{CPS (backdoor)}}
& centre & \multicolumn{1}{|c}{$55.3 \pm 15.8$} & \multicolumn{1}{c|}{$55.3 \pm 15.8$} & \multicolumn{1}{c}{$47.7 \pm 2.5$} & \multicolumn{1}{c}{$54.7 \pm 21.7$} & \multicolumn{1}{c}{$23.9 \pm 4.0$} & \multicolumn{1}{c|}{$28.2 \pm 19.8$} \\
& ens. (mean) & \multicolumn{1}{|c}{$55.3 \pm 15.7$} & \multicolumn{1}{c|}{$55.5 \pm 15.7$} & \multicolumn{1}{c}{$47.3 \pm 2.2$} & \multicolumn{1}{c}{$54.2 \pm 20.8$} & \multicolumn{1}{c}{$23.5 \pm 3.6$} & \multicolumn{1}{c|}{$28.2 \pm 19.1$} \\
& intp. (max) & \multicolumn{1}{|c}{$57.5 \pm 15.5$} & \multicolumn{1}{c|}{$57.5 \pm 15.5$} & \multicolumn{1}{c}{$60.5 \pm 1.2$} & \multicolumn{1}{c}{$61.7 \pm 19.8$} & \multicolumn{1}{c}{$27.0 \pm 3.8$} & \multicolumn{1}{c|}{$30.2 \pm 19.1$} \\
\hline
\multicolumn{1}{|r}{\multirow{3}{*}{CPS (stylization)}} 
& centre & \multicolumn{1}{|c}{$57.5 \pm 17.8$} & \multicolumn{1}{c|}{$21.2 \pm 1.4$} & \multicolumn{1}{c}{$50.6 \pm 2.1$} & \multicolumn{1}{c}{$50.3 \pm 19.4$} & \multicolumn{1}{c}{$24.2 \pm 2.4$} & \multicolumn{1}{c|}{$26.1 \pm 19.5$} \\
& ens. (mean) & \multicolumn{1}{|c}{$57.3 \pm 18.3$} & \multicolumn{1}{c|}{$20.8 \pm 1.4$} & \multicolumn{1}{c}{$50.4 \pm 2.3$} & \multicolumn{1}{c}{$59.2 \pm 19.3$} & \multicolumn{1}{c}{$24.1 \pm 2.8$} & \multicolumn{1}{c|}{$26.1 \pm 19.5$} \\
& intp. (max) & \multicolumn{1}{|c}{$59.7 \pm 18.2$} & \multicolumn{1}{c|}{$22.7 \pm 1.6$} & \multicolumn{1}{c}{$60.7 \pm 1.8$} & \multicolumn{1}{c}{$50.1 \pm 19.4$} & \multicolumn{1}{c}{$27.4 \pm 3.4$} & \multicolumn{1}{c|}{$29.1 \pm 19.2$} \\
\hline
\multicolumn{1}{|r}{\multirow{3}{*}{CPS (rotation)}}
& centre & \multicolumn{1}{|c}{$19.0 \pm 1.8$} & \multicolumn{1}{c|}{$18.1 \pm 6.1$} & \multicolumn{1}{c}{$19.0 \pm 1.4$} & \multicolumn{1}{c}{$19.3 \pm 12.7$} & \multicolumn{1}{c}{$13.1 \pm 1.1$} & \multicolumn{1}{c|}{$27.4 \pm 20.3$} \\
& ens. (mean) & \multicolumn{1}{|c}{$19.2 \pm 1.8$} & \multicolumn{1}{c|}{$25.3 \pm 9.2$} & \multicolumn{1}{c}{$18.1 \pm 1.2$} & \multicolumn{1}{c}{$19.2 \pm 11.0$} & \multicolumn{1}{c}{$13.1 \pm 1.5$} & \multicolumn{1}{c|}{$26.9 \pm 19.3$} \\
& intp. (max) & \multicolumn{1}{|c}{$59.6 \pm 17.2$} & \multicolumn{1}{c|}{$38.2 \pm 1.2$} & \multicolumn{1}{c}{$58.1 \pm 1.6$} & \multicolumn{1}{c}{$52.8 \pm 18.9$} & \multicolumn{1}{c}{$23.7 \pm 3.1$} & \multicolumn{1}{c|}{$48.5 \pm 26.6$} \\
\hline
\end{tabular}
}
}
\caption{
\textit{CPS vs single test-time distributions: }
Baselines and CPS trained on 3 subsets of perturbed CIFAR10 (backdoor / stylization / rotation) 
against
varying shifts w.r.t. test-time distributions (mean $\pm$ std w.r.t. number of test-time samples):
clean test set (out of 3),
backdoor attack (out of 3),
adversarial attack (out of 3),
random permutations (out of 100),
stylization (out of 100), 
and rotation (out of 100).
Base model is 3-layer CNN.
The backdoor attack is evaluated w.r.t. clean samples (higher accuracy is higher robustness), not w.r.t. poisoned labels.
}
\label{tab:source_baselines}
\end{table*}

\begin{figure*}[h]
    \centering
    \includegraphics[width=0.245\textwidth]{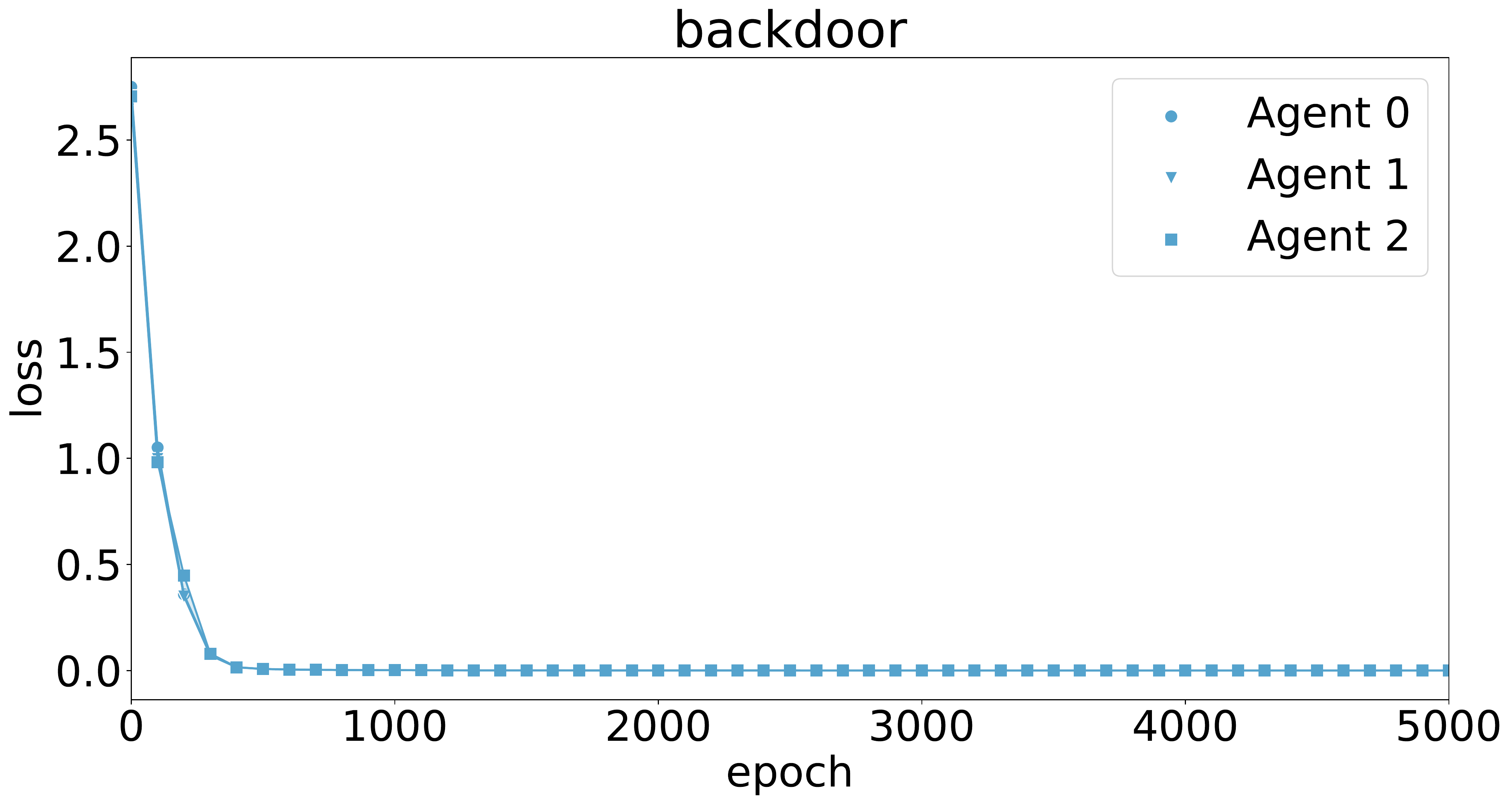}
    \includegraphics[width=0.245\textwidth]{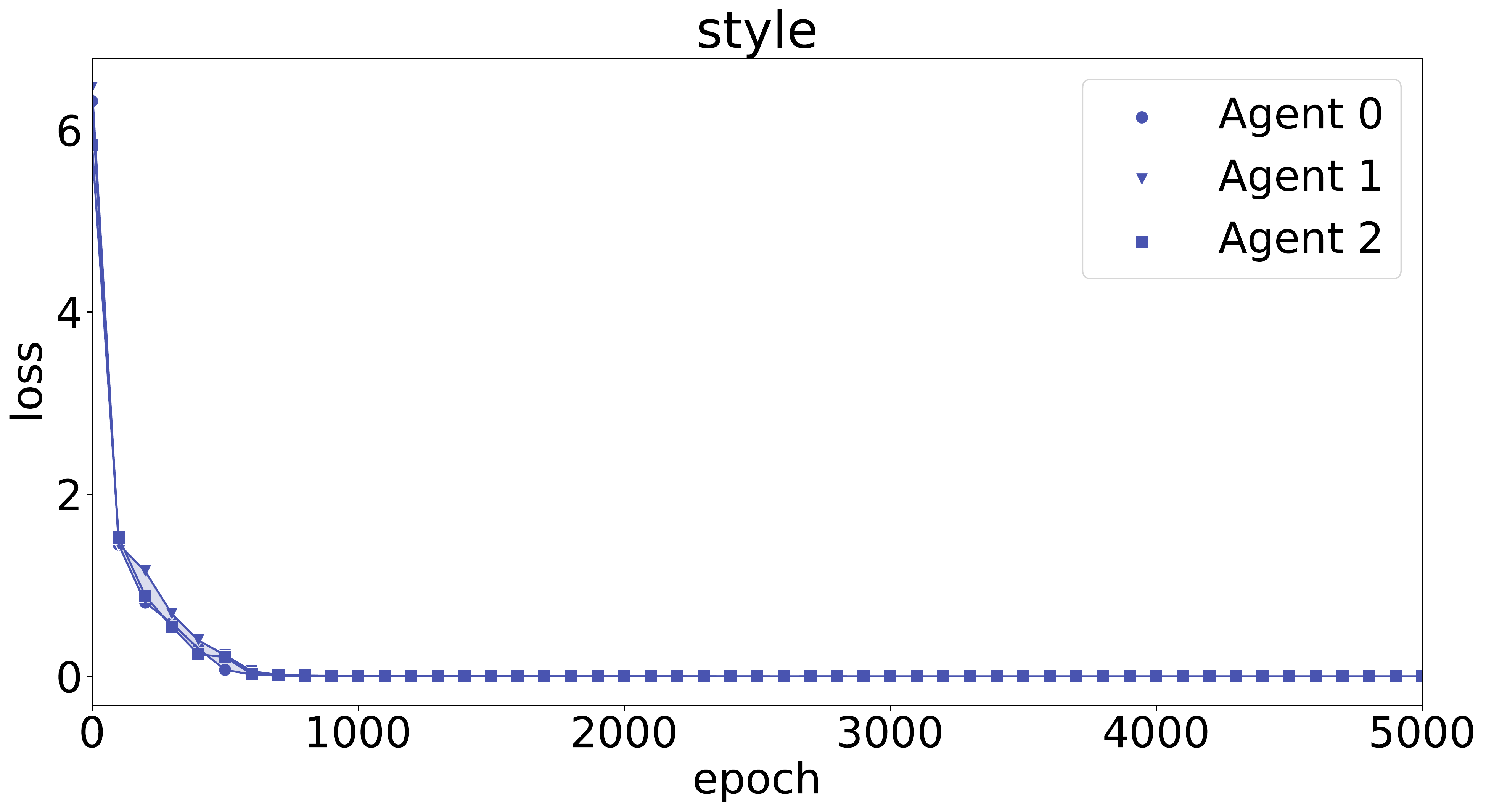}
    \includegraphics[width=0.245\textwidth]{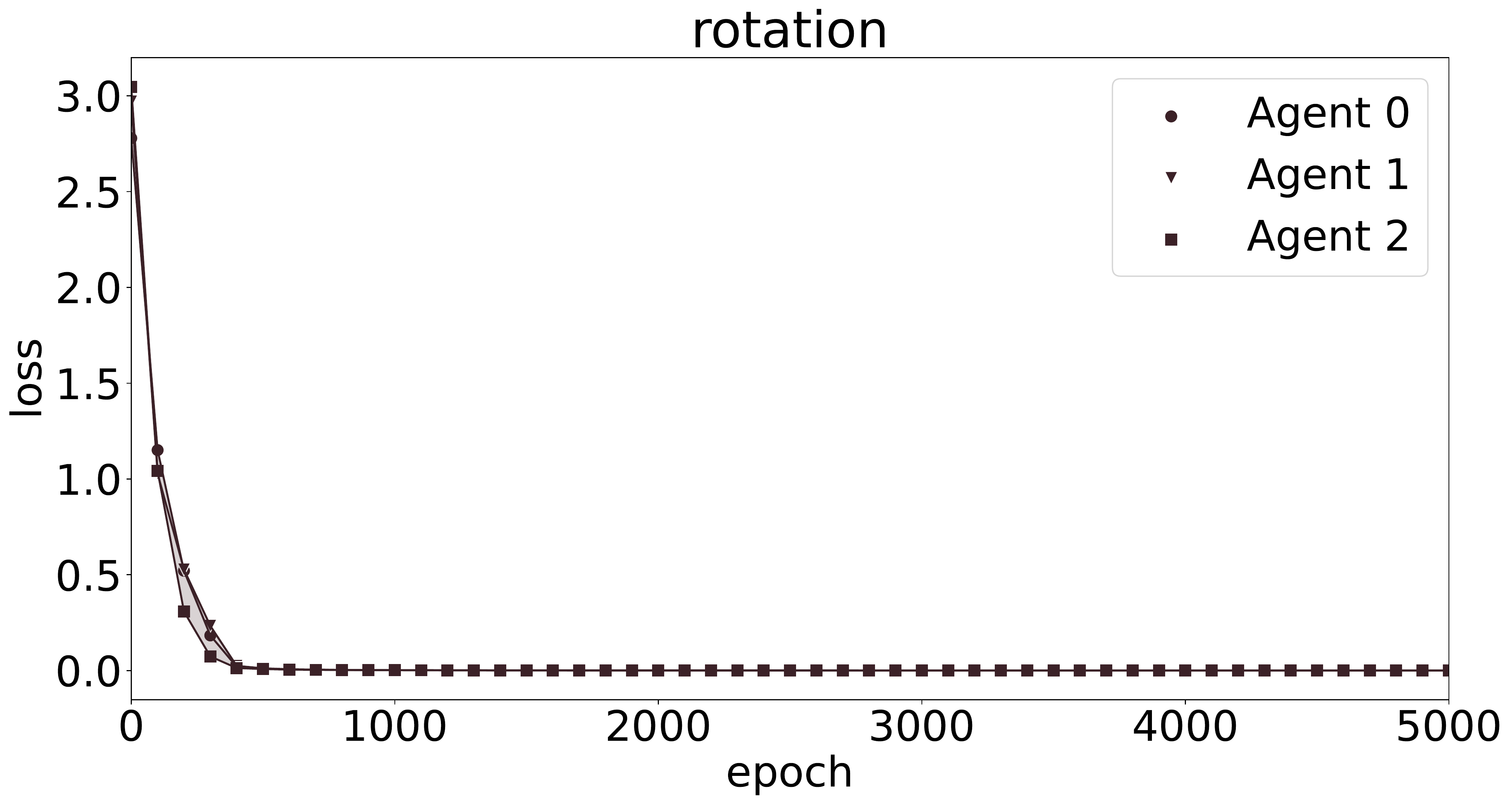}
    \includegraphics[width=0.245\textwidth]{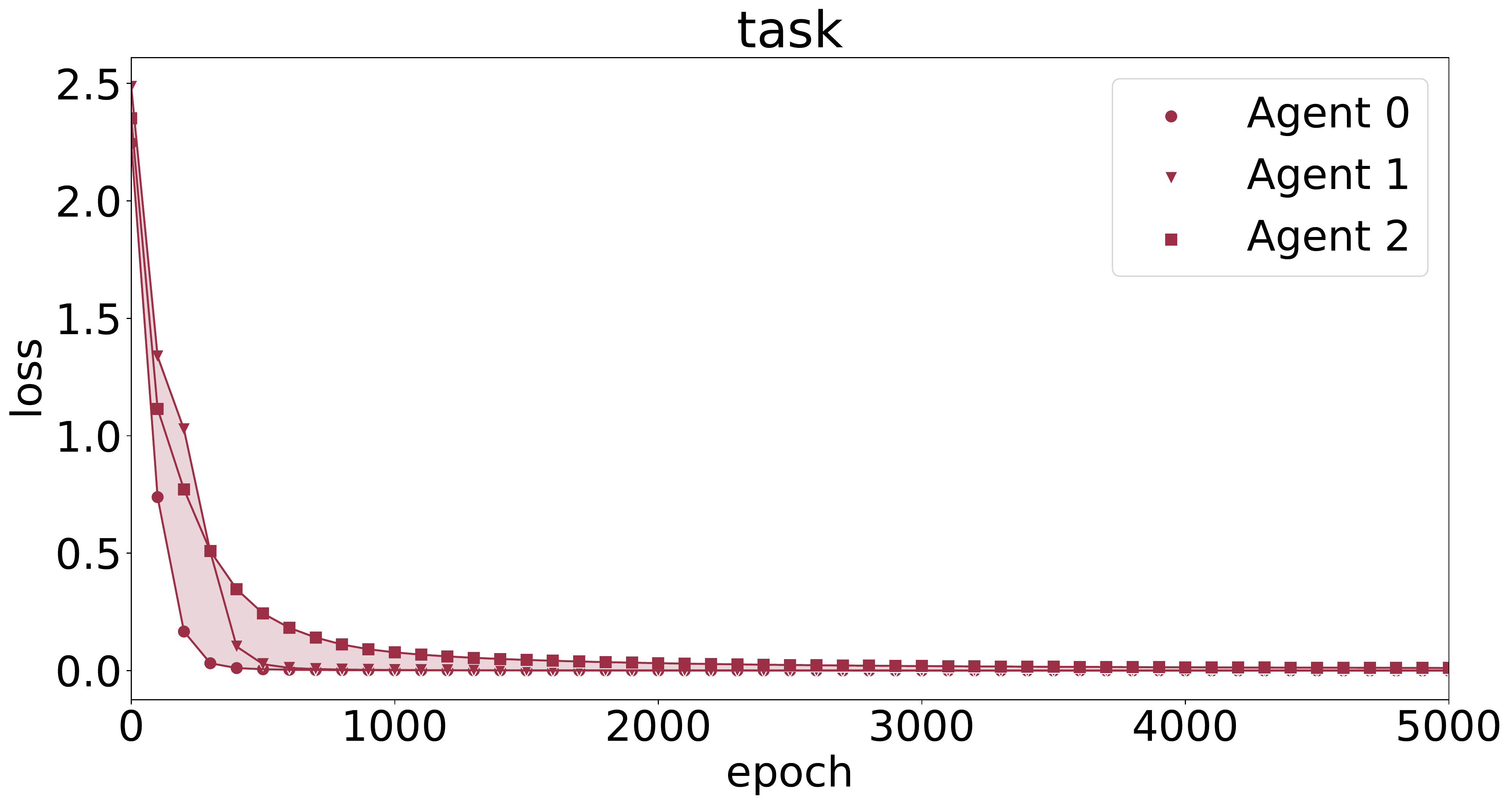}
    \includegraphics[width=0.245\textwidth]{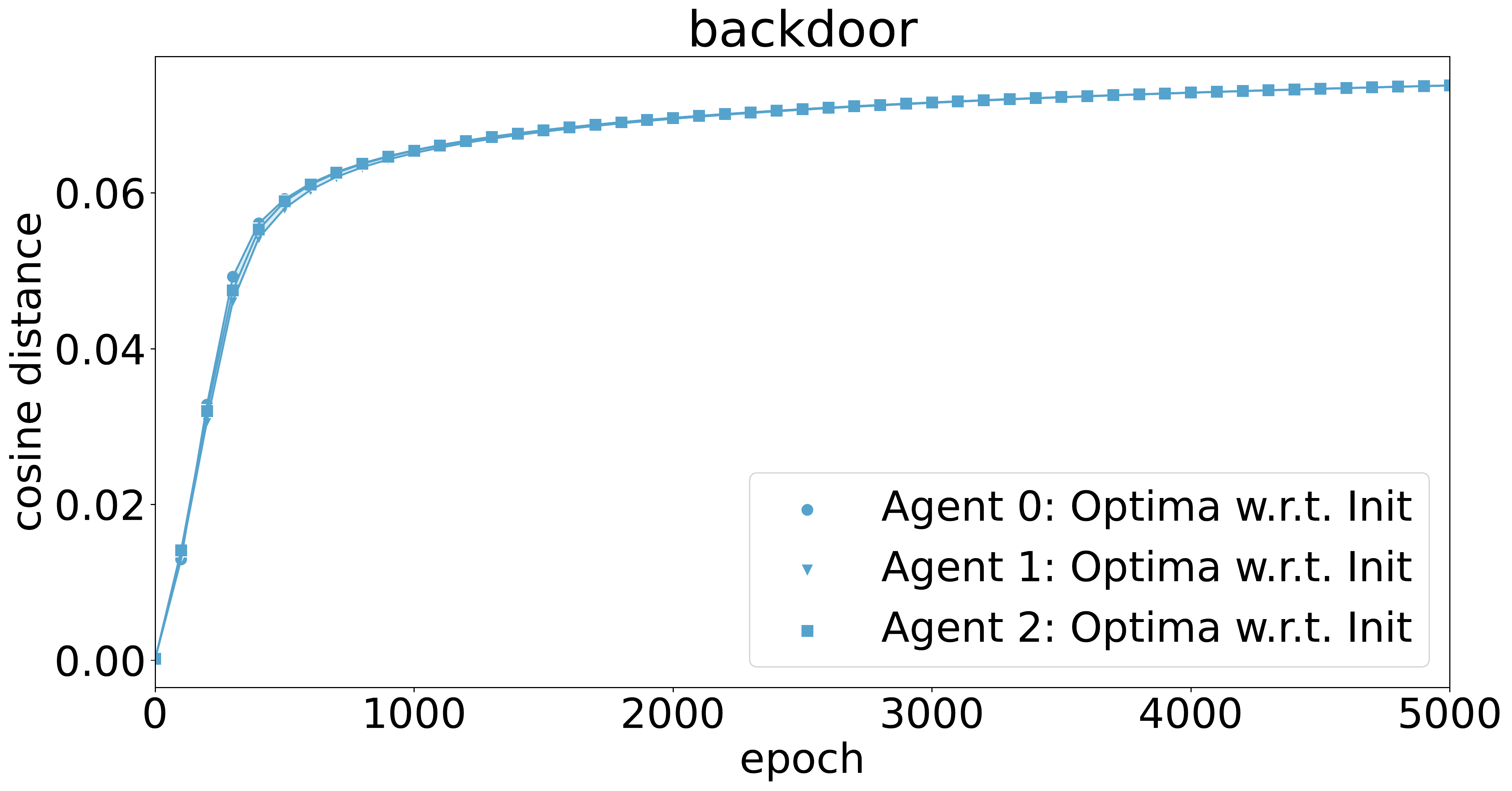}
    \includegraphics[width=0.245\textwidth]{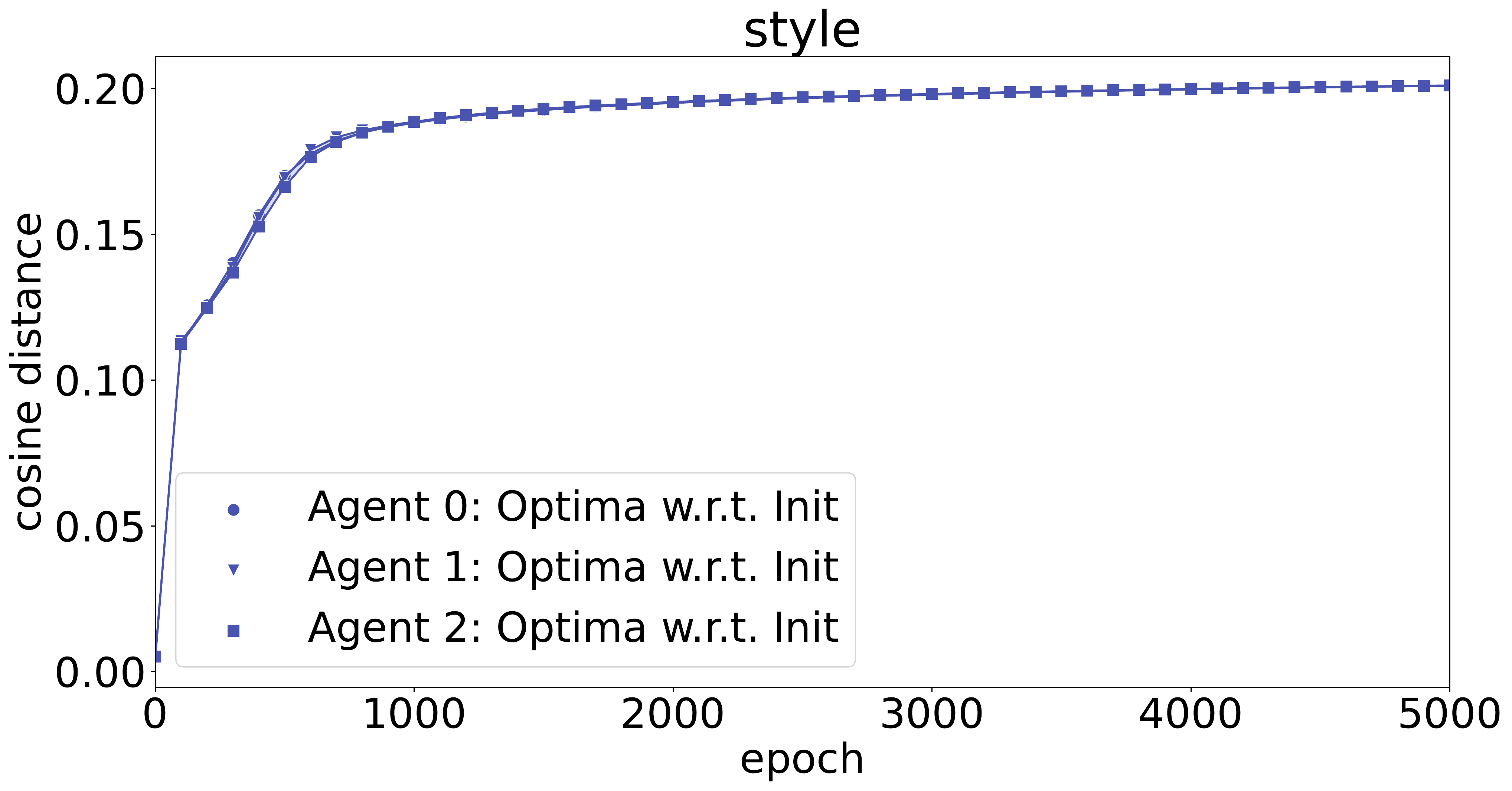}
    \includegraphics[width=0.245\textwidth]{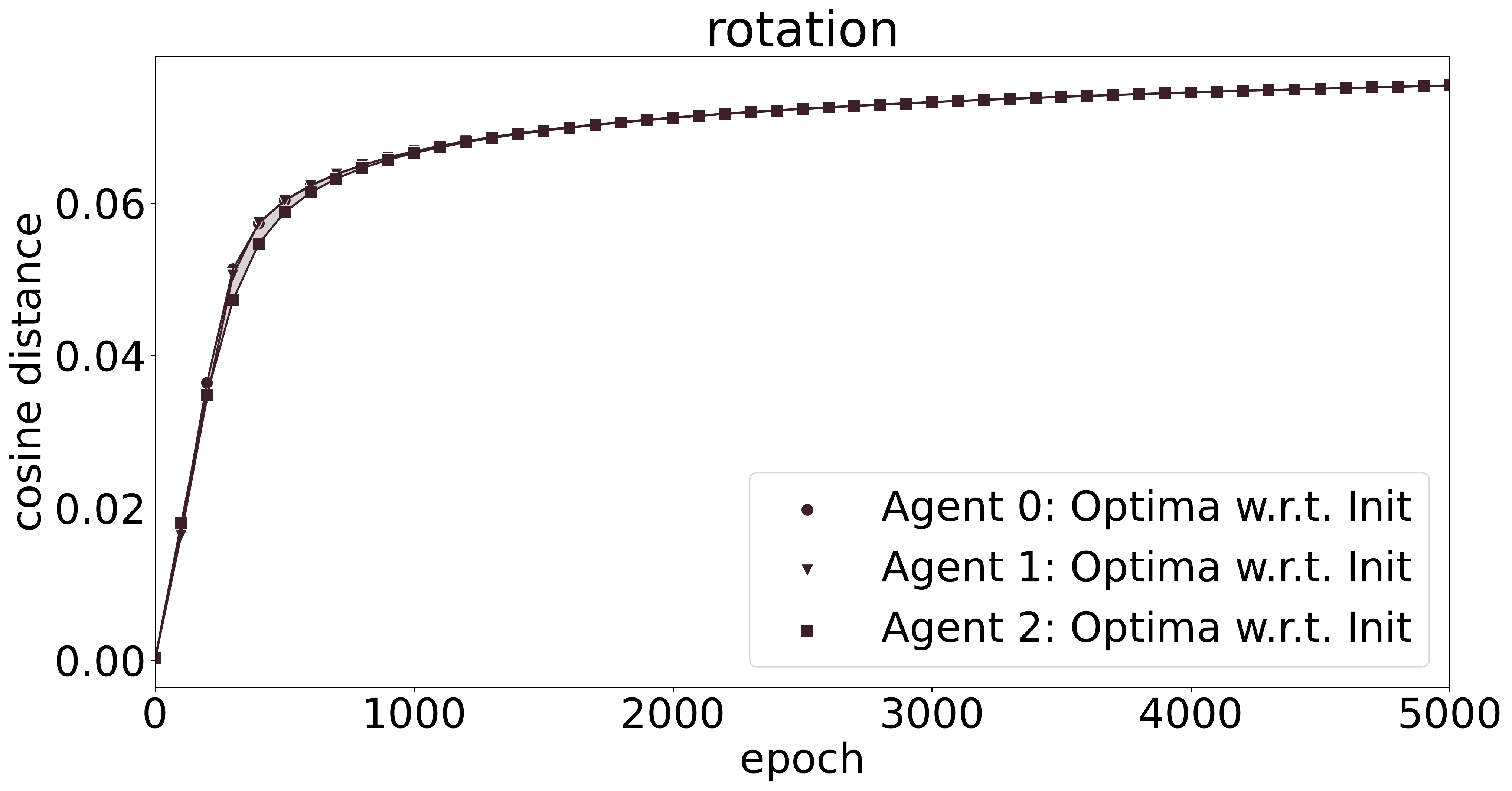}
    \includegraphics[width=0.245\textwidth]{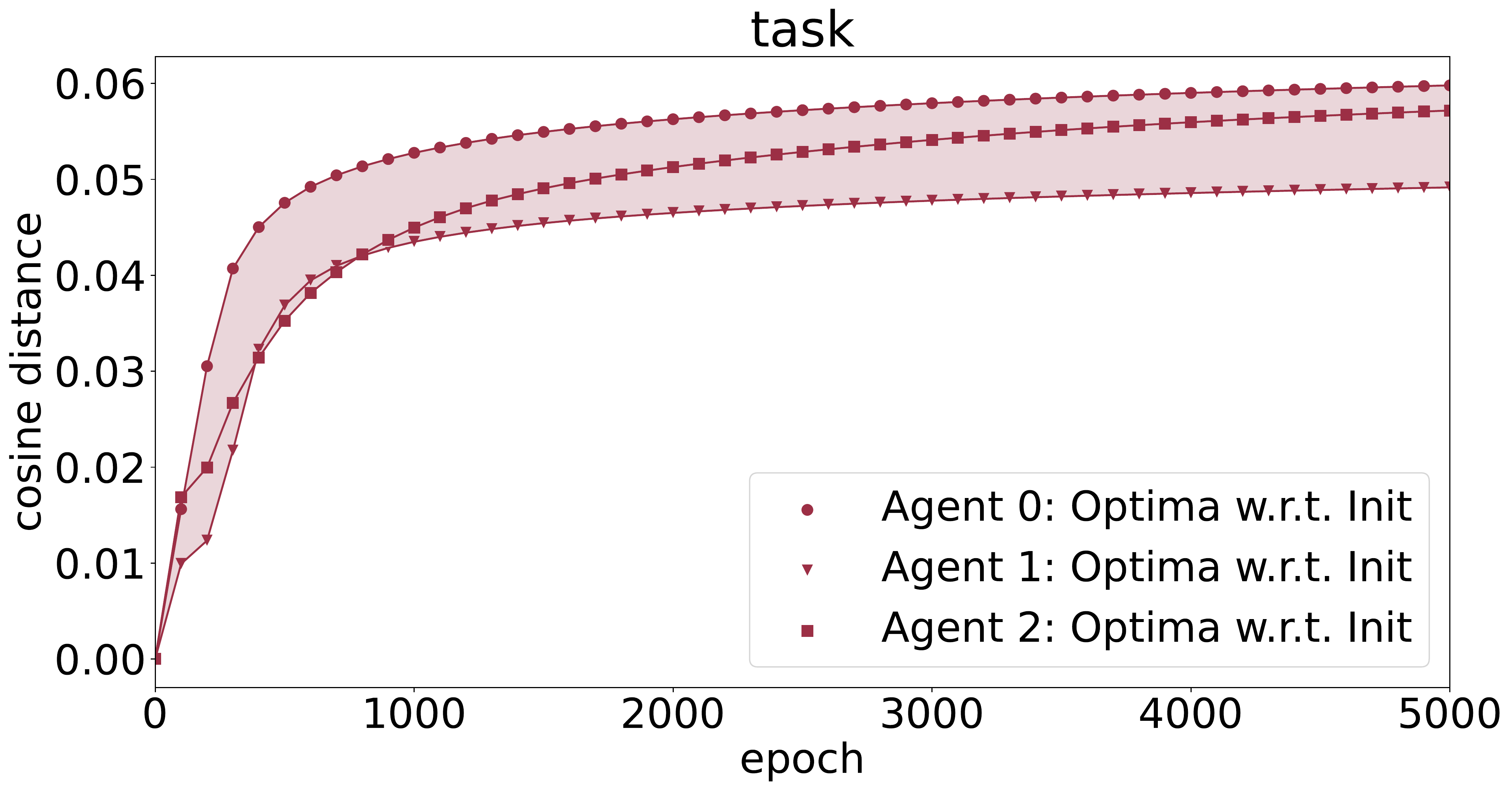}
    \includegraphics[width=0.245\textwidth]{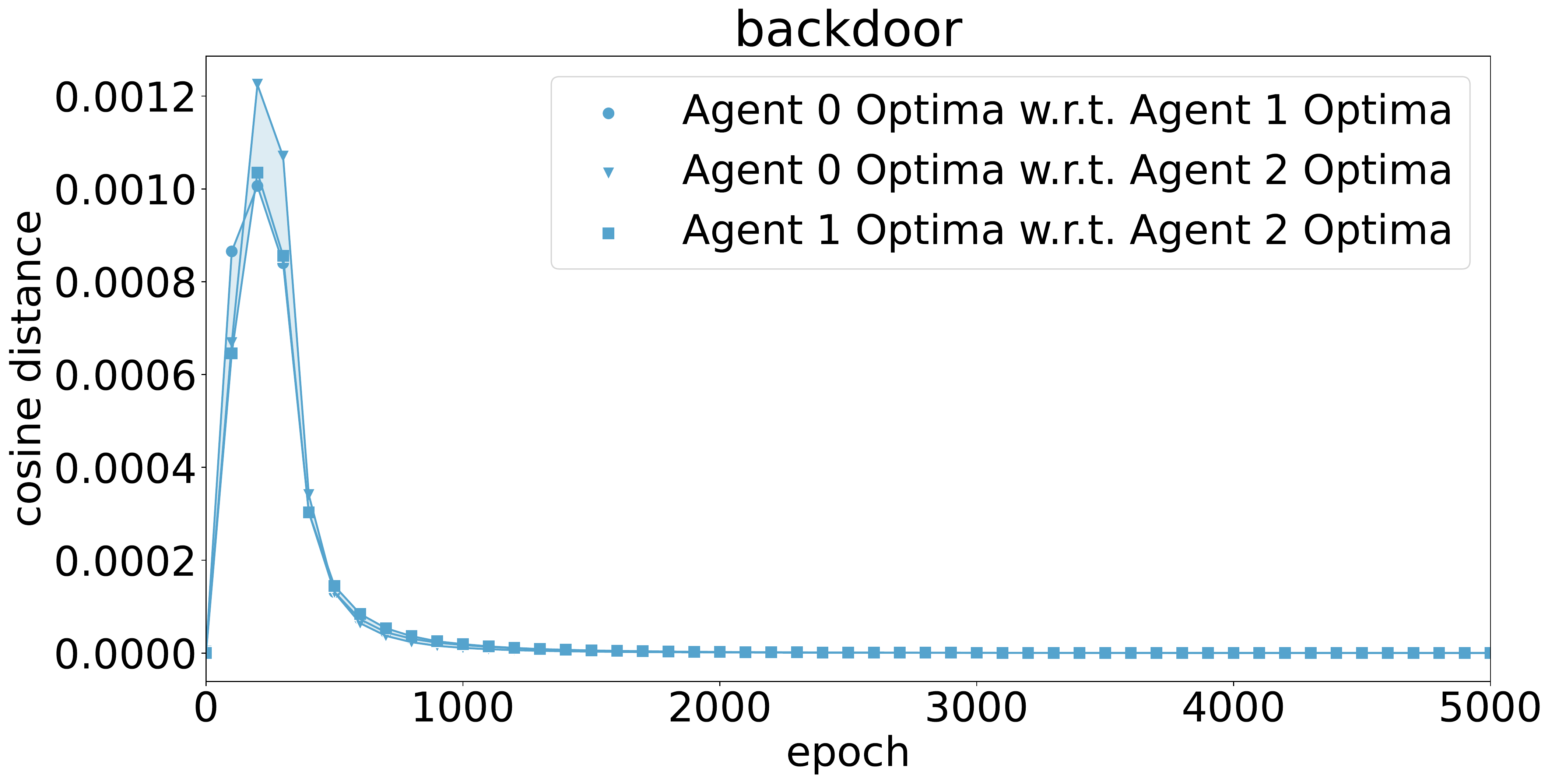}
    \includegraphics[width=0.245\textwidth]{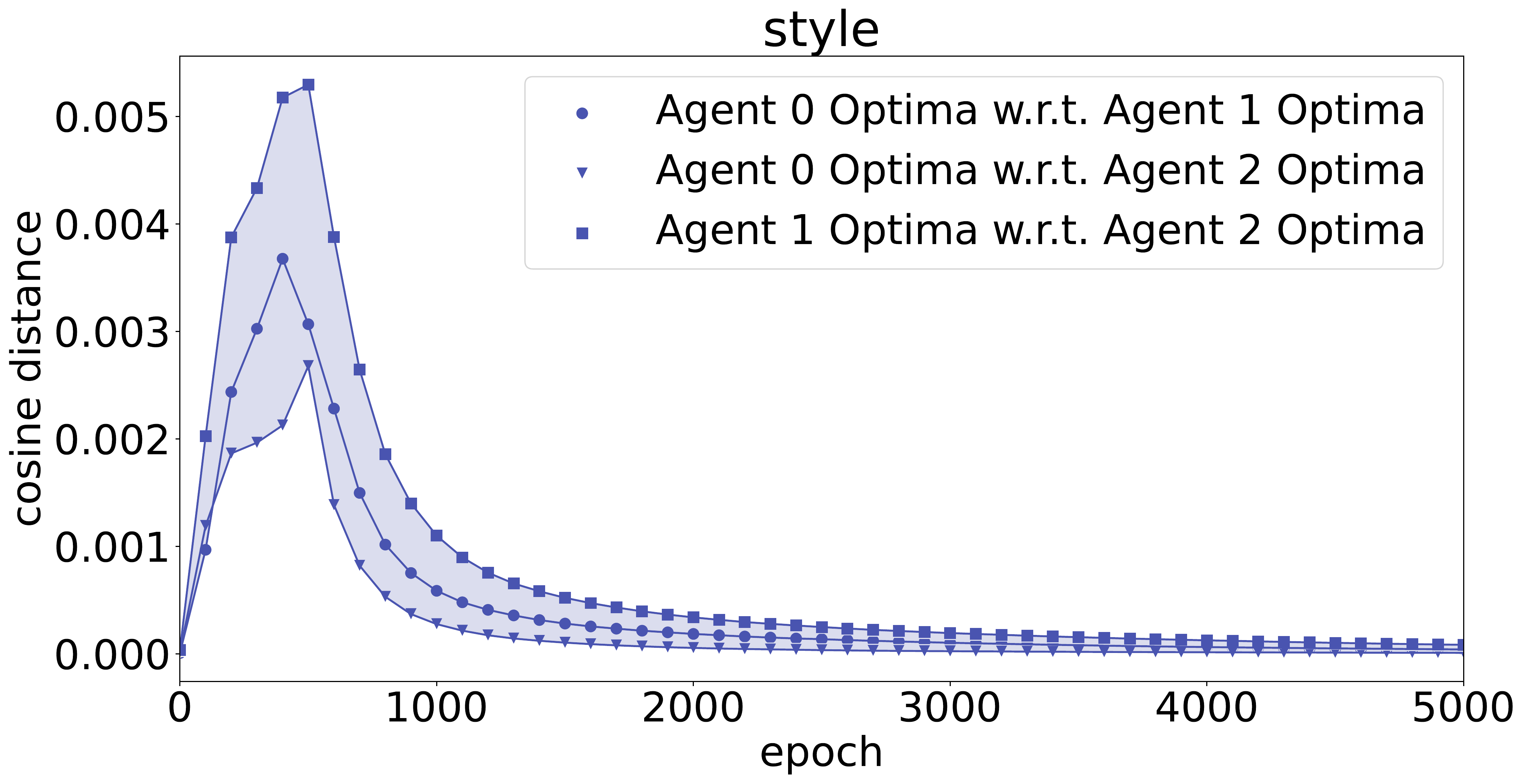}
    \includegraphics[width=0.245\textwidth]{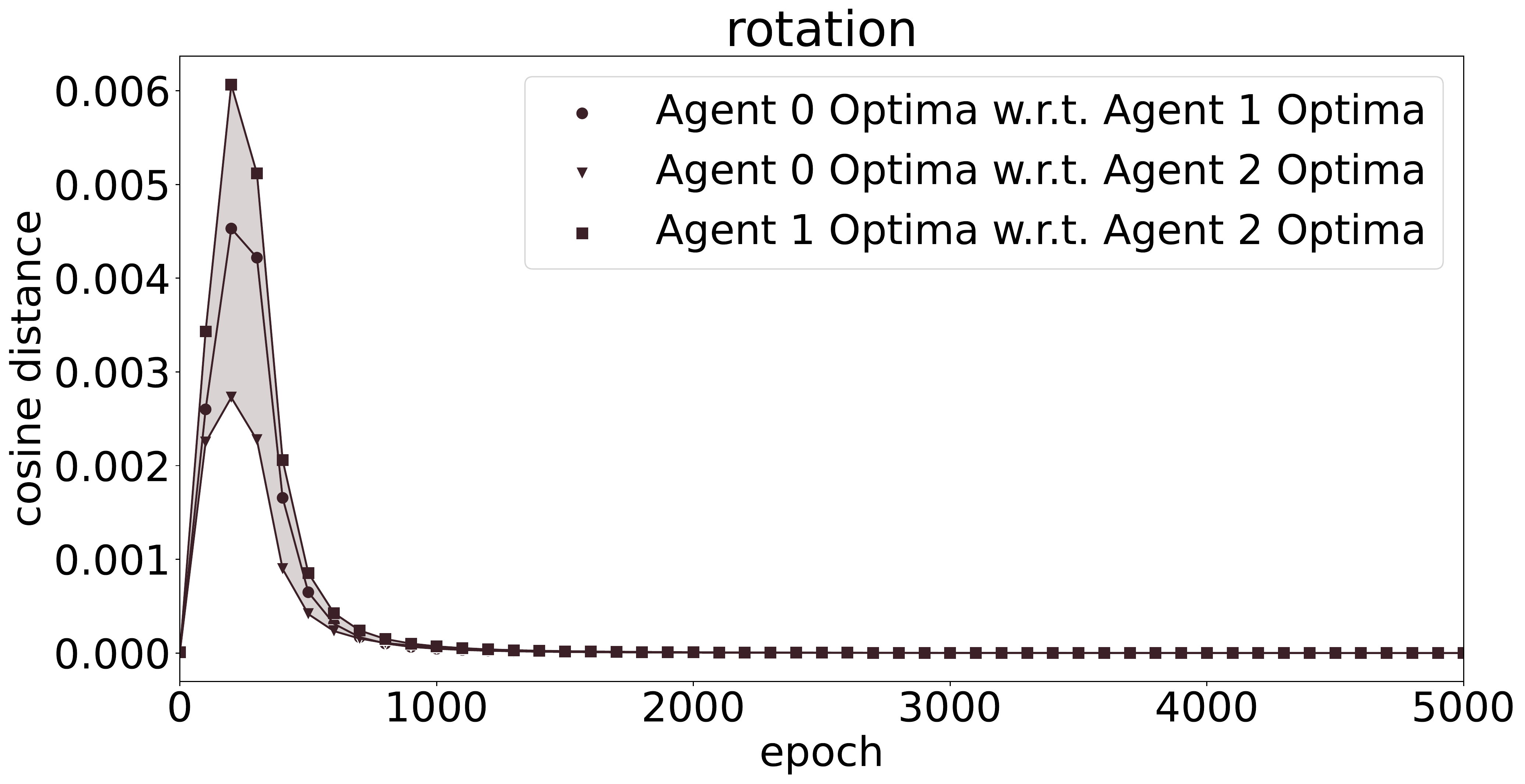}
    \includegraphics[width=0.245\textwidth]{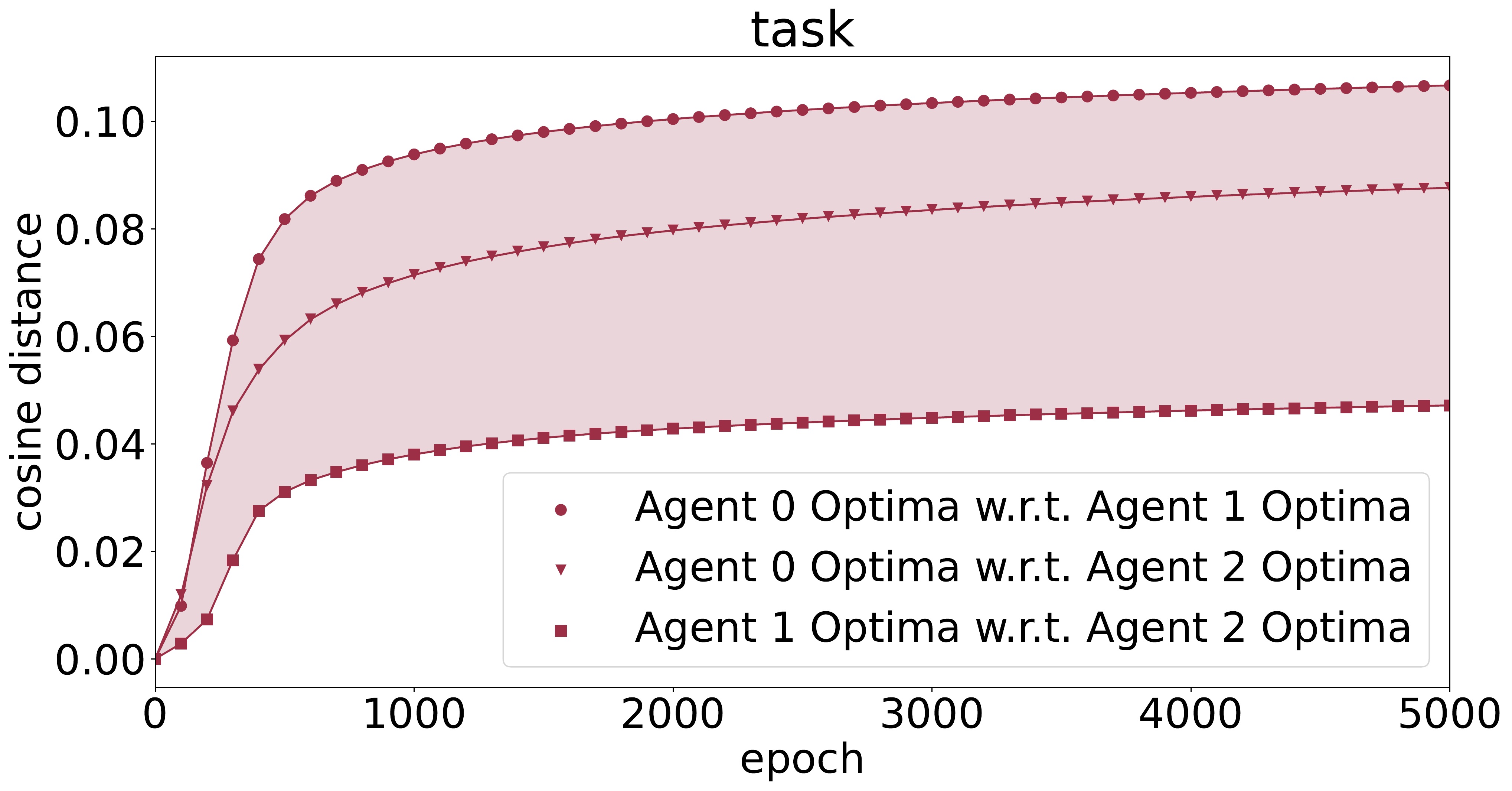}
    \includegraphics[width=0.245\textwidth]{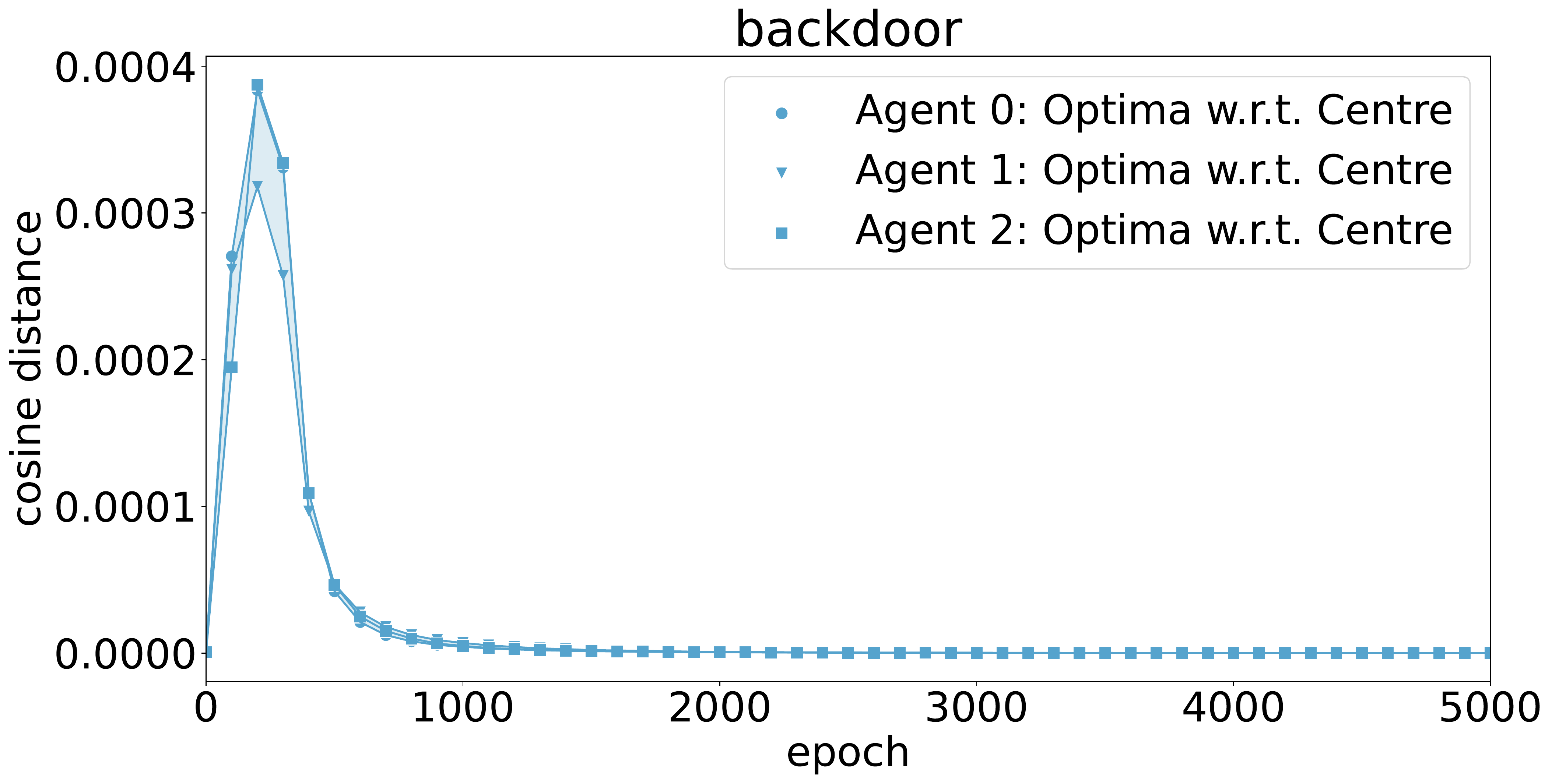}
    \includegraphics[width=0.245\textwidth]{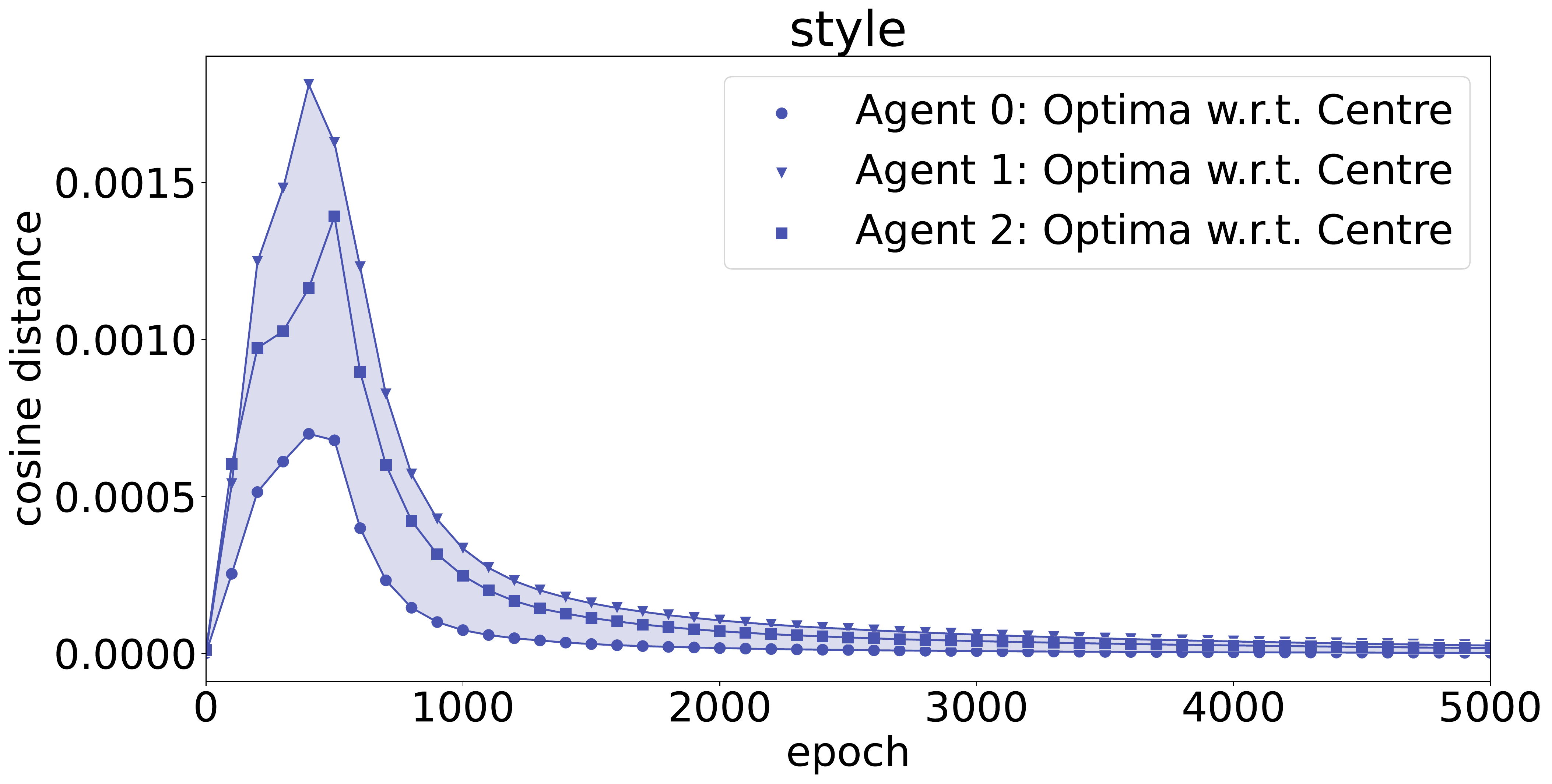}
    \includegraphics[width=0.245\textwidth]{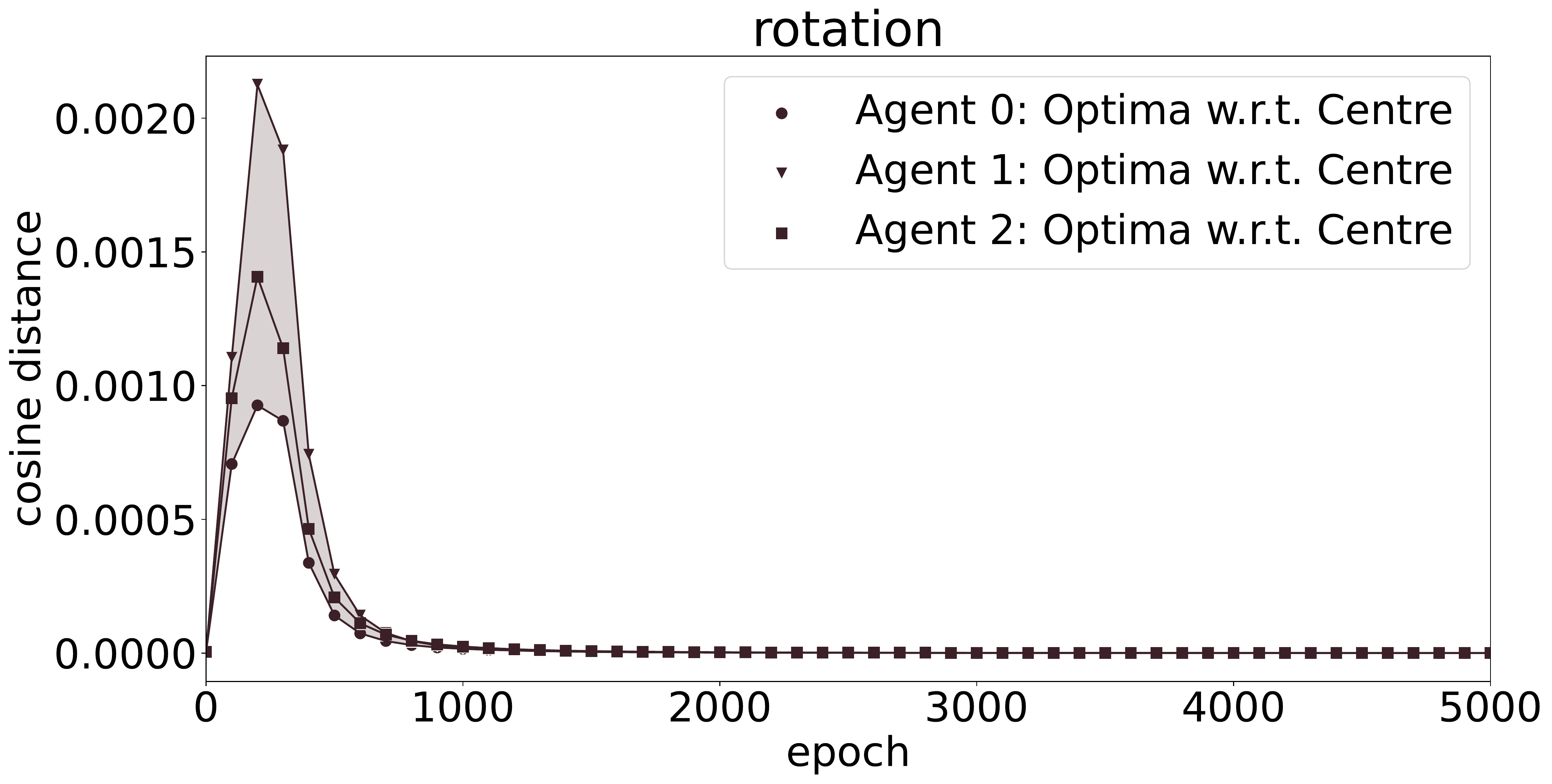}
    \includegraphics[width=0.245\textwidth]{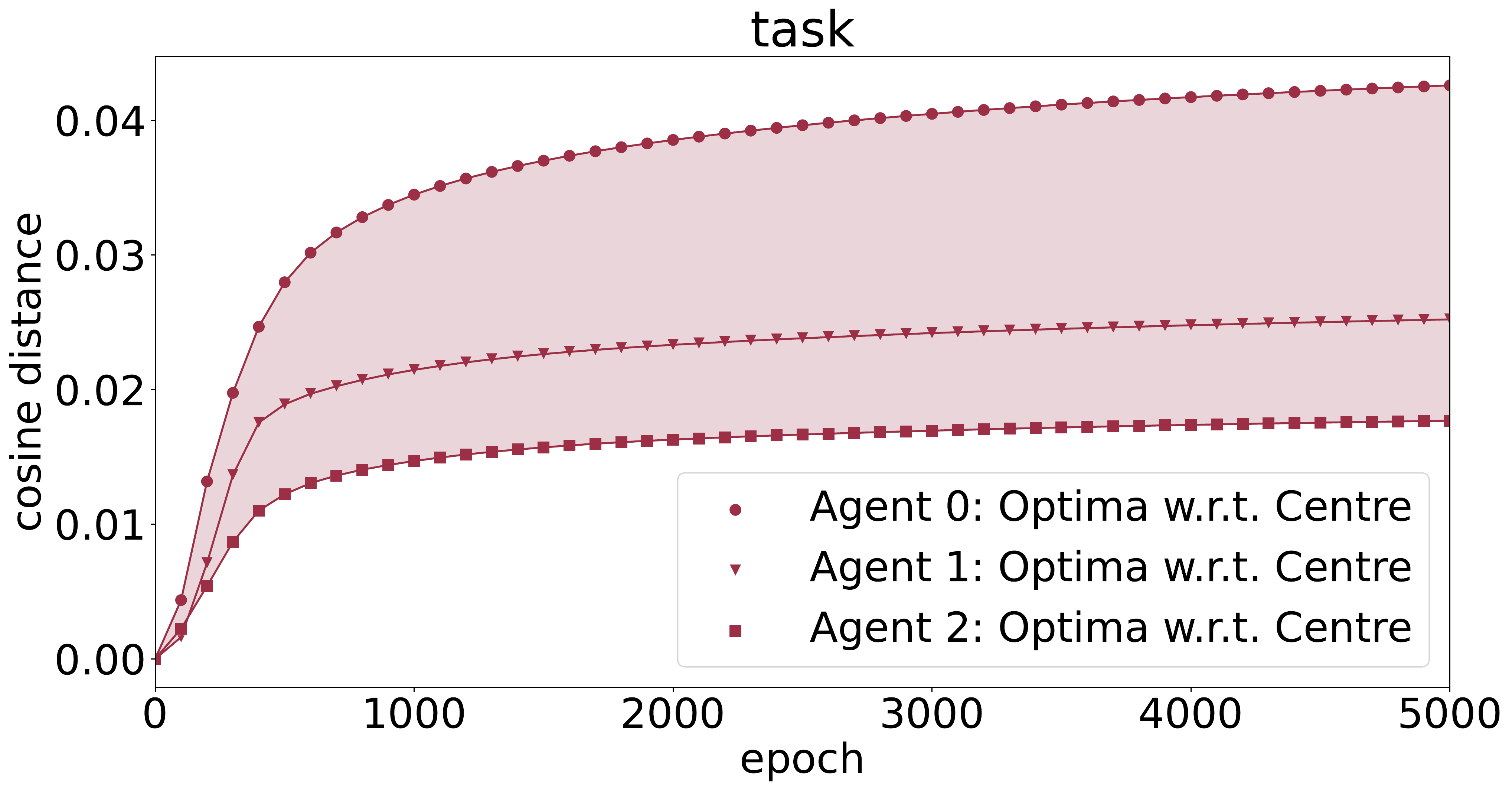}
    \caption{
    \textit{Change in parameter subspace dynamics: }
    Loss and cosine distance per epoch during training of CPS with 3 train-time distributions ($\beta=1.0$).
    }
    \label{fig:beta1_timeseries}
\end{figure*}

\subsection{Multiple Test-Time Distributions}

\noindent\textbf{(Methodology) Hypernetworks. }
We evaluate CPS-regularization in hypernetworks, 
an existing continual learning strategy.
In-line with \citet{10.5555/3305890.3306093, vonoswald2020continual},
we adopt the Split CIFAR10/100 benchmark,
where we train on CIFAR10 first, then train sequentially on the next 5 sets of 10-label CIFAR100.
All tasks share the same 10 coarse labels.

A \textbf{hypernetwork} $\mathscr{h}(\mathbf{x}, I) = \mathscr{f}(\mathbf{x}; \mathscr{mf}(\theta_{\mathscr{mf}}; I))$ 
is a pair of learners, the base learner $\mathscr{f}: \mathcal{X} \mapsto \mathcal{Y}$ and meta learner $\mathscr{mf}: \mathcal{I} \mapsto \Theta_{\mathscr{f}}$,
such that for the conditioning input $I$ of input $\mathbf{x}$ (where $\mathcal{X} \mapsto \mathcal{I}$),
$\mathscr{mf}$ produces the base learner parameters $\theta_{I} = \mathscr{mf}(\theta_{\mathscr{mf}}; I)$.
The function $\mathscr{mf}(\theta_{\mathscr{mf}}; I)$ takes a conditioning input $I$ 
to returns parameters $\theta_{I} \in \Theta_{\mathscr{f}}$ for $\mathscr{f}$. 
The meta learner parameters and each base learner parameters reside in their distinct parameter spaces $\theta_{\mathscr{mf}} \in \Theta_{\mathscr{mf}}$ and $\theta_{\mathscr{f}} \in \Theta_{\mathscr{f}}$.
The learner $\mathscr{f}$ takes an input $\mathbf{x}$ and returns an output $\bar{y} = \mathscr{f}(\theta_{I}; \mathbf{x})$ that depends on both $\mathbf{x}$ and the task-specific input $I$. 
$T$ is number of tasks, $t$ is index of specific task being evaluated ($t^{*}$ being current task), $\omega$ is task regularizer.
$\theta_{\mathscr{mf}}$ is the hypernetwork parameters at the current task's training timestep, 
$\theta_{\mathscr{mf}}^{*}$ is the hypernetwork parameters before attempting to learn task $t^{*}$, 
and $\Delta \theta_{\mathscr{mf}}$ is the candidate parameter change computed by the optimizer (Adam). 
We do not use any separate conditioning input (i.e. task embeddings), and use the test-time input as the primary argument, i.e. $\mathcal{I} \equiv \mathcal{X}$. 
\begin{equation}
\small
\begin{split}
\mathcal{L}_{\mathscr{mf}} 
&=
\mathcal{L}_{t^{*}}(\mathscr{mf}(\theta_{\mathscr{mf}}; \mathbf{x_{t^{*}}}); \mathbf{x_{t^{*}}}, \mathbf{y_{t^{*}}}) \\
& + 
\frac{\omega}{t^{*}-1} \sum_{t}^{t^{*}-1} 
\bigg|\bigg| 
\mathscr{f}(\mathscr{mf}(\theta_{\mathscr{mf}}^{*}; \mathbf{x_t}); \mathbf{x_t}, \mathbf{y_t})
-
\mathscr{f}(\mathscr{mf}(\theta_{\mathscr{mf}} + \Delta \theta_{\mathscr{mf}}; \mathbf{x_t}); \mathbf{x_t}, \mathbf{y_t})
\bigg|\bigg|^2 \\
&+
\colorboxed[rgb]{0.827, 0.827, 0.827}{\frac{\beta}{t^{*}-1} \sum_t^{t^{*}-1} \texttt{dist}(\theta_{\mathscr{mf}}, \theta_{\mathscr{mf}, t})}
\end{split}
\label{equation:hloss}
\end{equation}
Starting from a random initialization, 
we train the hypernetwork first on CIFAR10 (interpretable as re-initializing the network with pre-trained weights, which findings in \cite{NEURIPS2020_0607f4c7} may indicate that subsequent task parameters may reside in a shared low-loss basin), and store these parameters $\theta_{\mathscr{mf}, t}$ in a parameter set $\{\theta_{\mathscr{mf}, t}\}^T$.
Then we train on a subsequent CIFAR100 task with the Eqt. \ref{equation:hloss} loss function, 
where we compute loss w.r.t. the inputs and current timestep's parameters, 
change in loss w.r.t. a previous task between using the proposed parameters and the parameters last updated at that task (stored in the parameter set; this is enumerated for all past tasks in sequence), 
and the distance between the current parameters and all prior parameters.
Unlike our prior (multi-task) implementation, (i) we minimize the cosine distance between a current task's parameters against prior parameters sequentially, not in parallel (i.e. the subspace end-points are sequentially fixed in the parameter space, we cannot dynamically move the subspace towards a different region), 
(ii) we do not gain visibility to all task parameters at once, 
and (iii) we are computing distance w.r.t. multi-task parameters (i.e. each task's parameters is applicable to its own and prior tasks).

\noindent\textbf{(Observation 4)}
\textit{
(Tables 2-3, Appendices A.3.3, A.3.8)
Constructing a CPS
can yield a subspace containing low-loss parameters for corresponding inputs. 
}
Based on Table 2, compared to a non-CPS,
we observe that more points interpolated in the input space can be mapped to low-loss interpolated parameters in the parameter space. 
We would expect a larger subspace to contain more low-loss parameters, yet counter-intuitively 
a subspace compressed w.r.t. opposing boundary parameters contains more mappable
low-loss parameters (Theorem 3).
We also note that the interpolated points perform better than reusing boundary parameters. 
This result is visualized in 
Appendix A.3.3
(and Figure \ref{fig:loss_landscape}): more points in the input space can be mapped with higher accuracy, and fewer, sparser yet more-accurate points can be located in the parameter space that can map back to these inputs. 
In addition to these interpolated parameters that can be mapped back to shifted distributions, the average (ensemble) accuracy of the parameter space is higher than $\beta=0.0$.
In-line with Observation 1, 
we find from Appendix A.3.8
that, compared to a clean test-time distribution, we retain a comparable accuracy (centre / ensemble / interpolation) under different types of test-time perturbations.

To study improvements to task adaptation and meta learner capacity, 
we evaluate an additional CPS-regularization term to \cite{vonoswald2020continual}'s hypernetwork
to find reduced catastrophic forgetting across tasks (Table 3).
Though lowest-loss parameters are sparser in the compressed subspace, 
the hypernetwork can map parameters to tasks within the bounds of the parameter end-points. Though this is counter-intuitive as we would expect distant tasks to be located in distant subspaces in the input space, by forcing the parameters to be co-located in a single (yet interpolatable) subspace, we reduce the propensity for the hypernetwork to compute distant ($\beta=0.0$) parameters with high-loss regions in between computed parameters.

\begin{table*}[t]
\centering
	\begin{minipage}[t]{0.42\textwidth}
	\centering
    \resizebox{\textwidth}{!}{
        \begin{tabular}{|lr|c|}
        \hline
        && \multicolumn{1}{p{2.5cm}|}{Uniform interpolation (of inputs) }
        \\ \hline \hline
        \multirow{4}{2cm}{CPS: $\beta = 0.0$ (independently-trained)} 
        & \textbf{boundary} & \multicolumn{1}{c|}{$47.2 \pm 14.1$}  \\
        & centre & \multicolumn{1}{c|}{$28.5 \pm 4.1$}  \\
        & ens. (mean) & \multicolumn{1}{c|}{$29.2 \pm 6.8$}  \\
        & intp. (max) & \multicolumn{1}{c|}{$38.2 \pm 7.8$}  \\
        \hline
        \multirow{3}{2cm}{CPS: 3 seen tasks of same coarse label, low-capacity} 
        & \textbf{boundary}& \multicolumn{1}{c|}{$49.4 \pm 10.8$}  \\ 
        & centre & \multicolumn{1}{c|}{$33.4 \pm 8.4$}  \\
        & ens. (mean) & \multicolumn{1}{c|}{$32.9 \pm 6.1$}  \\
        & intp. (max) & \multicolumn{1}{c|}{$53.4 \pm 8.5$}  \\
        \hline
        \end{tabular}
        }
    \label{tab:task_intp}
    \caption{
    \textit{CPS vs multiple test-time distributions: }
    Between
    $\beta=0.0$ and $\beta=1.0$,
    with respect to a set of $50^3$ interpolated tasks,
    we evaluate accuracy
    with respect to the lowest-loss boundary parameter, centre parameter, ensembling 1000 parameters and averaging their predictions, and searching for the lowest-loss interpolated parameter (excluding boundary parameters).
    }
	\end{minipage}
	\hfill
	\begin{minipage}[t]{0.53\textwidth}
	\centering
    \resizebox{\textwidth}{!}{
    \begin{tabular}{|p{3.5cm}|p{1.5cm}|p{1.5cm}|p{1.5cm}|p{1.5cm}|p{1.5cm}|p{1.5cm}|}
    \hline
    Test Accuracy & \multicolumn{1}{c|}{CIFAR10} & \multicolumn{1}{c|}{1} & \multicolumn{1}{c|}{2} & \multicolumn{1}{c|}{3} & \multicolumn{1}{c|}{4} & \multicolumn{1}{c|}{5} \\ 
    \hline 
    Trained from scratch & \multicolumn{1}{c|}{$67.7$} & \multicolumn{1}{c|}{$70.9$} & \multicolumn{1}{c|}{$67.1$} & \multicolumn{1}{c|}{$61.7$} & \multicolumn{1}{c|}{$69.4$} & \multicolumn{1}{c|}{$70.9$} \\
    \hline
    \multicolumn{7}{|c|}{Standard Hypernetwork \citep{vonoswald2020continual}} \\
    \hline
    Hypernetwork, evaluated after each task & \multicolumn{1}{c|}{$63.7$} & \multicolumn{1}{c|}{$63.2$} & \multicolumn{1}{c|}{$60.1$} & \multicolumn{1}{c|}{$62.4$} & \multicolumn{1}{c|}{$61.7$} & \multicolumn{1}{c|}{$62.4$} \\ \hline
    Hypernetwork, evaluated after task 5 & \multicolumn{1}{c|}{$61.8$} & \multicolumn{1}{c|}{$61.9$} & \multicolumn{1}{c|}{$58.9$} & \multicolumn{1}{c|}{$59.3$} & \multicolumn{1}{c|}{$59.2$} & \multicolumn{1}{c|}{$59.9$} \\ \hline
    \multicolumn{7}{|c|}{CPS-regularized Hypernetwork} \\
    \hline
    Hypernetwork, evaluated after each task & \multicolumn{1}{c|}{$65.6$} & \multicolumn{1}{c|}{$68.2$} & \multicolumn{1}{c|}{$64.4$} & \multicolumn{1}{c|}{$68.7$} & \multicolumn{1}{c|}{$64.8$} & \multicolumn{1}{c|}{$64.9$} \\ \hline
    Hypernetwork, evaluated after task 5 & \multicolumn{1}{c|}{$63.6$} & \multicolumn{1}{c|}{$65.0$} & \multicolumn{1}{c|}{$60.7$} & \multicolumn{1}{c|}{$62.9$} & \multicolumn{1}{c|}{$61.6$} & \multicolumn{1}{c|}{$61.7$} \\ \hline
    \end{tabular}
        }
    \label{tab:task_hnet}
    \caption{
    \textit{CPS vs multiple test-time distributions: }
    Test set accuracy in continual learnign setting on CIFAR10 and subsequent CIFAR100
    splits of 10 classes. 
    We evaluate \cite{vonoswald2020continual}'s hypernetwork and our CPS-regularized hypernetwork, and compare between training on each task individually, evaluating a task right after learning a new task, and evaluating a task after learning all tasks.
    }
	\end{minipage}
	\label{tab:capacity}
\end{table*}

\noindent\textbf{(Observation 5)}
\textit{
(Appendix A.3.8)
Subspace capacity supports 
unseen tasks of different fine labels and different coarse labels. 
}
From Appendix A.3.8,
in addition to test set performance of seen tasks, 
we find that training a CPS on seen tasks of constant coarse label set
can retain above-chance accuracy on unseen tasks of different fine labels as well as distinctly different coarse labels, supported by the high mean and low standard deviation.
Semantic features may transfer across tasks of identical coarse labels, which can explain the performance with respect to unseen tasks of shared coarse labels. The performance with respect to unseen tasks of unshared coarse labels
could be interpreted as the CPS containing large enough sets of (random) parameters that fit this task without training, 
in-line with \citet{https://doi.org/10.48550/arxiv.1911.13299}
where a randomly-initialized network contains low-loss subnetworks.
These phenomena indicate an extremely large and flexible capacity supported within the parameter subspace; it has an increased number of low-loss parameters for shifted and unseen test-time distributions of shared labels, but also contains random parameters that can arbitrarily fit random tasks.

\noindent\textbf{(Observation 6)}
\textit{
(Appendices A.3.4-8)
Sampled parameters
can also be used as multi-task solutions.
}
To evaluate how well parameter point-estimates can operate as multi-task solutions, 
we consider over-parameterizaton of the subspace with respect to (i) minimizing the number of train-time distributions (which, according to Theorem 1, would reduce capacity per point-estimate allocated to an opposing end-point), and (ii) increasing model capacity (depth and/or width).
Considering varying over-parameterization w.r.t. a single test-time distribution (Appendix A.3.4),
though a larger capacity model can retain higher centre/ensemble/interpolation accuracy than a smaller capacity model, 
accuracy w.r.t. the test set of train-time distributions do not vary significantly when
the number of train-time distributions varies (i.e. a trained subspace can expand or change capacity to support more parameter end-points accordingly). 
Similarly in Appendix A.3.7,
when we vary the number of non-shared fine
and
coarse label sets, 
though the centre/ensemble accuracy does not improve significantly, 
we find that the accuracy of the lowest-loss interpolatable parameter increases with an increase in model capacity.

In Appendix A.3.8,
unique-task solutions outperform multi-task solutions.
From Appendices A.3.5 and A.3.6,
we find that increasing model capacity and decreasing non-label-sharing 
tasks 
increases performance for multi-task learning: an over-parameterized subspace yields high accuracy when evaluated with the subspace centre or ensemble.
An increase in centre/ensemble accuracy indicates that a sampled point has a higher propensity of providing low-loss inference on a sampled input set.
As the number of non-label-shared tasks increases, the general performance decreases; increasing the capacity helps mitigate this, though it only increases the ability to locate a linearly-interpolated multi-task solution.
When a subspace is over-parameterized for multi-task learning, 
though accuracy is increased for centre / ensemble / interpolation,
it does not linearly minimize the gap between them.

\section{Related Work}
\label{sec:rw}

Recent work demonstrate the effects or usage of multiple distribution shift sources for weakening (attacks) or strengthening (defenses) model robustness.
\citet{qi2021mind} used text style transfer to perform adversarial attacks.
\citet{datta2021learnweight} introduces adversarial perturbations to a target domain to fool a model trained on a source domain.
\citet{naseer2019crossdomain} generated domain-invariant adversarial perturbations to fool models of different domains.
AdvTrojan \citep{liu2021synergetic} combined adversarial perturbations with backdoor trigger perturbations to craft stealthy triggers to perform backdoor attacks.
\citet{10.5555/2946645.2946704} proposed domain-adapted adversarial training to improve domain adaptation.
\citet{10.1007/978-3-030-58580-8_4} proposed a robustness measure by augmenting a dataset with both adversarial noise and stylized perturbations, by evaluating a set of perturbation types including Gaussian noise, stylization and adversarial perturbations.
\citet{santurkar2020breeds} synthesized distribution shifts by combining random noise, adversarial perturbations, and domain shifts to contribute subpopulation shift benchmarks.
\citet{datta2022backdoors} demonstrated the low likelihood of backdooring a model in the presence of joint distribution shifts, including multiple perturbations of the same shift type (backdoor), and multiple perturbations of different shift types (backdoor, adversarial, domain).

Given joint distribution shifts can manifest at test-time, corresponding wide-spectrum robustness techniques are required. 
Methods of robustness for a single test-time distribution (e.g. adversarial/backdoor attacks) do not strongly overlap with that for multiple test-time distributions (e.g. domain adaptation, meta/continual learning). 
The methods tend to be optimized for train-time-specified disjoint distribution shifts. 
For example, \citet{NEURIPS2020_8b406655} found a trade-off between adversarial defenses optimized towards adversarial perturbations against backdoor defenses optimized towards backdoor perturbations, where the adoption of one defense worsened robustness to the other's attack.
Though data augmentation \citep{yun2019cutmix, borgnia2020strong} and adversarial training \citep{goodfellow2015explaining, geiping2021doesnt} as a technique can mitigate adversarial attacks and backdoor attacks, they need to be trained on one of the perturbation types to manifest robustness. 
We would be amongst the first to propose an adaptation procedure that tackles multiple/joint distribution shifts.

\section{Conclusion}

We learn that sampling points in the parameter subspace can return lower-loss mapped parameters if the space was compressed during training, yield robust accuracy across various perturbation types, and reduces catastrophic forgetting when adapted into a hypernetwork. 
We motivate further study into enforcing geometric structure to parameter subspace construction and methods of test-time adaptation towards (joint) distribution shifts.

\newpage
\bibliography{main}

\begin{thebibliography}{52}
\expandafter\ifx\csname natexlab\endcsname\relax\def\natexlab#1{#1}\fi

\bibitem[{Arpit et~al.(2021)Arpit, Wang, Zhou, and Xiong}]{arpit2021ensemble}
Devansh Arpit, Huan Wang, Yingbo Zhou, and Caiming Xiong. 2021.
\newblock \href {http://arxiv.org/abs/2110.10832} {Ensemble of averages:
  Improving model selection and boosting performance in domain generalization}.

\bibitem[{Borgnia et~al.(2020)Borgnia, Cherepanova, Fowl, Ghiasi, Geiping,
  Goldblum, Goldstein, and Gupta}]{borgnia2020strong}
Eitan Borgnia, Valeriia Cherepanova, Liam Fowl, Amin Ghiasi, Jonas Geiping,
  Micah Goldblum, Tom Goldstein, and Arjun Gupta. 2020.
\newblock \href {http://arxiv.org/abs/2011.09527} {Strong data augmentation
  sanitizes poisoning and backdoor attacks without an accuracy tradeoff}.

\bibitem[{Cha et~al.(2021)Cha, Chun, Lee, Cho, Park, Lee, and
  Park}]{cha2021swad}
Junbum Cha, Sanghyuk Chun, Kyungjae Lee, Han-Cheol Cho, Seunghyun Park, Yunsung
  Lee, and Sungrae Park. 2021.
\newblock \href {http://arxiv.org/abs/2102.08604} {Swad: Domain generalization
  by seeking flat minima}.

\bibitem[{Cheung et~al.(2019)Cheung, Terekhov, Chen, Agrawal, and
  Olshausen}]{cheung2019superposition}
Brian Cheung, Alex Terekhov, Yubei Chen, Pulkit Agrawal, and Bruno Olshausen.
  2019.
\newblock \href {http://arxiv.org/abs/1902.05522} {Superposition of many models
  into one}.

\bibitem[{Datta(2021)}]{datta2021learnweight}
Siddhartha Datta. 2021.
\newblock \href {https://openreview.net/forum?id=1-j4VLSHApJ} {Learn2weight:
  Weights transfer defense against similar-domain adversarial attacks}.

\bibitem[{Datta and Shadbolt(2022)}]{datta2022backdoors}
Siddhartha Datta and Nigel Shadbolt. 2022.
\newblock \href {https://arxiv.org/abs/2201.12211} {Backdoors stuck at the
  frontdoor: Multi-agent backdoor attacks that backfire}.
\newblock In \emph{International Conference on Learning Representations
  Workshop: Gamification and Multiagent Solutions}.

\bibitem[{Draxler et~al.(2019)Draxler, Veschgini, Salmhofer, and
  Hamprecht}]{draxler2019essentially}
Felix Draxler, Kambis Veschgini, Manfred Salmhofer, and Fred~A. Hamprecht.
  2019.
\newblock \href {http://arxiv.org/abs/1803.00885} {Essentially no barriers in
  neural network energy landscape}.

\bibitem[{Finn et~al.(2017)Finn, Abbeel, and Levine}]{finn2017modelagnostic}
Chelsea Finn, Pieter Abbeel, and Sergey Levine. 2017.
\newblock \href {http://arxiv.org/abs/1703.03400} {Model-agnostic meta-learning
  for fast adaptation of deep networks}.

\bibitem[{Fort et~al.(2020)Fort, Hu, and Lakshminarayanan}]{fort2020deep}
Stanislav Fort, Huiyi Hu, and Balaji Lakshminarayanan. 2020.
\newblock \href {http://arxiv.org/abs/1912.02757} {Deep ensembles: A loss
  landscape perspective}.

\bibitem[{Fort and Jastrzebski(2019)}]{fort2019large}
Stanislav Fort and Stanislaw Jastrzebski. 2019.
\newblock \href {http://arxiv.org/abs/1906.04724} {Large scale structure of
  neural network loss landscapes}.

\bibitem[{Ganin et~al.(2016{\natexlab{a}})Ganin, Ustinova, Ajakan, Germain,
  Larochelle, Laviolette, Marchand, and Lempitsky}]{10.5555/2946645.2946704}
Yaroslav Ganin, Evgeniya Ustinova, Hana Ajakan, Pascal Germain, Hugo
  Larochelle, Fran\c{c}ois Laviolette, Mario Marchand, and Victor Lempitsky.
  2016{\natexlab{a}}.
\newblock Domain-adversarial training of neural networks.
\newblock \emph{J. Mach. Learn. Res.}, 17(1):2096–2030.

\bibitem[{Ganin et~al.(2016{\natexlab{b}})Ganin, Ustinova, Ajakan, Germain,
  Larochelle, Laviolette, March, and Lempitsky}]{JMLR:v17:15-239}
Yaroslav Ganin, Evgeniya Ustinova, Hana Ajakan, Pascal Germain, Hugo
  Larochelle, Fran{\c{c}}ois Laviolette, Mario March, and Victor Lempitsky.
  2016{\natexlab{b}}.
\newblock \href {http://jmlr.org/papers/v17/15-239.html} {Domain-adversarial
  training of neural networks}.
\newblock \emph{Journal of Machine Learning Research}, 17(59):1--35.

\bibitem[{Ganin et~al.(2016{\natexlab{c}})Ganin, Ustinova, Ajakan, Germain,
  Larochelle, Laviolette, Marchand, and Lempitsky}]{ganin2016domainadversarial}
Yaroslav Ganin, Evgeniya Ustinova, Hana Ajakan, Pascal Germain, Hugo
  Larochelle, François Laviolette, Mario Marchand, and Victor Lempitsky.
  2016{\natexlab{c}}.
\newblock \href {http://arxiv.org/abs/1505.07818} {Domain-adversarial training
  of neural networks}.

\bibitem[{Garipov et~al.(2018)Garipov, Izmailov, Podoprikhin, Vetrov, and
  Wilson}]{garipov2018loss}
Timur Garipov, Pavel Izmailov, Dmitrii Podoprikhin, Dmitry Vetrov, and
  Andrew~Gordon Wilson. 2018.
\newblock \href {http://arxiv.org/abs/1802.10026} {Loss surfaces, mode
  connectivity, and fast ensembling of dnns}.

\bibitem[{Geiping et~al.(2021)Geiping, Fowl, Somepalli, Goldblum, Moeller, and
  Goldstein}]{geiping2021doesnt}
Jonas Geiping, Liam Fowl, Gowthami Somepalli, Micah Goldblum, Michael Moeller,
  and Tom Goldstein. 2021.
\newblock \href {http://arxiv.org/abs/2102.13624} {What doesn't kill you makes
  you robust(er): Adversarial training against poisons and backdoors}.

\bibitem[{Geirhos et~al.(2019)Geirhos, Rubisch, Michaelis, Bethge, Wichmann,
  and Brendel}]{geirhos2018imagenettrained}
Robert Geirhos, Patricia Rubisch, Claudio Michaelis, Matthias Bethge, Felix~A.
  Wichmann, and Wieland Brendel. 2019.
\newblock \href {https://openreview.net/forum?id=Bygh9j09KX} {Imagenet-trained
  {CNN}s are biased towards texture; increasing shape bias improves accuracy
  and robustness.}
\newblock In \emph{International Conference on Learning Representations}.

\bibitem[{Goodfellow et~al.(2015)Goodfellow, Shlens, and
  Szegedy}]{goodfellow2015explaining}
Ian Goodfellow, Jonathon Shlens, and Christian Szegedy. 2015.
\newblock \href {http://arxiv.org/abs/1412.6572} {Explaining and harnessing
  adversarial examples}.
\newblock In \emph{International Conference on Learning Representations}.

\bibitem[{Gu et~al.(2019)Gu, Liu, Dolan-Gavitt, and Garg}]{8685687}
Tianyu Gu, Kang Liu, Brendan Dolan-Gavitt, and Siddharth Garg. 2019.
\newblock \href {https://doi.org/10.1109/ACCESS.2019.2909068} {Badnets:
  Evaluating backdooring attacks on deep neural networks}.
\newblock \emph{IEEE Access}, 7:47230--47244.

\bibitem[{Hoffman et~al.(2017)Hoffman, Tzeng, Park, Zhu, Isola, Saenko, Efros,
  and Darrell}]{hoffman2017cycada}
Judy Hoffman, Eric Tzeng, Taesung Park, Jun-Yan Zhu, Phillip Isola, Kate
  Saenko, Alexei~A. Efros, and Trevor Darrell. 2017.
\newblock \href {http://arxiv.org/abs/1711.03213} {Cycada: Cycle-consistent
  adversarial domain adaptation}.

\bibitem[{Hsu et~al.(2019)Hsu, Levine, and Finn}]{hsu2019unsupervised}
Kyle Hsu, Sergey Levine, and Chelsea Finn. 2019.
\newblock \href {http://arxiv.org/abs/1810.02334} {Unsupervised learning via
  meta-learning}.

\bibitem[{Huang et~al.(2018)Huang, Lin, Chen, Wu, Hsu, and
  Lai}]{Huang_2018_ECCV}
Sheng-Wei Huang, Che-Tsung Lin, Shu-Ping Chen, Yen-Yi Wu, Po-Hao Hsu, and
  Shang-Hong Lai. 2018.
\newblock Auggan: Cross domain adaptation with gan-based data augmentation.
\newblock In \emph{Proceedings of the European Conference on Computer Vision
  (ECCV)}.

\bibitem[{Huang and Belongie(2017)}]{huang2017adain}
Xun Huang and Serge Belongie. 2017.
\newblock Arbitrary style transfer in real-time with adaptive instance
  normalization.
\newblock In \emph{ICCV}.

\bibitem[{Ilyas et~al.(2022)Ilyas, Park, Engstrom, Leclerc, and
  Madry}]{https://doi.org/10.48550/arxiv.2202.00622}
Andrew Ilyas, Sung~Min Park, Logan Engstrom, Guillaume Leclerc, and Aleksander
  Madry. 2022.
\newblock \href {https://doi.org/10.48550/ARXIV.2202.00622} {Datamodels:
  Predicting predictions from training data}.

\bibitem[{Khodadadeh et~al.(2019)Khodadadeh, Bölöni, and
  Shah}]{khodadadeh2019unsupervised}
Siavash Khodadadeh, Ladislau Bölöni, and Mubarak Shah. 2019.
\newblock \href {http://arxiv.org/abs/1811.11819} {Unsupervised meta-learning
  for few-shot image classification}.

\bibitem[{{Krizhevsky}(2009)}]{krizhevsky2009learning}
Alex {Krizhevsky}. 2009.
\newblock Learning multiple layers of features from tiny images.

\bibitem[{Liu et~al.(2021)Liu, Khalil, Khreishah, and Phan}]{liu2021synergetic}
Guanxiong Liu, Issa Khalil, Abdallah Khreishah, and NhatHai Phan. 2021.
\newblock \href {http://arxiv.org/abs/2109.01275} {A synergetic attack against
  neural network classifiers combining backdoor and adversarial examples}.

\bibitem[{Madry et~al.(2018)Madry, Makelov, Schmidt, Tsipras, and
  Vladu}]{madry2018towards}
Aleksander Madry, Aleksandar Makelov, Ludwig Schmidt, Dimitris Tsipras, and
  Adrian Vladu. 2018.
\newblock \href {https://openreview.net/forum?id=rJzIBfZAb} {Towards deep
  learning models resistant to adversarial attacks}.
\newblock In \emph{International Conference on Learning Representations}.

\bibitem[{Mirzadeh et~al.(2020)Mirzadeh, Farajtabar, Gorur, Pascanu, and
  Ghasemzadeh}]{mirzadeh2020linear}
Seyed~Iman Mirzadeh, Mehrdad Farajtabar, Dilan Gorur, Razvan Pascanu, and
  Hassan Ghasemzadeh. 2020.
\newblock \href {http://arxiv.org/abs/2010.04495} {Linear mode connectivity in
  multitask and continual learning}.

\bibitem[{Naseer et~al.(2019)Naseer, Khan, Khan, Khan, and
  Porikli}]{naseer2019crossdomain}
Muzammal Naseer, Salman~H. Khan, Harris Khan, Fahad~Shahbaz Khan, and Fatih
  Porikli. 2019.
\newblock \href {http://arxiv.org/abs/1905.11736} {Cross-domain transferability
  of adversarial perturbations}.

\bibitem[{Neyshabur et~al.(2020)Neyshabur, Sedghi, and
  Zhang}]{NEURIPS2020_0607f4c7}
Behnam Neyshabur, Hanie Sedghi, and Chiyuan Zhang. 2020.
\newblock \href
  {https://proceedings.neurips.cc/paper/2020/file/0607f4c705595b911a4f3e7a127b44e0-Paper.pdf}
  {What is being transferred in transfer learning?}
\newblock In \emph{Advances in Neural Information Processing Systems},
  volume~33, pages 512--523. Curran Associates, Inc.

\bibitem[{Pan and Yang(2010)}]{5288526}
Sinno~Jialin Pan and Qiang Yang. 2010.
\newblock \href {https://doi.org/10.1109/TKDE.2009.191} {A survey on transfer
  learning}.
\newblock \emph{IEEE Transactions on Knowledge and Data Engineering},
  22(10):1345--1359.

\bibitem[{Peng et~al.(2019)Peng, Bai, Xia, Huang, Saenko, and
  Wang}]{peng2019moment}
Xingchao Peng, Qinxun Bai, Xide Xia, Zijun Huang, Kate Saenko, and Bo~Wang.
  2019.
\newblock \href {http://arxiv.org/abs/1812.01754} {Moment matching for
  multi-source domain adaptation}.

\bibitem[{Qi et~al.(2021)Qi, Chen, Zhang, Li, Liu, and Sun}]{qi2021mind}
Fanchao Qi, Yangyi Chen, Xurui Zhang, Mukai Li, Zhiyuan Liu, and Maosong Sun.
  2021.
\newblock \href {http://arxiv.org/abs/2110.07139} {Mind the style of text!
  adversarial and backdoor attacks based on text style transfer}.

\bibitem[{Ramanujan et~al.(2019)Ramanujan, Wortsman, Kembhavi, Farhadi, and
  Rastegari}]{https://doi.org/10.48550/arxiv.1911.13299}
Vivek Ramanujan, Mitchell Wortsman, Aniruddha Kembhavi, Ali Farhadi, and
  Mohammad Rastegari. 2019.
\newblock \href {https://doi.org/10.48550/ARXIV.1911.13299} {What's hidden in a
  randomly weighted neural network?}

\bibitem[{Ren et~al.(2019)Ren, Liu, Fertig, Snoek, Poplin, DePristo, Dillon,
  and Lakshminarayanan}]{10.5555/3454287.3455604}
Jie Ren, Peter~J. Liu, Emily Fertig, Jasper Snoek, Ryan Poplin, Mark~A.
  DePristo, Joshua~V. Dillon, and Balaji Lakshminarayanan. 2019.
\newblock \emph{Likelihood Ratios for Out-of-Distribution Detection}. Curran
  Associates Inc., Red Hook, NY, USA.

\bibitem[{Rusak et~al.(2020)Rusak, Schott, Zimmermann, Bitterwolf, Bringmann,
  Bethge, and Brendel}]{10.1007/978-3-030-58580-8_4}
Evgenia Rusak, Lukas Schott, Roland~S. Zimmermann, Julian Bitterwolf, Oliver
  Bringmann, Matthias Bethge, and Wieland Brendel. 2020.
\newblock A simple way to make neural networks robust against diverse image
  corruptions.
\newblock In \emph{Computer Vision -- ECCV 2020}, pages 53--69, Cham. Springer
  International Publishing.

\bibitem[{Santurkar et~al.(2020)Santurkar, Tsipras, and
  Madry}]{santurkar2020breeds}
Shibani Santurkar, Dimitris Tsipras, and Aleksander Madry. 2020.
\newblock \href {http://arxiv.org/abs/2008.04859} {Breeds: Benchmarks for
  subpopulation shift}.

\bibitem[{Sastry and Oore(2020)}]{pmlr-v119-sastry20a}
Chandramouli~Shama Sastry and Sageev Oore. 2020.
\newblock \href {https://proceedings.mlr.press/v119/sastry20a.html} {Detecting
  out-of-distribution examples with {G}ram matrices}.
\newblock In \emph{Proceedings of the 37th International Conference on Machine
  Learning}, volume 119 of \emph{Proceedings of Machine Learning Research},
  pages 8491--8501. PMLR.

\bibitem[{Snoek et~al.(2015)Snoek, Rippel, Swersky, Kiros, Satish, Sundaram,
  Patwary, Prabhat, and Adams}]{10.5555/3045118.3045349}
Jasper Snoek, Oren Rippel, Kevin Swersky, Ryan Kiros, Nadathur Satish,
  Narayanan Sundaram, Md. Mostofa~Ali Patwary, Prabhat Prabhat, and Ryan~P.
  Adams. 2015.
\newblock Scalable bayesian optimization using deep neural networks.
\newblock In \emph{Proceedings of the 32nd International Conference on
  International Conference on Machine Learning - Volume 37}, ICML'15, page
  2171–2180. JMLR.org.

\bibitem[{Sun et~al.(2015)Sun, Feng, and Saenko}]{sun2015return}
Baochen Sun, Jiashi Feng, and Kate Saenko. 2015.
\newblock \href {http://arxiv.org/abs/1511.05547} {Return of frustratingly easy
  domain adaptation}.

\bibitem[{von Oswald et~al.(2020)von Oswald, Henning, Sacramento, and
  Grewe}]{vonoswald2020continual}
Johannes von Oswald, Christian Henning, João Sacramento, and Benjamin~F.
  Grewe. 2020.
\newblock \href {http://arxiv.org/abs/1906.00695} {Continual learning with
  hypernetworks}.

\bibitem[{Weng et~al.(2020)Weng, Lee, and Wu}]{NEURIPS2020_8b406655}
Cheng-Hsin Weng, Yan-Ting Lee, and Shan-Hung~(Brandon) Wu. 2020.
\newblock \href
  {https://proceedings.neurips.cc/paper/2020/file/8b4066554730ddfaa0266346bdc1b202-Paper.pdf}
  {On the trade-off between adversarial and backdoor robustness}.
\newblock In \emph{Advances in Neural Information Processing Systems},
  volume~33, pages 11973--11983. Curran Associates, Inc.

\bibitem[{Wortsman et~al.(2021)Wortsman, Horton, Guestrin, Farhadi, and
  Rastegari}]{wortsman2021learning}
Mitchell Wortsman, Maxwell Horton, Carlos Guestrin, Ali Farhadi, and Mohammad
  Rastegari. 2021.
\newblock \href {http://arxiv.org/abs/2102.10472} {Learning neural network
  subspaces}.

\bibitem[{Yao et~al.(2021)Yao, Zhang, and Finn}]{yao2021metalearning}
Huaxiu Yao, Linjun Zhang, and Chelsea Finn. 2021.
\newblock \href {http://arxiv.org/abs/2106.02695} {Meta-learning with fewer
  tasks through task interpolation}.

\bibitem[{Yun et~al.(2019{\natexlab{a}})Yun, Han, Oh, Chun, Choe, and
  Yoo}]{Yun_2019_ICCV}
Sangdoo Yun, Dongyoon Han, Seong~Joon Oh, Sanghyuk Chun, Junsuk Choe, and
  Youngjoon Yoo. 2019{\natexlab{a}}.
\newblock Cutmix: Regularization strategy to train strong classifiers with
  localizable features.
\newblock In \emph{Proceedings of the IEEE/CVF International Conference on
  Computer Vision (ICCV)}.

\bibitem[{Yun et~al.(2019{\natexlab{b}})Yun, Han, Oh, Chun, Choe, and
  Yoo}]{yun2019cutmix}
Sangdoo Yun, Dongyoon Han, Seong~Joon Oh, Sanghyuk Chun, Junsuk Choe, and
  Youngjoon Yoo. 2019{\natexlab{b}}.
\newblock \href {http://arxiv.org/abs/1905.04899} {Cutmix: Regularization
  strategy to train strong classifiers with localizable features}.

\bibitem[{Zeng et~al.(2020)Zeng, Qiu, Memmi, and Qiu}]{zeng2020data}
Yi~Zeng, Han Qiu, Gerard Memmi, and Meikang Qiu. 2020.
\newblock \href {http://arxiv.org/abs/2007.15290} {A data augmentation-based
  defense method against adversarial attacks in neural networks}.

\bibitem[{Zenke et~al.(2017)Zenke, Poole, and
  Ganguli}]{10.5555/3305890.3306093}
Friedemann Zenke, Ben Poole, and Surya Ganguli. 2017.
\newblock Continual learning through synaptic intelligence.
\newblock In \emph{Proceedings of the 34th International Conference on Machine
  Learning - Volume 70}, ICML'17, page 3987–3995. JMLR.org.

\bibitem[{Zhang et~al.(2017)Zhang, Bengio, Hardt, Recht, and
  Vinyals}]{zhang2017understanding}
Chiyuan Zhang, Samy Bengio, Moritz Hardt, Benjamin Recht, and Oriol Vinyals.
  2017.
\newblock \href {http://arxiv.org/abs/1611.03530} {Understanding deep learning
  requires rethinking generalization}.

\bibitem[{Zhang et~al.(2020)Zhang, Li, Guo, and Guo}]{zhang2020hybrid}
Hongjie Zhang, Ang Li, Jie Guo, and Yanwen Guo. 2020.
\newblock \href {http://arxiv.org/abs/2003.12506} {Hybrid models for open set
  recognition}.

\bibitem[{Zhang et~al.(2018)Zhang, Cisse, Dauphin, and
  Lopez-Paz}]{zhang2018mixup}
Hongyi Zhang, Moustapha Cisse, Yann~N. Dauphin, and David Lopez-Paz. 2018.
\newblock \href {http://arxiv.org/abs/1710.09412} {mixup: Beyond empirical risk
  minimization}.

\bibitem[{Zhou et~al.(2021)Zhou, Gu, Pang, Feng, Cheng, Lu, Shi, and
  Ma}]{zhou2021selfadversarial}
Qianyu Zhou, Qiqi Gu, Jiangmiao Pang, Zhengyang Feng, Guangliang Cheng, Xuequan
  Lu, Jianping Shi, and Lizhuang Ma. 2021.
\newblock \href {http://arxiv.org/abs/2108.03553} {Self-adversarial
  disentangling for specific domain adaptation}.

\end{thebibliography}
\bibliographystyle{acl_natbib}

\newpage
\appendix
\section{Appendix}

\subsection{Supporting Theory}

\noindent\textbf{Definition 1. }
\textit{
A \textbf{distribution shift} is the divergence between a train-time distribution $\mathbf{x_0}, \mathbf{y_0}$ and a test-time distribution $\hat{\mathbf{x}}, \hat{\mathbf{y}}$, where
$\hat{\mathbf{x}}, \hat{\mathbf{y}}$ is an interpolated distribution between $\mathbf{x_0}, \mathbf{y_0}$ and a target distribution $\mathbf{x_i^{\Delta}}, \mathbf{y_i^{\Delta}}$ such that
$\hat{\mathbf{x}} = \sum_{i}^{N} \alpha_i \mathbf{x_i^{\Delta}}$
and $\hat{\mathbf{y}} = \mathbf{y_i} | i := \arg\max_{i} \alpha_i$.
A \textbf{disjoint distribution shift} is a distribution shift of one target distribution $|\{ \alpha_i \}| = 2$ s.t. $\hat{\mathbf{x}} = \alpha \mathbf{x^{\Delta}} + (1-\alpha) \mathbf{x_0}$.
A \textbf{joint distribution shift} is a distribution shift of multiple target distributions $|\{ \alpha_i \}| > 2$ s.t. $\hat{\mathbf{x}} = \sum_{i}^{N} \alpha_i \mathbf{x_i^{\Delta}}$.
}

Let $\mathcal{X}$, $\mathcal{Y}$, $\mathcal{K}$, $\Theta$ be denoted as the input space, coarse label space, fine label space, and parameter space respectively, such that $\mathcal{X} \mapsto \mathcal{Y}$, $\mathcal{X} \mapsto \mathcal{K}$, $\mathcal{K} \mapsto \mathcal{Y}$, $\mathcal{X} \mapsto \Theta$.
Coarse labels are the super-class / higher-order labels of fine labels.
$\mathscr{f}$ is a base learner function
that accepts inputs $\mathbf{x}$, $\theta$ to return predicted labels $\bar{\mathbf{y}} = \mathscr{f}(\theta; \mathbf{x})$. 
The model parameters $\theta \sim \Theta$ are sampled from the parameter space, for example with an optimization procedure e.g. SGD, such that it minimizes the loss between the ground-truth and predicted labels: 
$\mathcal{L}(\theta; \mathbf{x}, \mathbf{y}) = \frac{1}{|\mathbf{x}|} \sum_{i}^{|\mathbf{x}|} (\mathscr{f}(\theta; \mathbf{x})-\mathbf{y})^2$.

In-line with notation in \citet{5288526},
for a marginal (or feature) distribution $\mathcal{P}(\mathcal{X})$ and a conditional (or label) distribution $\mathcal{P}(\mathcal{Y} | \mathcal{X})$,
a joint distribution shift is denoted as a joint transformation in the marginal-conditional distributions.
A covariate shift manifests as a change in the marginal distribution $\mathcal{P}(\mathcal{X})$ under constant conditional distribution $\mathcal{P}(\mathcal{Y} | \mathcal{X})$.
A label shift manifests as a change in the conditional distribution $\mathcal{P}(\mathcal{Y} | \mathcal{X})$ under constant marginal distribution $\mathcal{P}(\mathcal{X})$.
As elaborated in Section \ref{sec:rw}, deep neural networks may not perform well under this joint shift setting, and unifying robustness methods are limited.
To simplify our language in discussion, we use a set of points sampled in a space to pertain to a distribution.

To retain generality in notation and implementation, we re-iterate the aforementioned formalism in terms of sampled points in their respective spaces.
We sample mapped points in the input-label spaces as $\mathbf{x}, \mathbf{y} = \{ \{x, y\}|\texttt{condition} \sim \mathcal{X}, \mathcal{Y} \}$.
We describe shift here with respect to a single target label in order to highlight changes in the input-label mappings; the notation extends to non-singular label sets. 
Suppose we sample a clean / source / train-time set of points, constrained by a mapped target label $y^{*}$: $\mathbf{x_0}, \mathbf{y_0} = \{ \{x, y\}|y=y^{*} \sim \mathcal{X}, \mathcal{Y} \}$.

A distribution shift is the divergence between a train-time distribution $\mathbf{x_0}, \mathbf{y_0}$ and a test-time distribution $\hat{\mathbf{x}}, \hat{\mathbf{y}}$, where
$\hat{\mathbf{x}}, \hat{\mathbf{y}}$ is an interpolated distribution between $\mathbf{x_0}, \mathbf{y_0}$ and a target distribution $\mathbf{x_i^{\Delta}}, \mathbf{y_i^{\Delta}}$ such that
$\hat{\mathbf{x}} = \sum_{i}^{N} \alpha_i \mathbf{x_i^{\Delta}}$
and $\hat{\mathbf{y}} = \mathbf{y_i} | i := \arg\max_{i} \alpha_i$.
For a set of distributions $\{ \mathbf{x_0} \mapsto \mathbf{y_0},  \mathbf{x_1^{\Delta}} \mapsto \mathbf{y_1^{\Delta}}, ..., \mathbf{x_N^{\Delta}} \mapsto \mathbf{y_N^{\Delta}}\}$ containing $(N-1)$ target distributions, we sample interpolation coefficients $\alpha_i \sim [0, 1]$  s.t. $\alpha_i \mapsto \mathbf{x_i}$ and $\sum_{i}^{N} \alpha_i \leq N$. 
$N$ is the number of distributions being used; in CPS training, it is the number of train-time distributions used in training the subspace; in task interpolation, it is the number of train/test-time distributions used in interpolating between task sets.
A disjoint distribution shift is a distribution shift of only one target distribution $|\{ \alpha_i \}| = 2$ such that $\hat{\mathbf{x}} = \alpha \mathbf{x^{\Delta}} + (1-\alpha) \mathbf{x_0}$.
A joint distribution shift is a distribution shift of multiple target distributions $|\{ \alpha_i \}| > 2$ such that $\hat{\mathbf{x}} = \sum_{i}^{N} \alpha_i \mathbf{x_i^{\Delta}}$.

Each distinct target distribution is varied by the \textit{type} of perturbation (e.g. label, domain, task) and the \textit{variation} per type (e.g. multiple domains, multiple backdoor triggers).
For a clean test-time distribution where $|| \mathbf{x_1^{\Delta}} -\mathbf{x_0} ||_p \rightarrow 0$ and $y = y^{*}$, no shift takes place.
If $\hat{\mathbf{x}}, \hat{\mathbf{y}}$ manifests label shift ($y \neq y^{*}$, $\min || \mathbf{x_1^{\Delta}} -\mathbf{x_0} ||_p$), then $\mathbf{x_1^{\Delta}}, \mathbf{y_1^{\Delta}} = \{ \{x, y\}|y \neq y^{*} \sim \mathcal{X}, \mathcal{Y} \}$.
If $\hat{\mathbf{x}}, \hat{\mathbf{y}}$ manifests covariate shift ($|| \mathbf{x_2^{\Delta}} -\mathbf{x_0} ||_p > 0$), such as domain shift, then $\mathbf{x_2^{\Delta}}, \mathbf{y_2^{\Delta}} = \{ \{x, y\}|y=y_{\textnormal{ambg}} \sim \mathcal{X}, \mathcal{Y} \}$, where the sampled $x$ may not correspond to a clearly-defined class (i.e. a random / ambiguous class $y_{\textnormal{ambg}}$).
If $\hat{\mathbf{x}}, \hat{\mathbf{y}}$ manifests task shift (unrestricted $|| \mathbf{x_3^{\Delta}} -\mathbf{x_0} ||_p > 0$);
for label-shared task shift (identical coarse labels $y$, non-identical fine labels $k$) $\mathbf{x_3^{\Delta}}, \mathbf{y_3^{\Delta}} = \{ \{x, y\}|y=y^{*}, k \neq k^{*} \sim \mathcal{X}, \mathcal{Y}, \mathcal{K} \}$;
for non-label-shared task shift (non-identical coarse nor fine labels) $\mathbf{x_3^{\Delta}}, \mathbf{y_3^{\Delta}} = \{ \{x, y\}|y \neq y^{*}, k \neq k^{*} \sim \mathcal{X}, \mathcal{Y}, \mathcal{K} \}$.

For interpolated input sets between $\mathbf{x_0}$ and $\mathbf{x_i^{\Delta}}$, we would expect $0 < \alpha_0, \alpha_i < 1$, unless we expect the shifted distribution to completely manifest the target distribution (e.g. rotated sets, end-point task sets) such that $\alpha_i = 1$. 
The labels of the interpolated set is the label set with the larger corresponding $\alpha_i$.
In label shift, the perturbations are small enough that we retain the ground-truth labels.
For label-shared task shift, the ground-truth (coarse) labels between the end-point tasks are identical.
For non-label-shared task shift, we do not interpolate between non-identical label sets between the end-point tasks.

If we randomly-interpolate random input sets with random or ambiguous label assignments, 
we incur the risk of 
(i) the perturbed input set as being as being perceptually inconsistent with the assigned label, and 
(ii) the model overfitting towards random label assignments \citep{zhang2017understanding}.
To mitigate these concerns and 
generate semantically-valid (perceptually-interpreted as belonging to its assigned label) perturbations,
we adopt existing perturbation generation methods that have also been human-evaluated as perceptually consistent.
The use of these methods can be interpreted as the filtering of invalid points in the input space for the evaluation of mapping between $\mathcal{X}$ and $\Theta$.

\noindent\textbf{Corollary 1. }
\textit{
Test-time distributions can be decomposed into a weighted sum between known train-time distributions and computable interpolation coefficients.
}

We assume a given test-time distribution $\hat{\mathbf{x}}$ can be decomposed into a sum of parts $\hat{\mathbf{x}} := \sum_i^N \alpha_i \mathbf{x_i}$, where $N$ is the number of component distributions.
For the component distributions $\{ \mathbf{x_i} \}^N$, 
if a subset of the distributions exist in the set of train-time distributions, then the interpolation coefficient value per train-time distribution lies in $0 < \alpha_i \leq 1$;
if none of the distributions exist in the set of train-time distributions, then the interpolation coefficient value per train-time distribution is 0.

\noindent\textbf{Corollary 2. }
\textit{
For an input interpolated between a set of input end-points with interpolation coefficients $\{ \alpha_i \}^N$, 
the loss term per epoch can be decomposed into component/end-point loss terms of equivalent interpolation coefficients $\{ \alpha_i \}^N$, 
but the parameters may or may not be decomposed into component/end-point parameters of $\{ \alpha_i \}^N$. 
}

The loss of an interpolated set with respect to a parameter updated till epoch $e$ can be computed with respect to the weighted sum of the loss with respect to the boundary input points and interpolation coefficients for the interpolated input set. 

\begin{equation}
\small
\begin{split}
\frac{\partial \mathcal{L}(\theta_{t=e}; \hat{\mathbf{x}})}{\partial \theta_{t=e}} 
& = \frac{(\theta_{t=e} \sum_i^N \alpha_i \mathbf{x_i} - \mathbf{y})-(\theta_{t=e-1} \sum_i^N \alpha_i \mathbf{x_i} - \mathbf{y})}{\theta_{t=e} - \theta_{t=e-1}} \\
& = \sum_i^N \frac{(\theta_{t=e} \alpha_i \mathbf{x_i} - \mathbf{y})-(\theta_{t=e-1} \alpha_i \mathbf{x_i} - \mathbf{y})}{\theta_{t=e} - \theta_{t=e-1}} \\
&= \sum_i^N \alpha_i \frac{\partial \mathcal{L}(\theta_{t=e}; \mathbf{x_i})}{\partial \theta_{t=e}}
\end{split}
\end{equation}

$\theta_{t=e}$ is computed with respect to the total loss at epoch $e-1$.
$\theta_{t=e-1}$ updated with respect to multiple $\mathbf{x_i}$ may not be identical to $\theta_{t=e-1}$ updated with respect to a single set $\mathbf{x_0}$.
As a result, we cannot claim that an interpolated input set must correspond to or be satisfied with a parameter composed of the weighted sum of the interpolation coefficients and end-point parameters $\theta^{*} \neq \sum_i^N \alpha_i \theta_i$.

\textbf{Corollary 3. }
\textit{
Training and inference with a single parameter point-estimate $\theta$ may face a trade-off in loss w.r.t. $X_i$ and $X_n$ if interference exists between $X_i$ and $X_n$.
}

At a given epoch $e$, if $\frac{\partial \mathcal{L}(\theta; X_i)}{\partial \theta} \cdot \frac{\partial \mathcal{L}(\theta; X_n)}{\partial \theta} < 0$, then the decrease in loss w.r.t. one subset (either $X_i$ or $X_n$) would result in the increase in loss of the other during training, and consequently the same during inference. 

\noindent\textbf{Theorem 1. }
\textit{
For $N+1$ parameters trained in parallel,
any individual parameter $\theta_i$ is a function of its mapped data $X_i \mapsto Y_i$ as well as the remaining parameters $\{ \theta_n \}^{n \in N}$ (and by extension, $X_n \mapsto Y_n$).
Hence, a linearly-interpolated parameter between $\theta_i$ and $\{ \theta_n \}^{n \in N}$ is a function of $\{ \theta_n \}^{n \in N}$ and $\theta_i | \{ \theta_n \}^{n \in N}$.
}

\noindent\textbf{Proof sketch of Theorem 1. }
Suppose $N+1$ parameters are trained in parallel, and the number of dimensions of each parameter is $M = |\theta|$. 
To simplify the analysis, 
we use the Euclidean norm distance, 
and make relevant approximations. 

First, we decompose the loss w.r.t. $\theta_i$, in order to obtain $\theta_i$ in terms of $X_i$, $Y_i$, and $\{ \theta_n \}^{n \in N}$ such that $\mathcal{L}(\theta_i) = 0$.
We denote $\mathcal{C} = \sum_n^N \theta_n$.
By the Cauchy-Schwarz ineuality, we apply the following term replacement: $ (\sum_n^N \theta_n)^2 = (\sum_n^N 1 \cdot \theta_n)^2 \leq N \cdot \sum_n^N \theta_n^2 \Rightarrow \frac{\mathcal{C}}{N} \leq \sum_n^N \theta_n^2$.

\begin{equation}
\small
\begin{split}
\mathcal{L}(\theta_i) &= \mathcal{L}(\theta_i, X_i, Y_i) + \texttt{dist}(\theta_i, \{ \theta_n \}^{n \in N}) \\
&= (\theta_i X_i - Y_i) + \sum_n^N \sum_m^M (\theta_{i,m}-\theta_{n,m})^2 \\
& \approx (\theta_i X_i - Y_i) + \sum_n^N (\theta_i^2 - 2 \theta_i \theta_n + \theta_n^2) \\
& \approx (\theta_i X_i - Y_i) + N \theta_i^2 - 2 \theta_i \mathcal{C} + \frac{\mathcal{C}^2}{N} \\
& \approx N \theta_i^2 + (X_i - 2 \mathcal{C}) \theta_i + (\frac{\mathcal{C}^2}{N}-Y_i) \\
\theta_i &:= \frac{2\mathcal{C}-X_i \pm \sqrt{(X_i - 2 \mathcal{C})^2 - 4N(\frac{\mathcal{C}^2}{N}-Y_i)}}{2N} \phantom{=} \textnormal{s.t.} \phantom{=} \mathcal{L}(\theta_i) = 0 \\
\theta_i &:= \frac{2\mathcal{C}-X_i \pm \sqrt{-4 X_i \mathcal{C} + 4N Y_i + X_i^2}}{2N} \\
\end{split}
\label{equation:thm1_a}
\end{equation}

Next, we compute the change in $\theta_i$ w.r.t. change in $\mathcal{C}$, and find that the $\mathcal{C}$ term is persistent (non-eliminated) and is retained in the gradient calculation and optimization procedure of $\theta_i$.

\begin{equation}
\small
\begin{split}
\frac{\partial \theta_i}{\partial \mathcal{C}} 
&= \frac{1}{2N} \Bigg[ 2 \pm \frac{-4 X_i}{2\sqrt{-4 X_i \mathcal{C} + 4N Y_i + X_i^2}} \Bigg] \\
&= \frac{1}{N} \pm \frac{-2 X_i}{N\sqrt{-4 X_i \mathcal{C} + 4N Y_i + X_i^2}} \\
\end{split}
\label{equation:thm1_b}
\end{equation}

In order for Eqt \ref{equation:thm1_b} to remain valid, Eqt \ref{equation:thm1_c} needs to be satisfied, which also shows that $\mathcal{C}$ is dependent or contains artifacts from $X_i \mapsto Y_i$.

\begin{equation}
\small
\begin{split}
-4 X_i \mathcal{C} + 4N Y_i + X_i^2 \geq 0 \\
\mathcal{C} \leq \frac{4 X_i}{4N Y_i + X_i^2} \\
\end{split}
\label{equation:thm1_c}
\end{equation}

Further, suppose we linearly-interpolate a parameter $\theta$ within the bounded space of $\theta_i \cup \{ \theta_n \}^{n \in N}$, 
and the weighting per parameter is $\alpha_n \mapsto \theta_n$ such that $\sum_n^{N+1} \alpha_n \leq N+1$.
We find that the interpolated parameter $\theta$ has a non-eliminated term $\mathcal{C}$ regardless of set $\alpha$ or position in the bounded parameter space. 

\begin{equation}
\small
\begin{split}
\theta &= \alpha_i \theta_i + \sum_{n, n \neq i}^N \alpha_n \theta_n \\
\frac{\partial \theta}{\partial \mathcal{C}} & \approx \alpha_i \frac{\partial \theta_i}{\partial \mathcal{C}} + \frac{1}{N} \sum_{n, n \neq i}^N \alpha_n \\
\end{split}
\label{equation:thm1_d}
\end{equation}

\qed

\noindent\textbf{Theorem 2. }
\textit{
For $N+1$ parameters trained in parallel,
for an individual parameter $\theta_i$ and another parameter $\theta_n$ where $n \in N$,
the change in loss per iteration of SGD with respect to each parameter $\frac{\partial \mathcal{L}(\theta_i; X_i)}{\partial \theta_i}$, $\frac{\partial \mathcal{L}(\theta_n; X_n)}{\partial \theta_n}$ will both be in the same direction (i.e. identical signs), and will tend towards a local region of the parameter space upon convergence of loss.
}

\noindent\textbf{Proof sketch of Theorem 2. }
Suppose $N+1$ parameters are trained in parallel, and the number of dimensions of each parameter is $M = |\theta|$. 
We evaluate the incremental loss per iteration $\forall e \in E$, which is the distance regularization term.

\begin{equation}
\small
\begin{split}
\hspace{-1.75cm}
\Delta \mathcal{L}(\theta_i)_e & = \sum_n^N \sum_m^M (\theta_{i,m}-\theta_{n,m})^2 \\ 
& \approx \sum_n^N (\theta_{i}-\theta_{n})^2 \\
& := \sum_n^N \bigg[ \bigg(\theta_i - \frac{\partial \mathcal{L}(\theta_i; X_i)}{\partial \theta_i}\bigg) - \bigg(\theta_n - \frac{\partial \mathcal{L}(\theta_n; X_n)}{\partial \theta_n}\bigg)\bigg]^2 \\
& = \sum_n^N \bigg[ (\theta_i - \theta_n) + \bigg( \frac{\partial \mathcal{L}(\theta_n; X_n)}{\partial \theta_n} - \frac{\partial \mathcal{L}(\theta_i; X_i)}{\partial \theta_i} \bigg) \bigg]^2 \\
& = \sum_n^N \bigg[ (\theta_i - \theta_n)^2 + 2(\theta_i - \theta_n)\bigg( \frac{\partial \mathcal{L}(\theta_n; X_n)}{\partial \theta_n} - \frac{\partial \mathcal{L}(\theta_i; X_i)}{\partial \theta_i}\bigg) + \bigg( \frac{\partial \mathcal{L}(\theta_n; X_n)}{\partial \theta_n} - \frac{\partial \mathcal{L}(\theta_i; X_i)}{\partial \theta_i}\bigg)^2 \bigg]
\end{split}
\end{equation}

Given that minimizing the last term can result in minimizing the loss $\min \Delta \mathcal{L}(\theta_i)_e := \min (\theta_n - \frac{\partial \mathcal{L}(\theta_n; X_n)}{\partial \theta_n} - \frac{\partial \mathcal{L}(\theta_i; X_i)}{\partial \theta_i})$,
we next show that the gradient update (change in loss per iteration of SGD with respect to each parameter) terms at iteration $e$ must be positive and share the same direction when loss converges to 0.
For $\Delta \mathcal{L}(\theta_i)_e \geq 0$ and $\forall n$ where $n \in N$,

\begin{equation}
\small
\begin{split}
\hspace{-2.25cm}
\bigg(\frac{\partial \mathcal{L}(\theta_n; X_n)}{\partial \theta_n}\bigg)^2 + \bigg(\frac{\partial \mathcal{L}(\theta_i; X_i)}{\partial \theta_i}\bigg)^2 - 2\bigg(\frac{\partial \mathcal{L}(\theta_i; X_i)}{\partial \theta_i}\bigg)\bigg(\frac{\partial \mathcal{L}(\theta_n; X_n)}{\partial \theta_n}\bigg) & \geq 0 \\
\frac{\partial \mathcal{L}(\theta_i; X_i)}{\partial \theta_i} \cdot \frac{\partial \mathcal{L}(\theta_n; X_n)}{\partial \theta_n} & \leq \frac{1}{2}\bigg[\bigg(\frac{\partial \mathcal{L}(\theta_n; X_n)}{\partial \theta_n}\bigg)^2 + \bigg(\frac{\partial \mathcal{L}(\theta_i; X_i)}{\partial \theta_i}\bigg)^2\bigg] \\
\end{split}
\end{equation}

For $\Delta \mathcal{L}(\theta_i)_e \equiv 0$,

\begin{equation}
\small
\begin{split}
\frac{\partial \mathcal{L}(\theta_i; X_i)}{\partial \theta_i} \cdot \frac{\partial \mathcal{L}(\theta_n; X_n)}{\partial \theta_n} & \equiv \frac{1}{2}\bigg[\bigg(\frac{\partial \mathcal{L}(\theta_n; X_n)}{\partial \theta_n}\bigg)^2 + \bigg(\frac{\partial \mathcal{L}(\theta_i; X_i)}{\partial \theta_i}\bigg)^2\bigg] \geq 0\\
\end{split}
\end{equation}

Hence, 
$\frac{\partial \mathcal{L}(\theta_i; X_i)}{\partial \theta_i} \cdot \frac{\partial \mathcal{L}(\theta_n; X_n)}{\partial \theta_n}$ 
must be positive, and the component terms must be in the same direction (identical signs).

To further simplify analysis, we evaluate loss updates from iteration $e=0$ to $e=E-1$ where loss converges at iteration $E$, thus the component loss terms are non-zero. 
The two loss terms are thus non-orthogonal, co-directional, and the angle between them is less than $90^{\circ}$.

\begin{equation}
\small
\begin{split}
\frac{\partial \mathcal{L}(\theta_i; X_i)}{\partial \theta_i} \cdot \frac{\partial \mathcal{L}(\theta_n; X_n)}{\partial \theta_n} & > 0\\
\end{split}
\end{equation}

Given that $\theta_i$ and $\theta_n$ are initialized at the same constant random initialization $\theta_{i, e=0} \equiv \theta_{n, e=0}$, the change in each $\theta$ from iterations $e=0$ to $e=E-1$ is attributed to the difference in the summation of the gradient updates. 

\begin{numcases}{}
\begin{split}
\theta_i := \theta_{i, t=0} - \sum_{e}^{E-1} \frac{\partial \mathcal{L}(\theta_{i, t=e}; X_i)}{\partial \theta_{i, t=e}}
\end{split}
\\
\begin{split}
\theta_n := \theta_{n, t=0} - \sum_{e}^{E-1} \frac{\partial \mathcal{L}(\theta_{n, t=e}; X_n)}{\partial \theta_{n, t=e}}
\end{split}
\end{numcases}

We show next, that through enforcing distance regularization, we can exert a tendency for the distance between the summation of the loss term components to be smaller if they share the same direction, than if they were orthogonal or opposing directions. 

\begin{equation}
\small
\begin{split}
& \Bigg|\bigg[\bigg|\sum_{e}^{E-1} \frac{\partial \mathcal{L}(\theta_{i, t=e}; X_i)}{\partial \theta_{i, t=e}}\bigg|-\bigg|\sum_{e}^{E-1} \frac{\partial \mathcal{L}(\theta_{n, t=e}; X_n)}{\partial \theta_{n, t=e}}\bigg|\bigg]\bigg|_{\frac{\partial \mathcal{L}(\theta_i; X_i)}{\partial \theta_i} \cdot \frac{\partial \mathcal{L}(\theta_n; X_n)}{\partial \theta_n} > 0} \Bigg| \\
< & \Bigg|\bigg[\bigg|\sum_{e}^{E-1} \frac{\partial \mathcal{L}(\theta_{i, t=e}; X_i)}{\partial \theta_{i, t=e}}\bigg|-\bigg|\sum_{e}^{E-1} \frac{\partial \mathcal{L}(\theta_{n, t=e}; X_n)}{\partial \theta_{n, t=e}}\bigg|\bigg]\bigg|_{\frac{\partial \mathcal{L}(\theta_i; X_i)}{\partial \theta_i} \cdot \frac{\partial \mathcal{L}(\theta_n; X_n)}{\partial \theta_n} \leq 0} \Bigg|
\end{split}
\end{equation}

Thus we show that $\theta_i$ and $\theta_n$ are closer in distance in the parameter space. 
This can result in $\theta_i$ and $\theta_n$ sharing sub-matrices (intrinsic dimensions), and they may reside in the same or neighboring subspace. 

\qed

\noindent\textbf{Theorem 3. }
\textit{
For $N+1$ parameters trained in parallel,
the expected distance between the optimal parameter of
an interpolated input $\hat{\mathbf{x}} \mapsto \hat{\theta}$ 
would be smaller with respect to 
sampled points in a 
distance-regularized subspace
than 
non-distance-regularized subspace.
}

Informally, we conclude from Theorems 1 and 2 that, though non-linear and/or linear point-estimates may exist in a non-compressed parameter subspace, a compressed parameter subspace contains less interference attributed to changes in other dimensions; in other words, the CPS method reduces the dimensions of interpolation.
Practically,
this can manifest as most values in the weights matrix being relatively frozen or constant (and only few values or a submatrix being changed), or the range per value in a weights matrix may be narrower.

\noindent\textbf{Proof sketch of Theorem 3. }
Suppose we sample an interpolated input $\hat{\mathbf{x}} = \alpha_i \mathbf{x_i} + \alpha_n \mathbf{x_n}$. 
All trained parameters begin with a constant and equal initialization at $t=0$.

We would like to show here that any randomly interpolated input set of coefficients $\alpha_i$, $\alpha_n$ would be closer in distance to any parameter point-estimate interpolated between CPS-regularized parameter end-points with coefficients $\gamma_i$, $\gamma_n$, compared to non-CPS-regularized.

First we state the loss terms for a ground-truth parameter $\hat{\theta}$ corresponding to $\hat{\mathbf{x}}$. 
Recall from Theorem 2 that the loss terms in the end-point parameters converge in the same direction $\frac{\partial \mathcal{L}(\theta_i; X_i)}{\partial \theta_i} \cdot \frac{\partial \mathcal{L}(\theta_n; X_n)}{\partial \theta_n} > 0$.
We find here a similarity to CPS, where in order for the loss to converge $\mathcal{L}(\hat{\theta}) \rightarrow 0$
the product of the loss terms for the ground-truth parameter also converges in the same direction $\frac{\partial \mathcal{L}(\hat{\theta}; X_i)}{\partial \hat{\theta}} \cdot \frac{\partial \mathcal{L}(\hat{\theta}; X_n)}{\partial \hat{\theta}} > 0$.

\begin{equation}
\small
\begin{split}
\hat{\theta} & := \hat{\theta} -
\sum_{e}^{E-1}
\bigg[
\frac{\partial \mathcal{L}(\hat{\theta}_{t=e}; \alpha_i \mathbf{x_i} + \alpha_n \mathbf{x_n})}{\partial \hat{\theta}_{t=e}}
\bigg] \\
&= \hat{\theta} -
\sum_{e}^{E-1}
\bigg[
\alpha_i \frac{\partial \mathcal{L}(\hat{\theta}_{t=e}; \mathbf{x_i})}{\partial \hat{\theta}_{t=e}}
+ \alpha_n \frac{\partial \mathcal{L}(\hat{\theta}_{t=e}; \mathbf{x_n})}{\partial \hat{\theta}_{t=e}}
\bigg] \textnormal{\phantom{====}(by Corollary 2)}\\
\end{split}
\end{equation}

Next, as the summation of loss per epoch indicates the region in the parameter space with respect to a constant initialization (as in Theorem 2), 
we compute the distance in the summation of loss for $\hat{\theta}$ compared to the boundary or end-point parameters of an uncompressed space.
As the component loss terms are not regularized with respect to each other, their product may or may not be greater than 0, i.e. their product is in the range [-1, 1].

\begin{equation}
\small
\begin{split}
\mathbb{E}
\Bigg[
\Bigg|
& \bigg[
\alpha_i \frac{\partial \mathcal{L}(\hat{\theta}_{t=e}; \mathbf{x_i})}{\partial \hat{\theta}_{t=e}}
+ \alpha_n \frac{\partial \mathcal{L}(\hat{\theta}_{t=e}; \mathbf{x_n})}{\partial \hat{\theta}_{t=e}} 
-
\gamma_i \frac{\partial \mathcal{L}(\theta_{i, t=e}; \mathbf{x_i})}{\partial \theta_{i, t=e}} 
\bigg|_{\frac{\partial \mathcal{L}(\theta_i; \mathbf{x_i})}{\partial \theta_i} \cdot \frac{\partial \mathcal{L}(\theta_n; \mathbf{x_n})}{\partial \theta_n} \sim [-1,1]}
\bigg] \\
- 
& \bigg[
\alpha_i \frac{\partial \mathcal{L}(\hat{\theta}_{t=e}; \mathbf{x_i})}{\partial \hat{\theta}_{t=e}}
+ \alpha_n \frac{\partial \mathcal{L}(\hat{\theta}_{t=e}; \mathbf{x_n})}{\partial \hat{\theta}_{t=e}}
-
\gamma_n \frac{\partial \mathcal{L}(\theta_{n, t=e}; \mathbf{x_n})}{\partial \theta_{n, t=e}} 
\bigg|_{\frac{\partial \mathcal{L}(\theta_n; \mathbf{x_i})}{\partial \theta_i} \cdot \frac{\partial \mathcal{L}(\theta_n; \mathbf{x_n})}{\partial \theta_n} \sim [-1,1]}
\bigg]
\Bigg|
\Bigg] \\
= 
\mathbb{E}
\Bigg[
\Bigg|
& 2 \bigg[
\alpha_i \frac{\partial \mathcal{L}(\hat{\theta}_{t=e}; \mathbf{x_i})}{\partial \hat{\theta}_{t=e}}
+ \alpha_n \frac{\partial \mathcal{L}(\hat{\theta}_{t=e}; \mathbf{x_n})}{\partial \hat{\theta}_{t=e}}
\bigg] \\
+
& \bigg[
\gamma_n \frac{\partial \mathcal{L}(\theta_{n, t=e}; \mathbf{x_n})}{\partial \theta_{n, t=e}} 
\bigg|_{\frac{\partial \mathcal{L}(\theta_n; \mathbf{x_i})}{\partial \theta_i} \cdot \frac{\partial \mathcal{L}(\theta_n; \mathbf{x_n})}{\partial \theta_n} \sim [-1,1]} 
- 
\gamma_i \frac{\partial \mathcal{L}(\theta_{i, t=e}; \mathbf{x_i})}{\partial \theta_{i, t=e}} 
\bigg|_{\frac{\partial \mathcal{L}(\theta_n; \mathbf{x_i})}{\partial \theta_i} \cdot \frac{\partial \mathcal{L}(\theta_n; \mathbf{x_n})}{\partial \theta_n} \sim [-1,1]}
\bigg]
\Bigg|
\Bigg]
\end{split}
\hspace{1.5cm}
\end{equation}

Then, we evaluate the summation of loss for $\hat{\theta}$ against the boundary or end-point parameters of a compressed space. 

\begin{equation}
\small
\begin{split}
\mathbb{E}
\Bigg[
\Bigg|
& \bigg[
\alpha_i \frac{\partial \mathcal{L}(\hat{\theta}_{t=e}; \mathbf{x_i})}{\partial \hat{\theta}_{t=e}}
+ \alpha_n \frac{\partial \mathcal{L}(\hat{\theta}_{t=e}; \mathbf{x_n})}{\partial \hat{\theta}_{t=e}} 
-
\gamma_i \frac{\partial \mathcal{L}(\theta_{i, t=e}; \mathbf{x_i})}{\partial \theta_{i, t=e}} 
\bigg|_{\frac{\partial \mathcal{L}(\theta_i; \mathbf{x_i})}{\partial \theta_i} \cdot \frac{\partial \mathcal{L}(\theta_n; \mathbf{x_n})}{\partial \theta_n} > 0}
\bigg] \\
- 
& \bigg[
\alpha_i \frac{\partial \mathcal{L}(\hat{\theta}_{t=e}; \mathbf{x_i})}{\partial \hat{\theta}_{t=e}}
+ \alpha_n \frac{\partial \mathcal{L}(\hat{\theta}_{t=e}; \mathbf{x_n})}{\partial \hat{\theta}_{t=e}}
-
\gamma_n \frac{\partial \mathcal{L}(\theta_{n, t=e}; \mathbf{x_n})}{\partial \theta_{n, t=e}} 
\bigg|_{\frac{\partial \mathcal{L}(\theta_n; \mathbf{x_i})}{\partial \theta_i} \cdot \frac{\partial \mathcal{L}(\theta_n; \mathbf{x_n})}{\partial \theta_n} > 0}
\bigg]
\Bigg|
\Bigg] \\
= 
\mathbb{E}
\Bigg[
\Bigg|
& 2 \bigg[
\alpha_i \frac{\partial \mathcal{L}(\hat{\theta}_{t=e}; \mathbf{x_i})}{\partial \hat{\theta}_{t=e}}
+ \alpha_n \frac{\partial \mathcal{L}(\hat{\theta}_{t=e}; \mathbf{x_n})}{\partial \hat{\theta}_{t=e}}
\bigg] \\
+
& \bigg[
\gamma_n \frac{\partial \mathcal{L}(\theta_{n, t=e}; \mathbf{x_n})}{\partial \theta_{n, t=e}} 
\bigg|_{\frac{\partial \mathcal{L}(\theta_n; \mathbf{x_i})}{\partial \theta_i} \cdot \frac{\partial \mathcal{L}(\theta_n; \mathbf{x_n})}{\partial \theta_n} > 0} 
- 
\gamma_i \frac{\partial \mathcal{L}(\theta_{i, t=e}; \mathbf{x_i})}{\partial \theta_{i, t=e}} 
\bigg|_{\frac{\partial \mathcal{L}(\theta_n; \mathbf{x_i})}{\partial \theta_i} \cdot \frac{\partial \mathcal{L}(\theta_n; \mathbf{x_n})}{\partial \theta_n} > 0}
\bigg]
\Bigg|
\Bigg] \\
< 
\mathbb{E}
\Bigg[
\Bigg|
& 2 \bigg[
\alpha_i \frac{\partial \mathcal{L}(\hat{\theta}_{t=e}; \mathbf{x_i})}{\partial \hat{\theta}_{t=e}}
+ \alpha_n \frac{\partial \mathcal{L}(\hat{\theta}_{t=e}; \mathbf{x_n})}{\partial \hat{\theta}_{t=e}}
\bigg]  \textnormal{\phantom{==========================================}(by Theorem 2)} \\
+
& \bigg[
\gamma_n \frac{\partial \mathcal{L}(\theta_{n, t=e}; \mathbf{x_n})}{\partial \theta_{n, t=e}} 
\bigg|_{\frac{\partial \mathcal{L}(\theta_n; \mathbf{x_i})}{\partial \theta_i} \cdot \frac{\partial \mathcal{L}(\theta_n; \mathbf{x_n})}{\partial \theta_n} \sim [-1,1]} 
- 
\gamma_i \frac{\partial \mathcal{L}(\theta_{i, t=e}; \mathbf{x_i})}{\partial \theta_{i, t=e}} 
\bigg|_{\frac{\partial \mathcal{L}(\theta_n; \mathbf{x_i})}{\partial \theta_i} \cdot \frac{\partial \mathcal{L}(\theta_n; \mathbf{x_n})}{\partial \theta_n} \sim [-1,1]}
\bigg]
\Bigg|
\Bigg]
\end{split}
\hspace{1.46cm}
\end{equation}

We find that a ground-truth interpolated parameter is closer to a compressed space than an uncompressed space. 
This motivates our work that a compressed subspace contains a higher density of low-loss interpolated parameters than an uncompressed subspace.

\qed

\textbf{Assumptions. }
We also note a few (non-exhaustive) assumptions or criteria for interpreting the problem and evaluating the CPS.

\noindent\textbf{(Assumption 1)}
\textit{Each parameter point-estimate in the parameter space $\Theta$ can be evaluated against more than one input distribution.}

\noindent\textbf{(Assumption 2)}
\textit{The input space and parameter space are continuous, and interpolated points can be sampled from both spaces.}

\noindent\textbf{(Assumption 3)}
\textit{A low-loss parameter point-estimate for a test-time distribution $\theta \mapsto \hat{\mathbf{x}}$ should be computable with minimal storage and time complexity. }

\noindent\textbf{(Assumption 4)}
\textit{We assume no task index or conditioning input or meta-data of the test-time distribution. }

\newpage
\noindent\textbf{(Methodology) Conditioning inputs. }
Adaptation without conditioning inputs is a more difficult scenario; though it is used in adapt model parameters when the target distribution changes, it is not commonly used when the source distribution is constant (i.e. not used in adversarial/backdoor defenses). 
Detecting when shift has occurred is a major implementation hurdle in most work tackling domain shift and task shift. 
Conditioning inputs are not used w.r.t. a single test-time distribution (adversarial/backdoor attacks), but are usually provided in meta/continual learning, or domain adaptation. 
Many implementations use support sets in domain adaptation \citep{ganin2016domainadversarial, hoffman2017cycada, peng2019moment, sun2015return} and meta/continual learning \citep{finn2017modelagnostic} literature.
The implementations assume a task/domain $T_i$ is sampled from a task/domain distribution $p(T)$ associated with a dataset $D_i$, from which we sample a support set $D^{s}_{i}=(X^{s}_{i}, Y^{s}_{i})$ and a query set $D^{q}_{i}=(X^{q}_{i}, Y^{q}_{i})$, where a meta learner computes the optimal model parameters $\theta$ with respect to $D^{s}_{i}$ in order to predict accurately on $D^{q}_{i}$. The support set distribution approximates the query set distribution here.  
Other than support/query sets, conditioning inputs can manifest as known/seen embeddings:
\cite{cheung2019superposition} learns a set of parameters for each of the $K$ tasks, where these parameters are stored in superposition with each other. 
The task-specific models are accessed using task-specific “context” information $C_k$ that dynamically “routes” an input towards a specific model retrieved from this superposition. 
The context vectors $C_k$ are the headers for each task, and they are all orthogonal to each other, hence making no assumption of a metric that can connect or predict these vectors on new or unseen tasks. 
While there is work that aims to perform unsupervised domain adaptation \citep{ganin2016domainadversarial, zhou2021selfadversarial}, domain generalization \citep{arpit2021ensemble, cha2021swad} and unsupervised meta learning \citep{hsu2019unsupervised, khodadadeh2019unsupervised} without explicit conditioning inputs, 
there is a gap between their performance in an unsupervised setting compared to methods in supervised or settings with conditioning inputs.
To exemplify the difficulty of inference without conditioning inputs at test-time, 
we can refer to the gap between correctly detecting near-OOD (inputs that are near but not i.i.d to the input distribution) and far-OOD (inputs that are far from the input distribution) inputs as being outliers.
For far-OOD, an AUROC close to 99\% is attainable on CIFAR-100 (in) vs SVHN (out) \citep{pmlr-v119-sastry20a}. 
However, the AUROC for near-OOD is around 85\% for CIFAR-100 (in) vs CIFAR-10 (out) \citep{zhang2020hybrid}.
In other modalities, such as genomics, the AUROC of near-OOD detection can fall to 66\% \citep{10.5555/3454287.3455604}.
Thus, though removing the conditioning input assumption limits the maximum adaptation performance, it is a practical and realistic assumption we retain in our evaluation.

\subsection{Experiment configurations}

\textbf{Training regime}
Given that a large number of models are trained to evaluate this work (i.e. 3 models are trained in parallel per CPS configuration across all results),
except for Figure \ref{fig:beta1_timeseries} and Appendix A.3.1
which were trained to the full 5000 epochs, 
all models trained in this work (including CPS and baselines) are early-stopped when training loss reaches 1.0 (test set accuracy ranging from 55-65\%).
Though further convergence is possible, we make note of this computational constraint, and recommend evaluating the results in the context of clean test set accuracy presented in tables/figures. 
All train-time procedures use a seed of 1.
All test-time procedures use a seed of 100.
The exception to both is any use of Random-BadNet (backdoor perturbations used in CPS, backdoor attack, random permutations and labels), where the seed is the index of perturbation set starting from 1.
The defender uses a train-test split of 80-20\%.
For backdoor attack settings, the attacker uses a traintime-runtime split of 80-20\%; this means the attacker contributes 80\% of their private dataset (of which $p, \varepsilon = 0.4$ is poisoned) to the defender's dataset, and we evaluate the backdoor attack on the remaining 20\% at test-time (of which 1.0 is poisoned). 
All experiments were performed with respect to the constraints of $1 \times$ NVIDIA GeForce RTX 3070.

\begin{algorithm}[t]
  \footnotesize 
  \caption{CPS: Evaluate}
  \SetKwInOut{Input}{Input}
  \SetKwInOut{Output}{Output}
  \SetKwProg{evalSpace}{evalSpace}{}{}

    \evalSpace{$x, \{\theta_i\}^{N}, \{\alpha_i\}^{N}$}{
    \Input{Input $x$, Trained parameters $\{\theta_i\}^{N}$, Interpolation coefficients $\alpha_i \sim [0, 1]$ s.t. $\sum_{i}^{N} \alpha_i \leq N$}
    \Output{Predicted output $\hat{y}$}
    $\theta \gets \sum_{i}^{N} \alpha_i \theta_i$\\
    $\hat{y} \gets f(x, \theta)$\\
    \KwRet{$\hat{y}$}\;
  }
  \label{alg:inf}
\end{algorithm}

\textbf{Models}
In-line with work in loss landscape analysis \citep{fort2020deep}, we opt to use the minimum model required to fit/generalize a model to a test set. 
In-line with the implementation of \citet{10.5555/3045118.3045349} (to ensure the minimum CNN capcity to fit CIFAR100 test sets), 
we train each CNN with a cross entropy loss function and SGD optimizer for the following depth and width variations:
3-layer [16, 16, 16],
6-layer [16, 16, 16, 32, 32, 32],
9-layer [16, 16, 16, 32, 32, 32, 32, 32, 32],
6-layer-wide [256, 256, 256, 512, 512, 512].

\textbf{Adversarial Attack}
We use the Projected Gradient Descent (PGD) attack \citep{madry2018towards}. 
With this attack method, adversarial perturbation rate $\varepsilon_{\textnormal{a}}=0.1$.
can sufficiently bring down the attack success rate comparable and similar to that of $\varepsilon_{\textnormal{a}}=1.0$. 
Hence, we scale the perturbation against an upper limit $1.0$ in our experiments, i.e. $\varepsilon_{\textnormal{a}}^{'} = 0.4 \times \varepsilon_{\textnormal{a}}$.
We computes perturbations with respect to the gradients of the defender's model (a more pessimistic, white-box attack). 

\textbf{Backdoor Attack \& Random Permutations}
As we would like different trigger perturbations and target poison labels, such as when sampling 100 non-identical random permutations/labels or constructing 3 backdoored sets for CPS training, 
we implement Random-BadNet \citep{datta2022backdoors}, a variant of the baseline dirty-label backdoor attack algorithm, BadNet \citep{8685687}.
Alike to BadNet and many existing backdoor implementations in literature, Random-BadNet constructs a static trigger.
Instead of a single square in the corner as in the case of BadNet, Random-BadNet generates randomized pixels to return unique trigger patterns.
Attackers need to specify a set of inputs mapped to target poisoned labels $\{X_i:Y_i^{\textnormal{poison}}\} \in D_i$ to specify the intended label classification, backdoor perturbation rate $\varepsilon_i$ to specify the proportion of an input to be perturbed, and the poison rate 
$p_i = \frac{|X^{\textnormal{poison}}|}{|X^{\textnormal{clean}}| + |X^{\textnormal{poison}}|}$ 
to specify the proportion of the private dataset to contain backdoored inputs, to return $X^{\textnormal{poison}} = b_i(X_i, Y_i^{\textnormal{poison}}, \varepsilon_i, p_i)$. 
In this work,
the target poison labels are randomly-sampled,
the perturbation rate and poison rate are $p, \varepsilon = 0.4$.

\textbf{Stylization} We use the Adaptive Instance Normalization
(AdaIN) stylization method \citep{huang2017adain}, which is a standard method to stylize datasets such as stylized-ImageNet \citep{geirhos2018imagenettrained}.
Dataset stylization is considered as texture shift or domain shift in different literature. 
We randomly sample a distinct (non-repeating) style for each attacker.
$\alpha$ is the degree of stylization to apply; 1.0 means 100\% stylization, 0\% means no stylization; we set $\alpha=0.5$.
We follow the implementation in \citet{huang2017adain} and \citet{geirhos2018imagenettrained} and stylize CIFAR-10 with the Paintings by Numbers style dataset. We adapt the method for our attack, by first randomly sampling a distinct set of styles for each attacker, and stylizing each attacker's sub-dataset before the insertion of backdoor or adversarial perturbations. This shift also contributes to the realistic scenario that different agents may have shifted datasets given heterogenous sources. 

\textbf{Rotation}
At train-time we rotate the $N$ sets according to their respective index out of $N$: 
[90, 0, 270, 180, 45, 135, 61, 315, 60, 315, 20, 75]. These rotation degrees were hand-picked and intended to be diverse and varied.
With seed 3407, 
At test-time, we randomly-sampled 100 seeded rotations:
[20, 47, 77, 109, 304, 87, 304, 254, 146, 94, 98, 39, 306, 114, 267, 42, 231, 120, 40, 339, 352, 14, 264, 288, 203, 175, 308, 355, 324, 76, 213, 209, 167, 4, 170, 234, 120, 87, 43, 337, 300, 358, 29, 237, 107, 62, 84, 95, 9, 327, 203, 331, 1, 27, 59, 122, 52, 294, 64, 128, 263, 39, 141, 291, 25, 39, 176, 79, 104, 243, 265, 166, 270, 113, 23, 65, 297, 19, 196, 134, 119, 169, 42, 178, 250, 253, 276, 354, 291, 298, 20, 0, 343, 263, 164, 246, 217, 184, 163, 98].

\textbf{Data Augmentation} (e.g.,  CutMix~\citep{Yun_2019_ICCV} or MixUp \citep{zhang2018mixup}) is a common method to robustify models against domain shift \citep{Huang_2018_ECCV}, adversarial attacks \citep{zeng2020data}, and backdoor attacks \citep{borgnia2020strong}. 
We implement CutMix~\citep{Yun_2019_ICCV}, where augmentation takes place per batch, and training completes in accordance with aforementioned early stopping.
In the case of CutMix, for example, 
instead of removing pixels and filling them with black or grey pixels or Gaussian noise, we replace the removed regions with a patch from another image, while the ground truth labels are mixed proportionally to the number of pixels of combined images. 
$50\%$ of the defender's allocation of the dataset is assigned to augmentation.

\textbf{Adversarial Training} is a method that 
increases model robustness by injecting adversarial examples into the training set, and commonly used against adversarial attacks \citep{goodfellow2015explaining}, backdoor attacks \citep{geiping2021doesnt}, and domain adaptation \citep{JMLR:v17:15-239}. 
We retain the same adversarial configurations as in the adversarial attack (PGD, $\varepsilon=0.4$).
At each epoch, $50\%$ of the defender's allocation of the dataset is adversarially-perturbed.

\textbf{Backdoor Adversarial Training} \citep{geiping2021doesnt} extend the concept of adversarial training 
on defender-generated backdoor examples to insert their own triggers to existing labels.
We implement backdoor adversarial training~\citep{geiping2021doesnt}, where the generation of backdoor perturbations is through Random-BadNet~\citep{datta2022backdoors}), 
where $50\%$ of the defender's allocation of the dataset is assigned to backdoor perturbation,
$p, \varepsilon = 0.4$, 
and 20 different backdoor triggers used
(i.e. allocation of defender's dataset for each backdoor trigger pattern is
$(1-0.5) \times 0.8 \times \frac{1}{20}$
).

\textbf{Fast Geometric Ensembles \citep{garipov2018loss}}
train multiple networks to find their respective modes such that a high-accuracy path is formed that connects the modes in the parameter space. The method collects parameters at different checkpoints, which can subsequently be used as an ensemble.
We retain the same model configurations as the clean setting (3-layer CNN), train 5 ensembles on randomly-sampled random initializations until the early-stopping condition.

\noindent\textbf{Subspace Inference}
Inferencing w.r.t. the \textit{centre} of the subspace is denoted as computing a prediction $\bar{y} = \mathscr{f}(\theta^{*}; \hat{\mathbf{x}})$ with respect to a given input and the centre parameter point-estimate $\theta^{*} = \sum_i^N \frac{1}{N} \theta_i$ (the average of the end-point parameters). 
This form of inference has also been used in \cite{wortsman2021learning}.
Inferencing w.r.t. an \textit{ensemble} in the subspace is denoted as computing a mean prediction $\bar{y} = \frac{1}{M} \sum_j^M \mathscr{f}(\theta_j^{*}; \hat{\mathbf{x}})$ with respect to a given input and $M$ randomly-sampled parameter point-estimates in the subpace (we randomly sample $M=1000$ interpolation coefficients $\{ \alpha_{j, i}\}^{M \times N}$ to return an ensemble set of parameter point estimates $\{ \theta_j^{*} \}^M = \{ \alpha_{j, i}\}^{M \times N} \cdot \{\theta_i\}^N$.
For our ensemble, we evaluate against the mean prediction.
This form of inference has also been used in \cite{wortsman2021learning, garipov2018loss, fort2020deep}.
Inferencing w.r.t. the maximum-accuracy (lowest-loss) \textit{interpolated} point-estimate in the subspace is denoted as 
computing a prediction $\bar{y} = \mathscr{f}(\theta_{j^{*}}; \hat{\mathbf{x}})$ with respect to a given input and the lowest-loss parameter point-estimate 
where $j^{*} := \arg\min_{j \sim M} \mathcal{L}(\theta_j; \mathbf{x}, \mathbf{y})$; this is specifically a unique-task solution, where a low-loss parameter maps back to one task/distribution $\hat{\mathbf{x}} \mapsto \theta_{j^{*}}$.
For a multi-task solution, a low-loss parameter is mapped back to a set of $T$ task/distributions $\{\hat{\mathbf{x_t}}\}^T \mapsto \theta_{j^{*}}$, where $j^{*} := \arg\min_{j \sim M} \sum_t^T \mathcal{L}(\theta_j; \mathbf{x_t}, \mathbf{y_t})$.
We linearly-space 50 segments per parameter end-point pair (resulting in $50^3$ point-estimates to evaluate). There is enough distance between parameter end-points such that the interpolated points are not close approximations of the end-point (further verifiable from Figure \ref{fig:beta1_timeseries}). 
Inferencing w.r.t. interpolated points has also been used in \cite{NEURIPS2020_0607f4c7}. 
An additional baseline used in Table 2
is inferencing w.r.t. the lowest-loss \textit{boundary} parameter in the subspace, denoted as computing a prediction $\bar{y} = \mathscr{f}(\theta_{i^*}; \hat{\mathbf{x}})$ with respect to a given input and lowest-loss boundary parameter,
where the latter is the parameter from a set of $N$ boundary or end-point parameters that returns the lowest loss 
 $i^{*} := \arg\min_{i \sim N} \mathcal{L}(\theta_i; \mathbf{x}, \mathbf{y})$.
This baseline informs us whether the lowest-loss interpolated parameter can outperform an uninterpolated SGD-trained parameter.
In Table 2
specifically, we do not include the boundary parameters in the interpolated parameters set (interpolation coefficients: [0,0,1],[0,1,0],[1,0,0]), 
so as to ensure the 2 sets are mutually-exclusive and avoid the reuse of the boundary parameter as the lowest-loss interpolated parameter; for example, in case there is another point that can also achieve similar accuracy.

\noindent\textbf{Hypernetworks. }
A hypernetwork $\mathscr{h}(\mathbf{x}, I) = \mathscr{f}(\mathbf{x}; \mathscr{mf}(\theta_{\mathscr{mf}}; I))$ 
is a pair of learners, the base learner $\mathscr{f}: \mathcal{X} \mapsto \mathcal{Y}$ and meta learner $\mathscr{mf}: \mathcal{I} \mapsto \Theta_{\mathscr{f}}$,
such that for the conditioning input $I$ of input $\mathbf{x}$ (where $\mathcal{X} \mapsto \mathcal{I}$),
$\mathscr{mf}$ produces the base learner parameters $\theta_{I} = \mathscr{mf}(\theta_{\mathscr{mf}}; I)$.
The function $\mathscr{mf}(\theta_{\mathscr{mf}}; I)$ takes a conditioning input $I$ 
to returns parameters $\theta_{I} \in \Theta_{\mathscr{f}}$ for $\mathscr{f}$. 
The meta learner parameters and base learner (of each respective distribution) parameters reside in their distinct parameter spaces $\theta_{\mathscr{mf}} \in \Theta_{\mathscr{mf}}$ and $\theta_{\mathscr{f}} \in \Theta_{\mathscr{f}}$.
The learner $\mathscr{f}$ takes an input $\mathbf{x}$ and returns an output $\bar{y} = \mathscr{f}(\theta_{I}; \mathbf{x})$ that depends on both $\mathbf{x}$ and the task-specific input $I$. 
$T$ is number of tasks, $t$ is index of specific task being evaluated ($t^{*}$ being current task), $\omega$ is task regularizer (from \cite{vonoswald2020continual} implementation).
We retain all other implementation configurations, such as the use of Adam optimizer, setting task regularizer $\omega =  0.01$.
$\theta_{\mathscr{mf}}$ is the hypernetwork parameters at the current task's training timestep, 
$\theta_{\mathscr{mf}}^{*}$ is the hypernetwork parameters before attempting to learn task $t^{*}$, 
and $\Delta \theta_{\mathscr{mf}}$ is the candidate parameter change computed by the optimizer (Adam). 
Our base learner is the 6-layer CNN.
Unlike \citet{vonoswald2020continual}, we wish to retain the assumption of the rest of the paper where we do not require conditioning inputs; hence
we do not use any separate conditioning input (i.e. task embeddings), and use the test-time input as the primary argument, i.e. $\mathcal{I} \equiv \mathcal{X}$. As such, the lack of use of task embeddings is one of the primary changes to our hypernetworks baseline based on \citet{vonoswald2020continual}.
In terms of parameter storage, only the weights of the hypernetwork (and not individual weights of points in the CPS) would need to be stored.
As we only adapt \citet{vonoswald2020continual}'s hypernetwork implementation, we use this as our primary baseline, and recommend comparison to other baselines listed in \citet{vonoswald2020continual}.

Starting from a random initialization, 
we train the hypernetwork first on CIFAR10 (this can be interpreted as re-initializing the network with pre-trained weights, which based on findings in \cite{NEURIPS2020_0607f4c7} may indicate that subsequent task parameters may reside in a shared low-loss basin), and store these parameters $\theta_{\mathscr{mf}, t}$ in a parameter set $\{\theta_{\mathscr{mf}, t}\}^T$.
Then we train on a subsequent CIFAR100 task with the Eqt. \ref{equation:hloss} loss function, 
where we compute loss w.r.t. the inputs and current timestep's parameters, 
change in loss w.r.t. a previous task between using the proposed parameters and the parameters last updated at that task (stored in the parameter set; this is enumerated for all past tasks in sequence), 
and the distance between the current parameters and all prior parameters.
Unlike our prior (multi-task) implementation, (i) we minimize the cosine distance between a current task's parameters against prior parameters sequentially, not in parallel (i.e. the subspace end-points are sequentially fixed in the parameter space, we cannot dynamically move the subspace towards a different region), 
(ii) we do not gain visibility to all task parameters at once, 
and (iii) we are computing distance w.r.t. multi-task parameters (i.e. each task's parameters is applicable to its own and prior tasks).

\textbf{Task Interpolation}
We evaluate interpolated tasks by linearly interpolating a set of tasks: 
for a set of tasks $\{X\}^N$ where each task $X$ is a set of inputs $x$ (and the number of inputs per task is identical, i.e. $\{\{x\}^M\}^N$),
we linearly-interpolate between indexed inputs across a task set to return an interpolated task: $\hat{\mathbf{x}} = \sum_i^N \sum_j^M \alpha_i x_{i,j}$.
In this work, we only interpolate between task sets with identical coarse labels (label-shared). 
We do not interpolate batches, meaning we do not take a subset of one batch in a task and merge it with another subset of another batch in a task (i.e. all inputs are unperturbed, but each interpolated task contains distinct subsets of different task). We do not implement this setting, as we present results on multi-task solutions, which are evaluated on multiple unperturbed tasks with a single parameter point-estimate, thus an interpolated batch setting is already evaluatable.

We linearly-space the range [0,1] into 5 segments (returning $5^3$ different tasks).
We ignore the [0,0,0] case, as this results in a black image set, containing no identifiable task-specific features, and thus likely to return random labels - we leave this case out of our evaluation to filter out this noise.
For each interpolated image, we clip the colour values per pixel in the range [0, 255].
We keep the [1,1,1] case as, even though it could be all white theoretically, it would only be all white if each and every pixel exceeded 255; in practice (for example, as shown in Appendix A.2.1,
pixels may contain colour channel values whose weighted sum per channel may still be less than 255.

We do not interpolate batches, meaning we do not take a subset of one batch in a task and merge it with another subset of another batch in a task (i.e. all inputs are unperturbed, but each interpolated task contains distinct subsets of different task). We do not implement this setting, as we present results on multi-task solutions, which are evaluated on multiple unperturbed tasks with a single parameter point-estimate, thus an interpolated batch setting is already evaluatable. 

\begin{figure*}[h]
\renewcommand{\fnum@table}{Appendix~A.2.\thetable}
\renewcommand{\fnum@figure}{Appendix~A.2.\thefigure}
\setcounter{figure}{0}
\setcounter{table}{0}
    \centering
    \includegraphics[width=\textwidth]{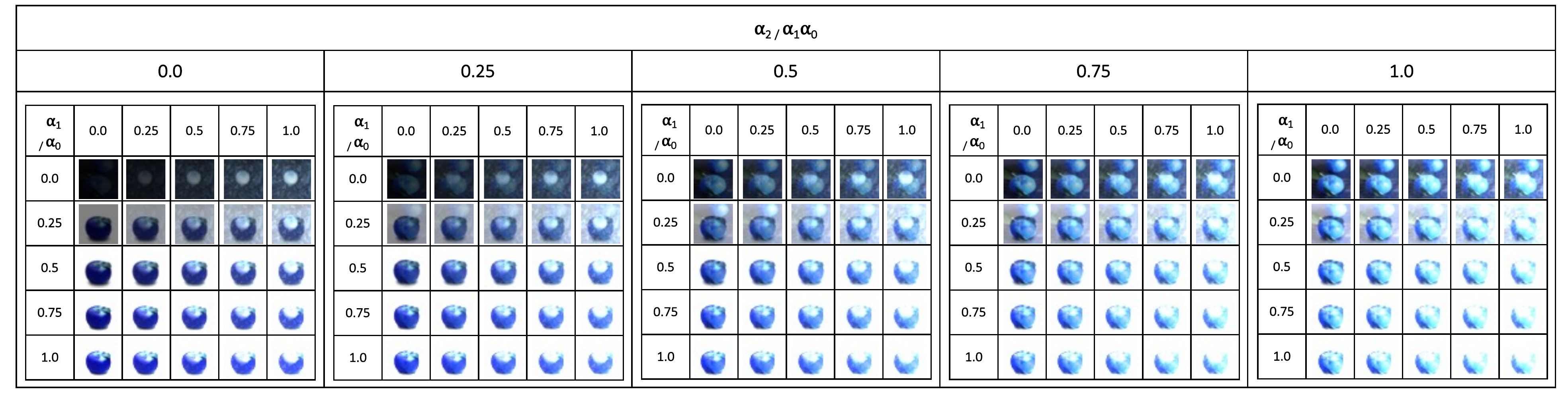}
    \caption{
    Samples of interpolated images.
    The label of the variations of interpolated image across tasks is 4. 
    The task end-points are: 3 tasks, 3 $\times$ same coarse label, train-time task set.
    }
    \label{fig:intp_images}
\end{figure*}

\textbf{Single Test-time Distribution Dataset}
We use the CIFAR10 dataset, which contains 10 classes, 60,0000 inputs, and 3 colour channels \citep{krizhevsky2009learning}.
For test-time perturbations, we evaluate on 100 different examples and tabulate the mean and standard deviation.
For train-time perturbations (backdoor), we insert a single backdoor trigger into the defender set before CPS initiated.

\textbf{Multiple Test-time Distributions Dataset}
We use CIFAR100 dataset, a common baseline for meta/continual learning settings \citep{krizhevsky2009learning}.
Similar to CIFAR10 except in that it contains 100 classes containing 600 images each. 
There are 500 training images and 100 testing images per class. The 100 classes in the CIFAR100 are grouped into 20 superclasses. Each image comes with a "fine" label (the class to which it belongs) and a "coarse" label (the superclass to which it belongs).
There are at most 5 tasks of the same coarse label (given there are 5 fine labels per class) per coarse label set. 
For 5-label-per-set tasks, there are at most 5 different coarse label sets available (for 10-label-per-set tasks, there are at most 2).
To evaluate the use of CPS-regularization in hypernetworks for continual learning, 
the dataset is used in-line with the Split-CIFAR10/100 setting as described in \citet{10.5555/3305890.3306093} and \citet{vonoswald2020continual}.
\textit{N-label-set per task} refers to the number of coarse labels contained/mapped in the task; 5(10)-label-set per task means we have 5(10) coarse labels mapped within the task, while 2 $\times$ 5(10)-label-set per task means we have 5(10) coarse labels mapped within the task but of 2 different fine label sets. 
\textit{Distinctively-different coarse labels (per task)} is a distinction from \textit{different coarse labels (of the same set)}, where the former has a unique coarse label set assigned per task (no shared coarse labels across the tasks), while the latter has a different set of tasks that share coarse labels (but different fine labels). 
We provide a breakdown of the labels (both coarse and fine) used per task; each task (as a self-contained training set/dataloader) is indicated by the square brackets [].

\begin{itemize}
    \item (Appendix A.3.8)
    Seen tasks (3 tasks, 3 $\times$ same coarse label, train-time task set):
    [4 (aquatic mammals $\rightarrow$ beaver), 1 (fish $\rightarrow$ aquarium fish), 54 (flowers $\rightarrow$ orchids), 9 (food $\rightarrow$ containers), 0 (fruit and vegetables $\rightarrow$ apples)], 
    [30 (aquatic mammals $\rightarrow$ dolphin), 32 (fish $\rightarrow$ flatfish), 62 (flowers $\rightarrow$ poppies), 10 (food $\rightarrow$ bottles), 51 (fruit and vegetables $\rightarrow$ mushrooms)], 
    [55 (aquatic mammals $\rightarrow$ otter), 67 (fish $\rightarrow$ ray), 70 (flowers $\rightarrow$ roses), 16(food $\rightarrow$ bowls), 53 (fruit and vegetables $\rightarrow$ oranges)]
    
    \item (Appendix A.3.8)
    Unseen tasks (2 tasks, 2 $\times$ same coarse label but unseen fine label):
    [72 (aquatic mammals $\rightarrow$ seal), 73 (fish $\rightarrow$ shark), 82 (flowers $\rightarrow$ sunflowers), 28 (food $\rightarrow$ cups), 57 (fruit and vegetables $\rightarrow$ pears)], 
    [95 (aquatic mammals $\rightarrow$ whale), 91 (fish $\rightarrow$ trout), 92 (flowers $\rightarrow$ tulips), 61 (food $\rightarrow$ plates), 83 (fruit and vegetables $\rightarrow$ sweet peppers)]
    
    \item (Appendix A.3.8)
    Unseen tasks (5 tasks, 1 $\times$ 5-label-set per task, 3 $\times$ same coarse label but unseen fine label, 2 $\times$ distinctly different coarse label): 
    [22 (household electrical devices $\rightarrow$ clock), 5 (household $\rightarrow$ furniture), 6 (insects $\rightarrow$ bee), 3 (large carnivores $\rightarrow$ bear), 12 (large man-made outdoor things $\rightarrow$ bridge)], 
    [39 (household electrical devices $\rightarrow$ computer keyboard), 20 (household $\rightarrow$ bed), 7(insects $\rightarrow$ beetle), 42 (large carnivores $\rightarrow$ leopard), 17 (large man-made outdoor things $\rightarrow$ castle)], 
    [40 (household electrical devices $\rightarrow$ lamp), 25 (household $\rightarrow$ chair), 14 (insects $\rightarrow$ butterfly), 43 (large carnivores $\rightarrow$ lion), 37 (large man-made outdoor things $\rightarrow$ house)], 
    [23 (large natural outdoor scenes $\rightarrow$ cloud), 15 (large omnivores and herbivores $\rightarrow$ camel), 34 (medium-sized mammals $\rightarrow$ fox), 26 (non-insect invertebrates $\rightarrow$ crab), 2 (people $\rightarrow$ baby)], 
    [27 (reptiles $\rightarrow$ crocodile), 36 (small mammals $\rightarrow$ hamster), 47 (trees $\rightarrow$ maple), 8 (vehicles 1 $\rightarrow$ bicycle), 41 (vehicles 2 $\rightarrow$ lawn-mower)]
    
    \item (Appendix A.3.6)
    2 tasks, 2 $\times$ same coarse label:
    [4 (aquatic mammals $\rightarrow$ beaver), 1 (fish $\rightarrow$ aquarium fish), 54 (flowers $\rightarrow$ orchids), 9 (food $\rightarrow$ containers), 0 (fruit and vegetables $\rightarrow$ apples)], 
    [30 (aquatic mammals $\rightarrow$ dolphin), 32 (fish $\rightarrow$ flatfish), 62 (flowers $\rightarrow$ poppies), 10 (food $\rightarrow$ bottles), 51 (fruit and vegetables $\rightarrow$ mushrooms)]
    
    \item (Appendix A.3.6)
    2 tasks, 2 $\times$ different coarse label: [4 (aquatic mammals $\rightarrow$ beaver), 1 (fish $\rightarrow$ aquarium fish), 54 (flowers $\rightarrow$ orchids), 9 (food $\rightarrow$ containers), 0 (fruit and vegetables $\rightarrow$ apples)], 
    [22 (household electrical devices $\rightarrow$ clock), 5 (household $\rightarrow$ furniture), 6 (insects $\rightarrow$ bee), 3 (large carnivores $\rightarrow$ bear), 12 (large man-made outdoor things $\rightarrow$ bridge)]
    
    \item (Appendix A.3.5)
    2 tasks (2 $\times$ 5-label-set per task, 2 $\times$ same coarse label per task, 0 $\times$ distinctly different coarse label per task): 
    [4 (aquatic mammals $\rightarrow$ beaver), 30 (aquatic mammals $\rightarrow$ dolphin), 1 (fish $\rightarrow$ aquarium fish), 32 (fish $\rightarrow$ flatfish), 54 (flowers $\rightarrow$ orchids), 62 (flowers $\rightarrow$ poppies), 9 (food $\rightarrow$ containers), 10 (food $\rightarrow$ bottles), 0 (fruit and vegetables $\rightarrow$ apples), 51 (fruit and vegetables $\rightarrow$ mushrooms)], 
    [55 (aquatic mammals $\rightarrow$ otter), 72 (aquatic mammals $\rightarrow$ seal), 67 (fish $\rightarrow$ ray), 73 (fish $\rightarrow$ shark), 70 (flowers $\rightarrow$ roses), 82 (flowers $\rightarrow$ sunflowers), 16 (food $\rightarrow$ bowls), 28 (food $\rightarrow$ cups), 53 (fruit and vegetables $\rightarrow$ oranges), 57 (fruit and vegetables $\rightarrow$ pears)]
    
    \item (Appendix A.3.5)
    3 tasks (2 $\times$ 5-label-set per task, 2 $\times$ same coarse label per task, 1 $\times$ distinctly different coarse label per task): 
    [4 (aquatic mammals $\rightarrow$ beaver), 30 (aquatic mammals $\rightarrow$ dolphin), 1 (fish $\rightarrow$ aquarium fish), 32 (fish $\rightarrow$ flatfish), 54 (flowers $\rightarrow$ orchids), 62 (flowers $\rightarrow$ poppies), 9 (food $\rightarrow$ containers), 10 (food $\rightarrow$ bottles), 0 (fruit and vegetables $\rightarrow$ apples), 51 (fruit and vegetables $\rightarrow$ mushrooms)], 
    [55 (aquatic mammals $\rightarrow$ otter), 72 (aquatic mammals $\rightarrow$ seal), 67 (fish $\rightarrow$ ray), 73 (fish $\rightarrow$ shark), 70 (flowers $\rightarrow$ roses), 82 (flowers $\rightarrow$ sunflowers), 16 (food $\rightarrow$ bowls), 28 (food $\rightarrow$ cups), 53 (fruit and vegetables $\rightarrow$ oranges), 57 (fruit and vegetables $\rightarrow$ pears)], 
    [22 (household electrical devices $\rightarrow$ clock), 39 (household electrical devices $\rightarrow$ clock), 5 (household $\rightarrow$ furniture), 20 (household $\rightarrow$ furniture), 6 (insects $\rightarrow$ bee), 7 (insects $\rightarrow$ bee), 3 (large carnivores $\rightarrow$ bear), 42 (large carnivores $\rightarrow$ bear), 12 (large man-made outdoor things $\rightarrow$ bridge), 17 (large man-made outdoor things $\rightarrow$ bridge)]
    
    \item (Appendix A.3.5)
    4 tasks (2 $\times$ 5-label-set per task, 2 $\times$ same coarse label per task, 2 $\times$ distinctly different coarse label per task): 
    [4 (aquatic mammals $\rightarrow$ beaver), 30 (aquatic mammals $\rightarrow$ dolphin), 1 (fish $\rightarrow$ aquarium fish), 32 (fish $\rightarrow$ flatfish), 54 (flowers $\rightarrow$ orchids), 62 (flowers $\rightarrow$ poppies), 9 (food $\rightarrow$ containers), 10 (food $\rightarrow$ bottles), 0 (fruit and vegetables $\rightarrow$ apples), 51 (fruit and vegetables $\rightarrow$ mushrooms)],  
    [55 (aquatic mammals $\rightarrow$ otter), 72 (aquatic mammals $\rightarrow$ seal), 67 (fish $\rightarrow$ ray), 73 (fish $\rightarrow$ shark), 70 (flowers $\rightarrow$ roses), 82 (flowers $\rightarrow$ sunflowers), 16 (food $\rightarrow$ bowls), 28 (food $\rightarrow$ cups), 53 (fruit and vegetables $\rightarrow$ oranges), 57 (fruit and vegetables $\rightarrow$ pears)], 
    [22 (household electrical devices $\rightarrow$ clock), 39 (household electrical devices $\rightarrow$ clock), 5 (household $\rightarrow$ furniture), 20 (household $\rightarrow$ furniture), 6 (insects $\rightarrow$ bee), 7 (insects $\rightarrow$ bee), 3 (large carnivores $\rightarrow$ bear), 42 (large carnivores $\rightarrow$ bear), 12 (large man-made outdoor things $\rightarrow$ bridge), 17 (large man-made outdoor things $\rightarrow$ bridge)], 
    [23 (large natural outdoor scenes $\rightarrow$ cloud), 33 (large natural outdoor scenes $\rightarrow$ forest), 15 (large omnivores and herbivores $\rightarrow$ camel), 19 (large omnivores and herbivores $\rightarrow$ cattle), 34 (medium-sized mammals $\rightarrow$ fox), 63 (medium-sized mammals $\rightarrow$ porcupine), 26 (non-insect invertebrates $\rightarrow$ crab), 45 (non-insect invertebrates $\rightarrow$ lobster), 2 (people $\rightarrow$ baby), 11 (people $\rightarrow$ boy)]
    
    \item (Appendix A.3.7)
    3 tasks (1 $\times$ 5-label-set per task, 3 $\times$ same coarse label): 
    [4 (aquatic mammals $\rightarrow$ beaver), 1 (fish $\rightarrow$ aquarium fish), 54 (flowers $\rightarrow$ orchids), 9 (food $\rightarrow$ containers), 0 (fruit and vegetables $\rightarrow$ apples)], 
    [30 (aquatic mammals $\rightarrow$ dolphin), 32 (fish $\rightarrow$ flatfish), 62 (flowers $\rightarrow$ poppies), 10 (food $\rightarrow$ bottles), 51 (fruit and vegetables $\rightarrow$ mushrooms)], 
    [55 (aquatic mammals $\rightarrow$ otter), 67 (fish $\rightarrow$ ray), 70 (flowers $\rightarrow$ roses), 16(food $\rightarrow$ bowls), 53 (fruit and vegetables $\rightarrow$ oranges)]
    
    \item (Appendix A.3.7)
    3 tasks (1 $\times$ 5-label-set per task, 2 $\times$ same coarse label, 1 $\times$ different coarse label): 
    [4 (aquatic mammals $\rightarrow$ beaver), 1 (fish $\rightarrow$ aquarium fish), 54 (flowers $\rightarrow$ orchids), 9 (food $\rightarrow$ containers), 0 (fruit and vegetables $\rightarrow$ apples)], 
    [30 (aquatic mammals $\rightarrow$ dolphin), 32 (fish $\rightarrow$ flatfish), 62 (flowers $\rightarrow$ poppies), 10 (food $\rightarrow$ bottles), 51 (fruit and vegetables $\rightarrow$ mushrooms)]], 
    [22 (household electrical devices $\rightarrow$ clock), 5 (household $\rightarrow$ furniture), 6 (insects $\rightarrow$ bee), 3 (large carnivores $\rightarrow$ bear), 12 (large man-made outdoor things $\rightarrow$ bridge)]
    
    \item (Appendix A.3.7)
    3 tasks (1 $\times$ 5-label-set per task, 3 $\times$ distinctly different coarse label): 
    [4 (aquatic mammals $\rightarrow$ beaver), 1 (fish $\rightarrow$ aquarium fish), 54 (flowers $\rightarrow$ orchids), 9 (food $\rightarrow$ containers), 0 (fruit and vegetables $\rightarrow$ apples)], 
    [22 (household electrical devices $\rightarrow$ clock), 5 (household $\rightarrow$ furniture), 6 (insects $\rightarrow$ bee), 3 (large carnivores $\rightarrow$ bear), 12 (large man-made outdoor things $\rightarrow$ bridge)],  
    [23 (large natural outdoor scenes $\rightarrow$ cloud), 15 (large omnivores and herbivores $\rightarrow$ camel), 34 (medium-sized mammals $\rightarrow$ fox), 26 (non-insect invertebrates $\rightarrow$ crab), 2 (people $\rightarrow$ baby)]
    
    \item (Appendix A.3.7)
    5 tasks (1 $\times$ 5-label-set per task, 5 $\times$ same coarse label): 
    [4 (aquatic mammals $\rightarrow$ beaver), 1 (fish $\rightarrow$ aquarium fish), 54 (flowers $\rightarrow$ orchids), 9 (food $\rightarrow$ containers), 0 (fruit and vegetables $\rightarrow$ apples)], 
    [30 (aquatic mammals $\rightarrow$ dolphin), 32 (fish $\rightarrow$ flatfish), 62 (flowers $\rightarrow$ poppies), 10 (food $\rightarrow$ bottles), 51 (fruit and vegetables $\rightarrow$ mushrooms)], 
    [55 (aquatic mammals $\rightarrow$ otter), 67 (fish $\rightarrow$ ray), 70 (flowers $\rightarrow$ roses), 16(food $\rightarrow$ bowls), 53 (fruit and vegetables $\rightarrow$ oranges)], 
    [72 (aquatic mammals $\rightarrow$ seal), 73 (fish $\rightarrow$ shark), 82 (flowers $\rightarrow$ sunflowers), 28 (food $\rightarrow$ cups), 57 (fruit and vegetables $\rightarrow$ pears)],  
    [95 (aquatic mammals $\rightarrow$ whale), 91 (fish $\rightarrow$ trout), 92 (flowers $\rightarrow$ tulips), 61 (food $\rightarrow$ plates), 83 (fruit and vegetables $\rightarrow$ sweet peppers)]
    
    \item (Appendix A.3.7)
    5 tasks (1 $\times$ 5-label-set per task, 2 $\times$ same coarse label, 3 $\times$ distinctly different coarse label): 
    [4 (aquatic mammals $\rightarrow$ beaver), 1 (fish $\rightarrow$ aquarium fish), 54 (flowers $\rightarrow$ orchids), 9 (food $\rightarrow$ containers), 0 (fruit and vegetables $\rightarrow$ apples)], 
    [30 (aquatic mammals $\rightarrow$ dolphin), 32 (fish $\rightarrow$ flatfish), 62 (flowers $\rightarrow$ poppies), 10 (food $\rightarrow$ bottles), 51 (fruit and vegetables $\rightarrow$ mushrooms)]], 
    [22 (household electrical devices $\rightarrow$ clock), 5 (household $\rightarrow$ furniture), 6 (insects $\rightarrow$ bee), 3 (large carnivores $\rightarrow$ bear), 12 (large man-made outdoor things $\rightarrow$ bridge)], 
    [23 (large natural outdoor scenes $\rightarrow$ cloud), 15 (large omnivores and herbivores $\rightarrow$ camel), 34 (medium-sized mammals $\rightarrow$ fox), 26 (non-insect invertebrates $\rightarrow$ crab), 2 (people $\rightarrow$ baby)], 
    [27 (reptiles $\rightarrow$ crocodile), 36 (small mammals $\rightarrow$ hamster), 47 (trees $\rightarrow$ maple), 8 (vehicles 1 $\rightarrow$ bicycle), 41 (vehicles 2 $\rightarrow$ lawn-mower)]

    \item 
    (Appendix A.3.7)
    5 tasks (1 $\times$ 5-label-set per task, 2 $\times$ same coarse label, 3 $\times$ different coarse label of same set): 
    [4 (aquatic mammals $\rightarrow$ beaver), 1 (fish $\rightarrow$ aquarium fish), 54 (flowers $\rightarrow$ orchids), 9 (food $\rightarrow$ containers), 0 (fruit and vegetables $\rightarrow$ apples)], 
    [30 (aquatic mammals $\rightarrow$ dolphin), 32 (fish $\rightarrow$ flatfish), 62 (flowers $\rightarrow$ poppies), 10 (food $\rightarrow$ bottles), 51 (fruit and vegetables $\rightarrow$ mushrooms)], 
    [22 (household electrical devices $\rightarrow$ clock), 5 (household $\rightarrow$ furniture), 6 (insects $\rightarrow$ bee), 3 (large carnivores $\rightarrow$ bear), 12 (large man-made outdoor things $\rightarrow$ bridge)],  
    [39 (household electrical devices $\rightarrow$ clock), 20 (household $\rightarrow$ furniture), 7 (insects $\rightarrow$ bee), 42 (large carnivores $\rightarrow$ bear), 17 (large man-made outdoor things $\rightarrow$ bridge)], 
    [40 (household electrical devices $\rightarrow$ lamp), 25 (household $\rightarrow$ chair), 14 (insects $\rightarrow$ butterfly), 43 (large carnivores $\rightarrow$ lion), 37 (large man-made outdoor things $\rightarrow$ house)]
    
\end{itemize}

\newpage

\subsection{Supporting Results}

\begin{table*}[t]
\renewcommand{\fnum@table}{Appendix~A.3.\thetable}
\renewcommand{\fnum@figure}{Appendix~A.3.\thefigure}
\setcounter{figure}{2}
\setcounter{table}{2}
\centering
	\begin{minipage}{\textwidth}
	\centering
	\includegraphics[width=0.49\textwidth]{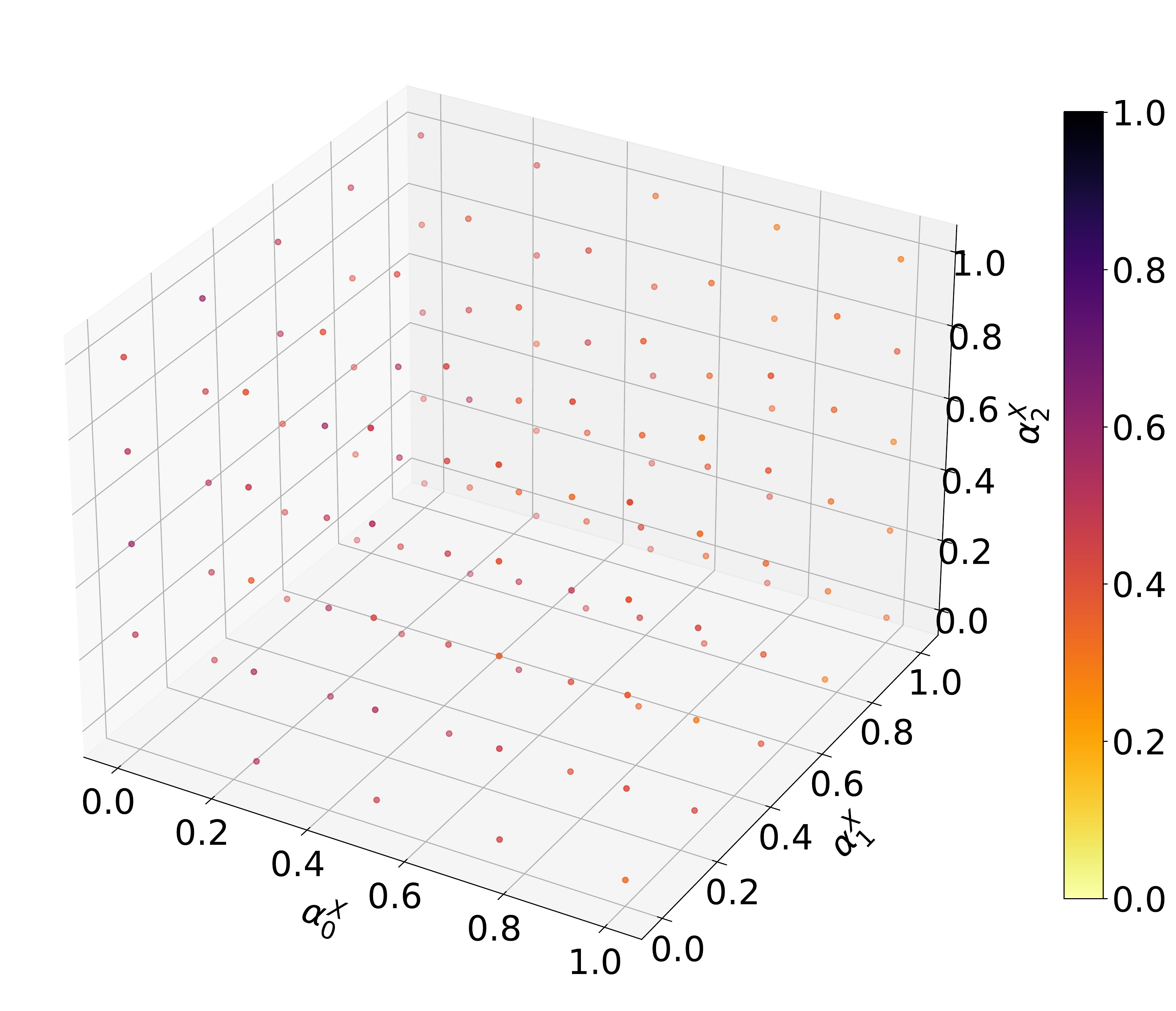}
    \includegraphics[width=0.49\textwidth]{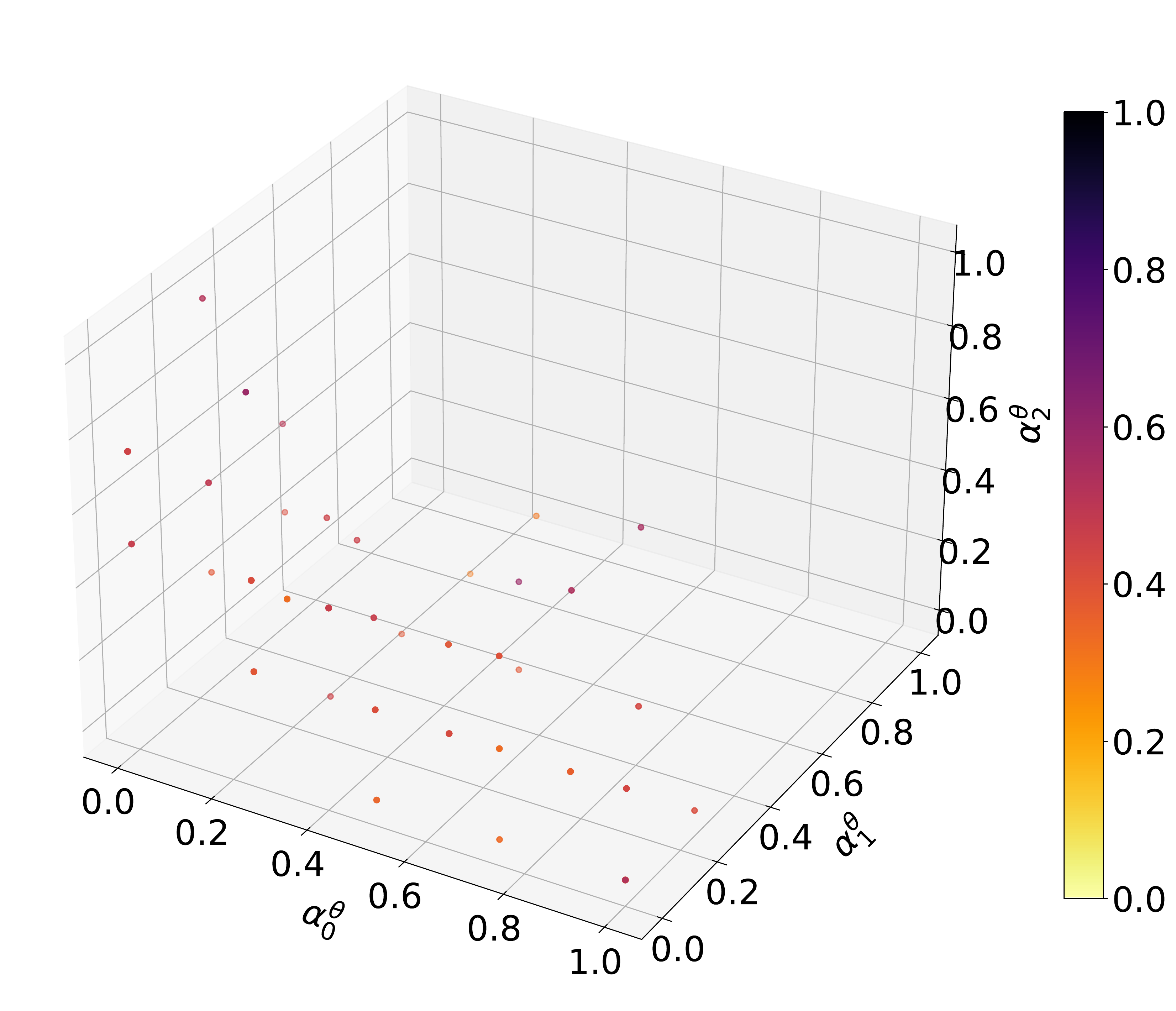}
    \begin{subfigure}[]{\textwidth}
         \centering
         \includegraphics[width=0.3\textwidth]{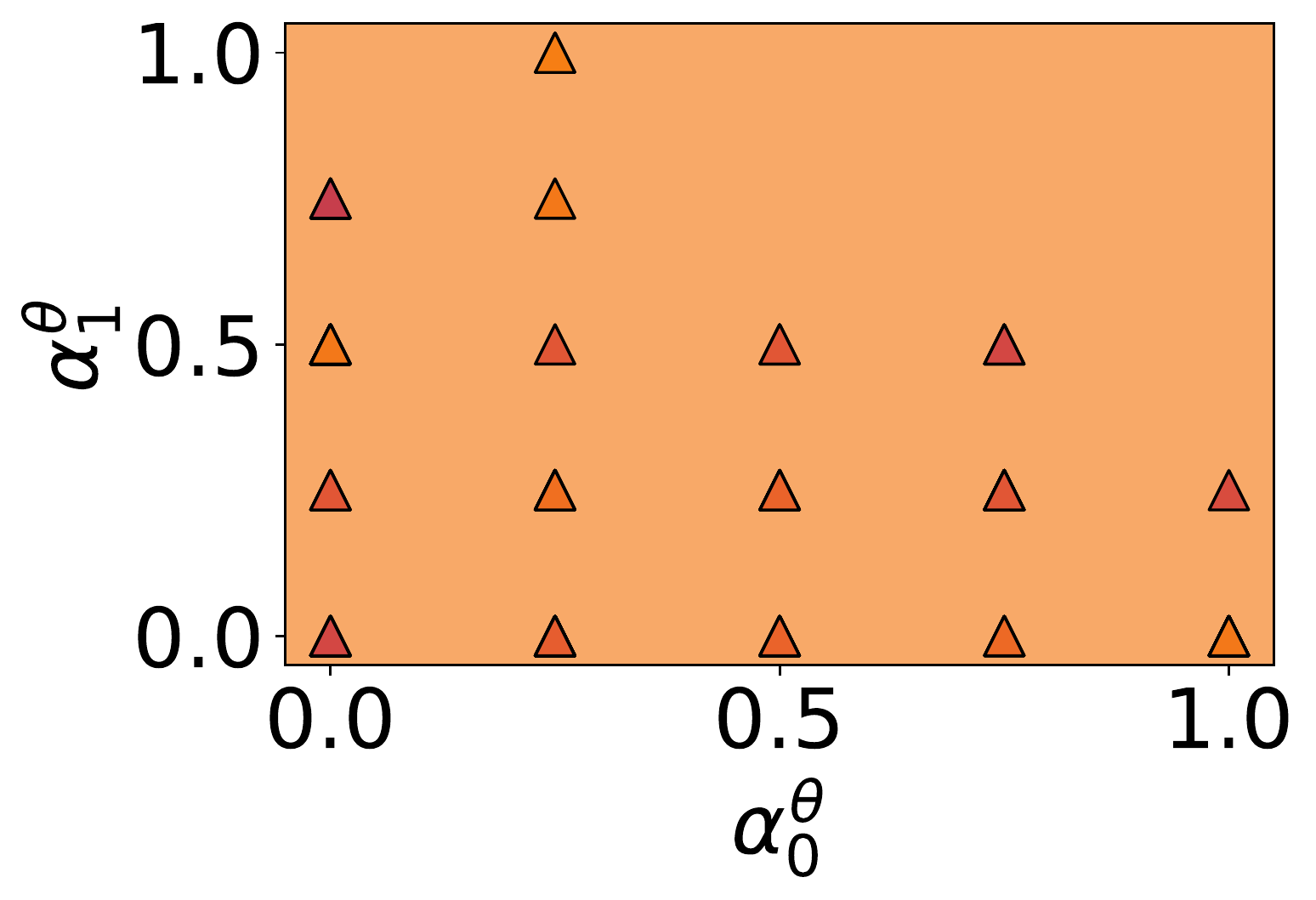}
        \includegraphics[width=0.3\textwidth]{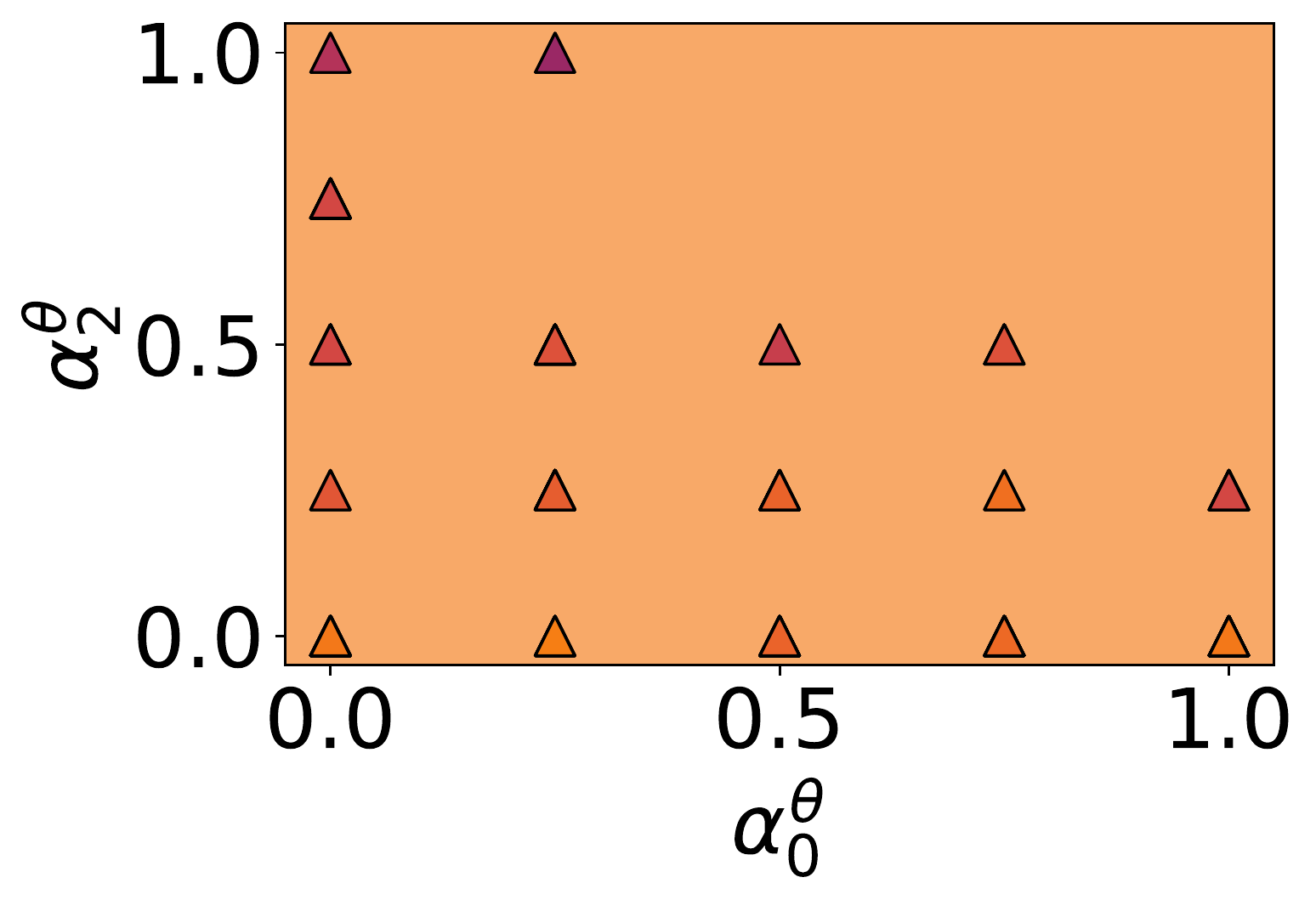}
        \includegraphics[width=0.3\textwidth]{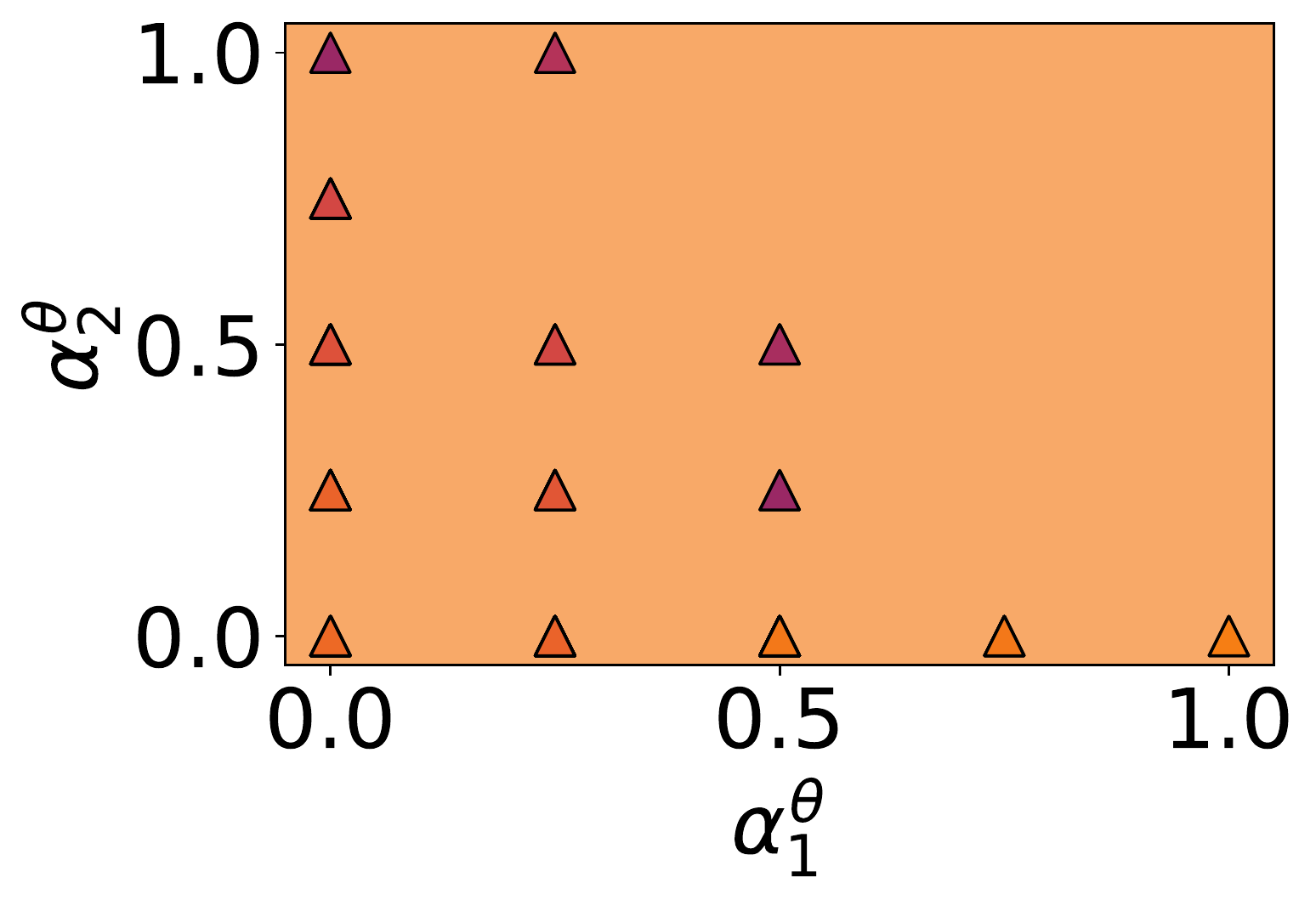}
         \caption{$\beta=0.0$}
     \end{subfigure}
	\end{minipage}
	\hfill
	\begin{minipage}{\textwidth}
	\centering
	\includegraphics[width=0.49\textwidth]{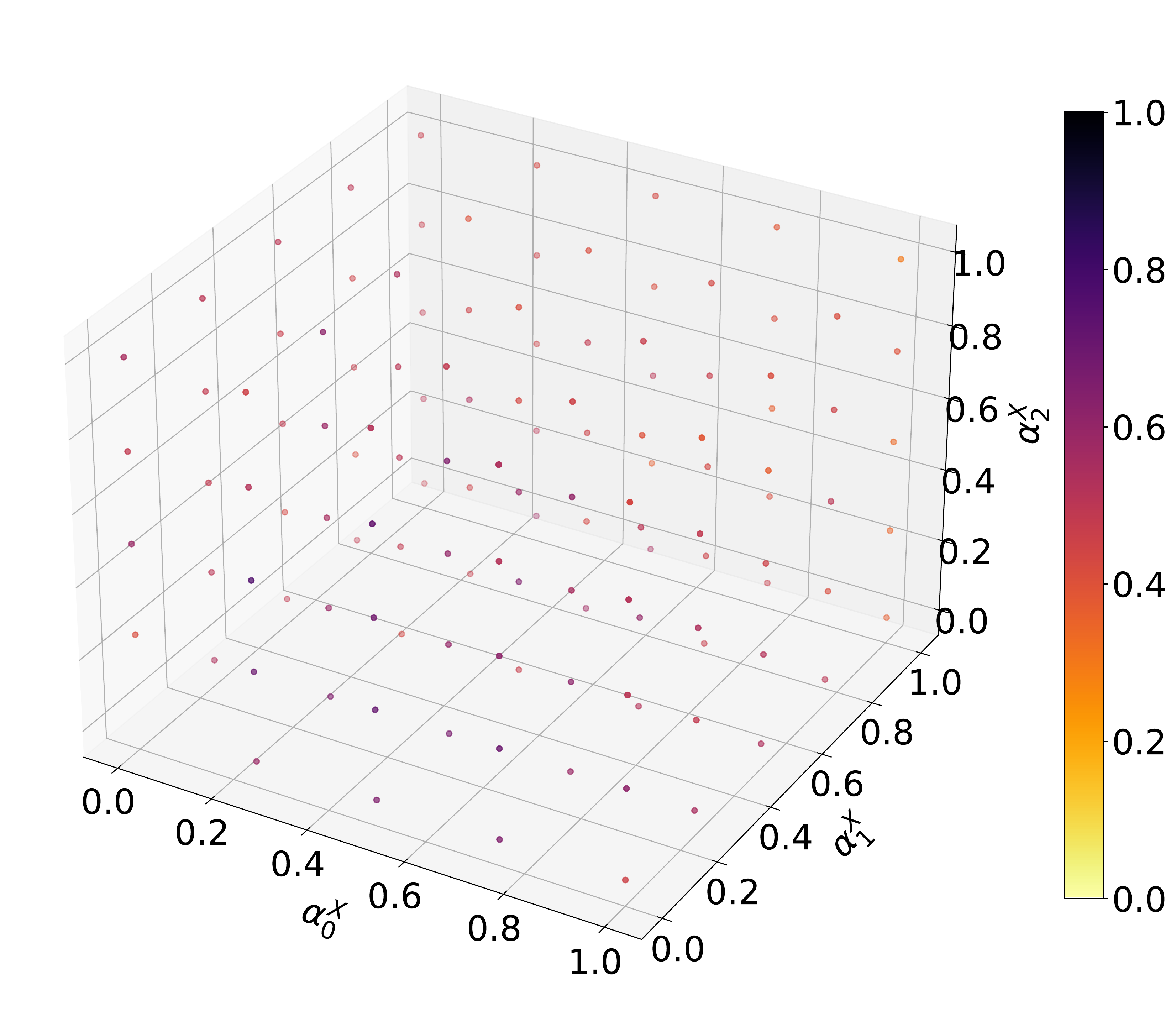}
    \includegraphics[width=0.49\textwidth]{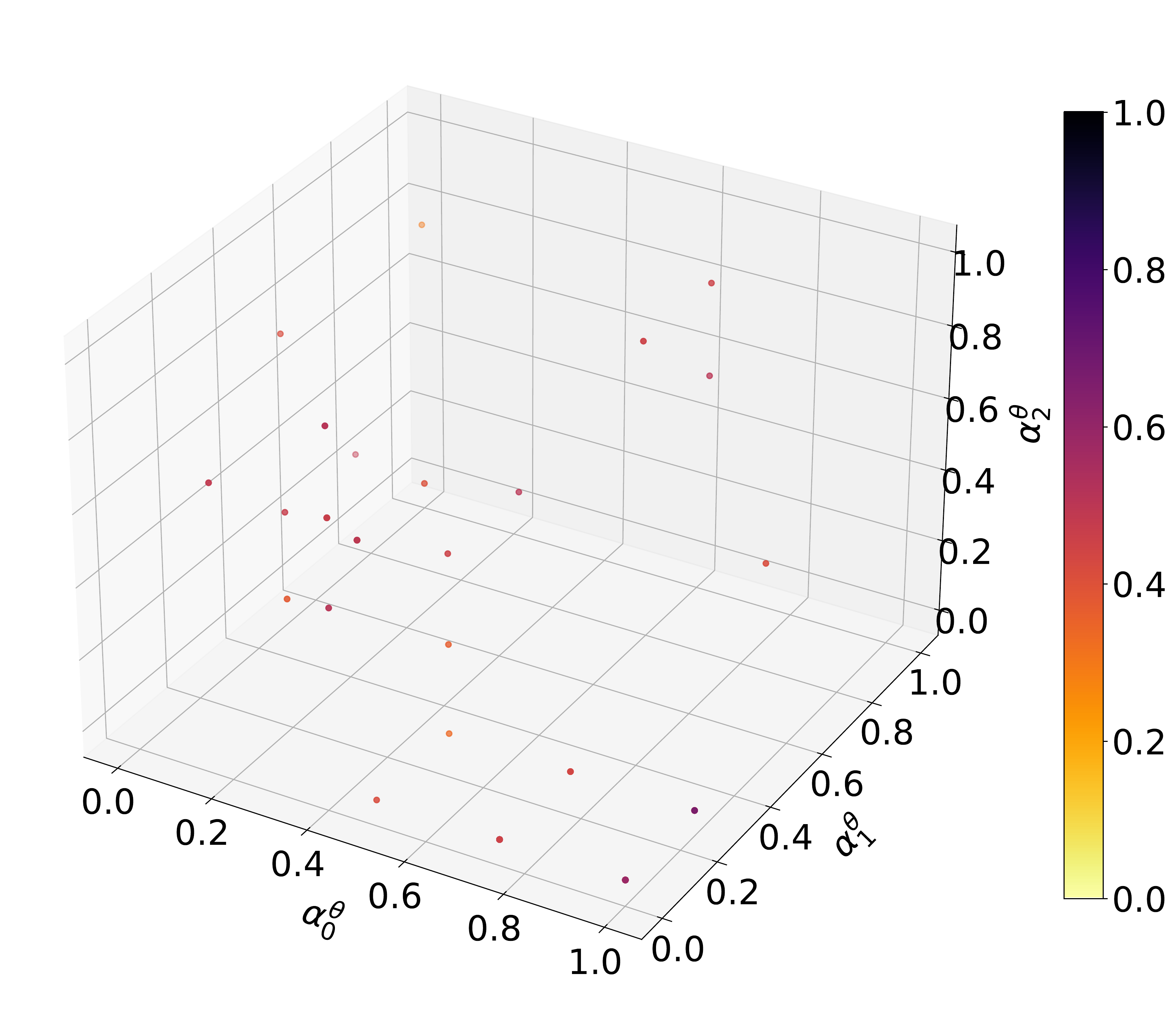}
	\begin{subfigure}[]{\textwidth}
         \centering
         \includegraphics[width=0.3\textwidth]{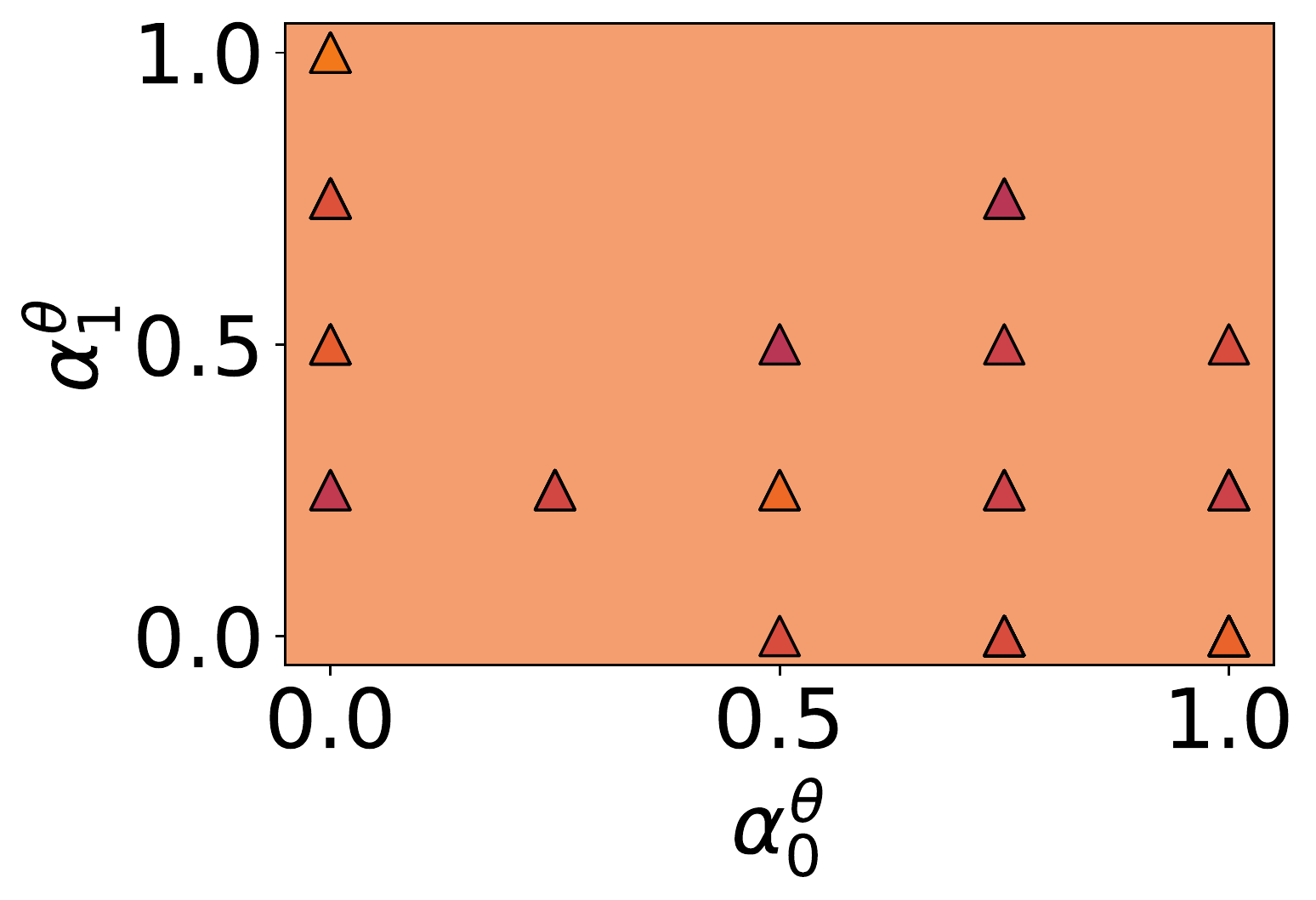}
        \includegraphics[width=0.3\textwidth]{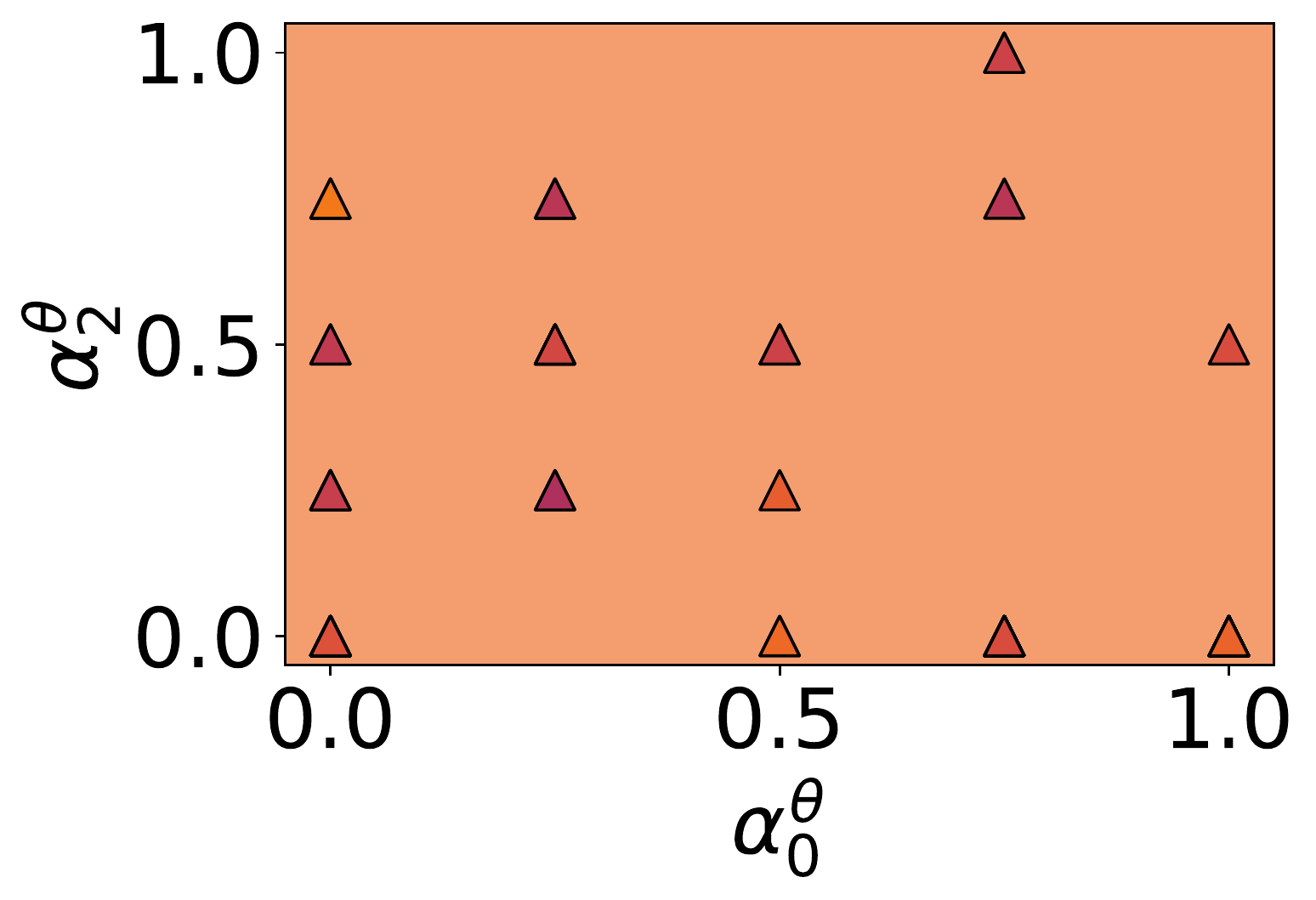}
        \includegraphics[width=0.3\textwidth]{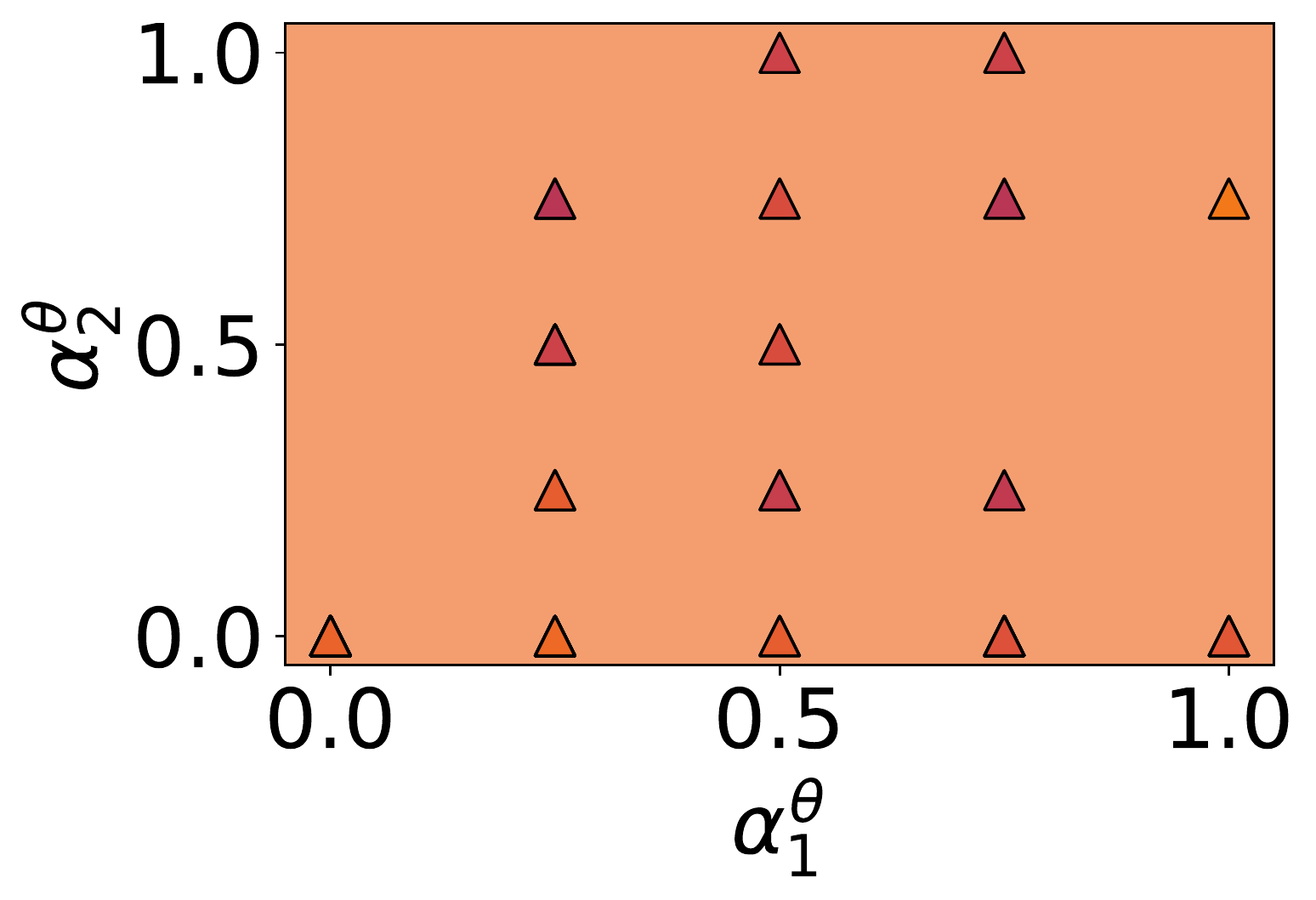}
         \caption{$\beta=1.0$}
     \end{subfigure}
	\end{minipage}
	\caption{
	\textit{Change in parameter subspace dynamics: }
	3D plot of interpolated input space (left) and interpolated parameter space (right) for parameter spaces trained with $\beta=0.0$ (a) and $\beta=1.0$ (b).
	The cross-sectional diagrams reflect the corresponding points in 2D.
	All colours correspond to the color bar in the figure. 
	We only plot the first-instance (i.e. there may be multiple) lowest-loss / maximum-accuracy mappable parameter in the parameter subspace for each interpolated task.
	}
	\label{tab:space_plot}
\end{table*}

\begin{figure*}[h]
    \renewcommand{\fnum@table}{Appendix~A.3.\thetable}
    \renewcommand{\fnum@figure}{Appendix~A.3.\thefigure}
    \setcounter{figure}{0}
    \setcounter{table}{0}
    \centering
    \includegraphics[width=0.245\textwidth]{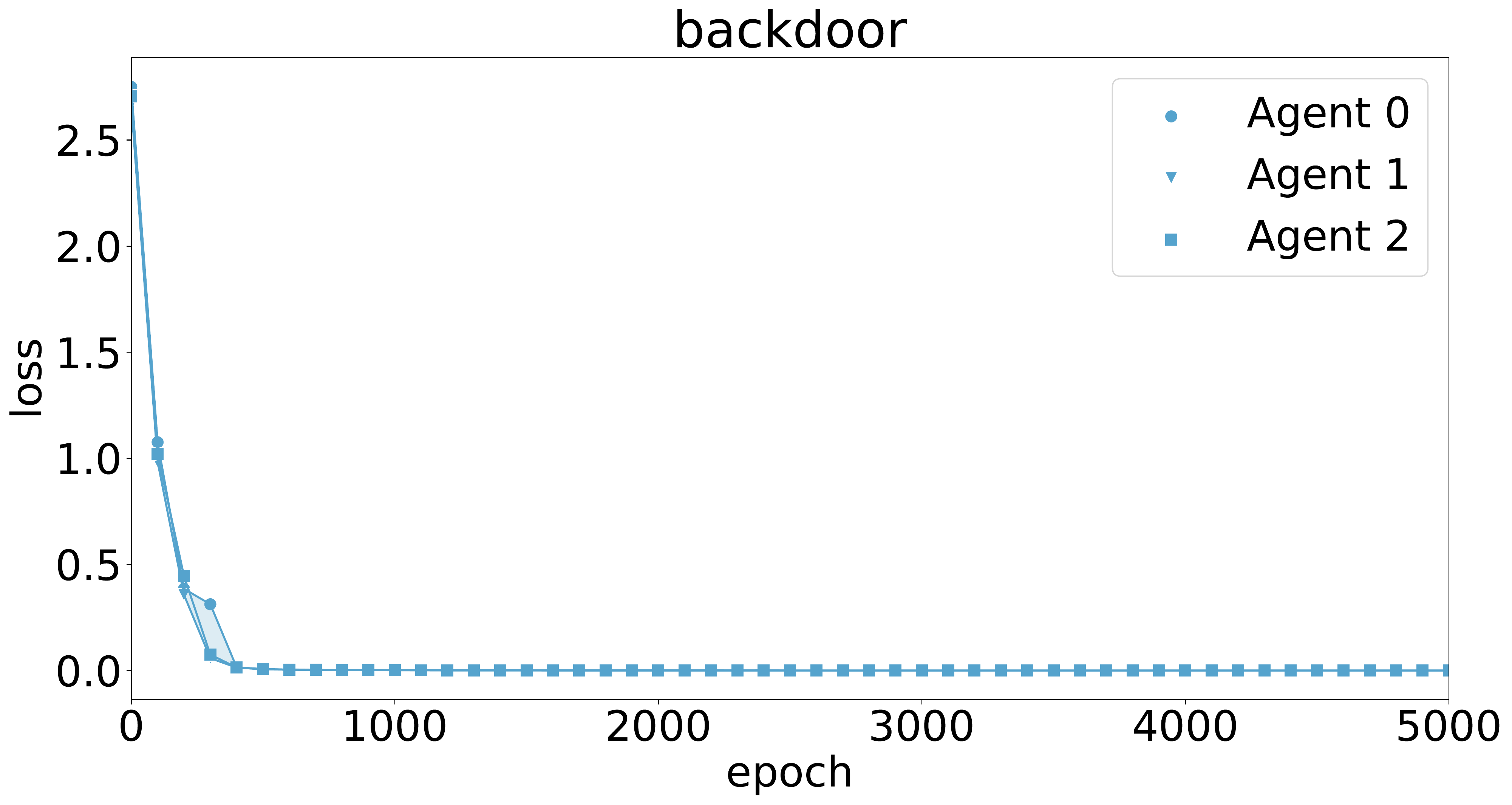}
    \includegraphics[width=0.245\textwidth]{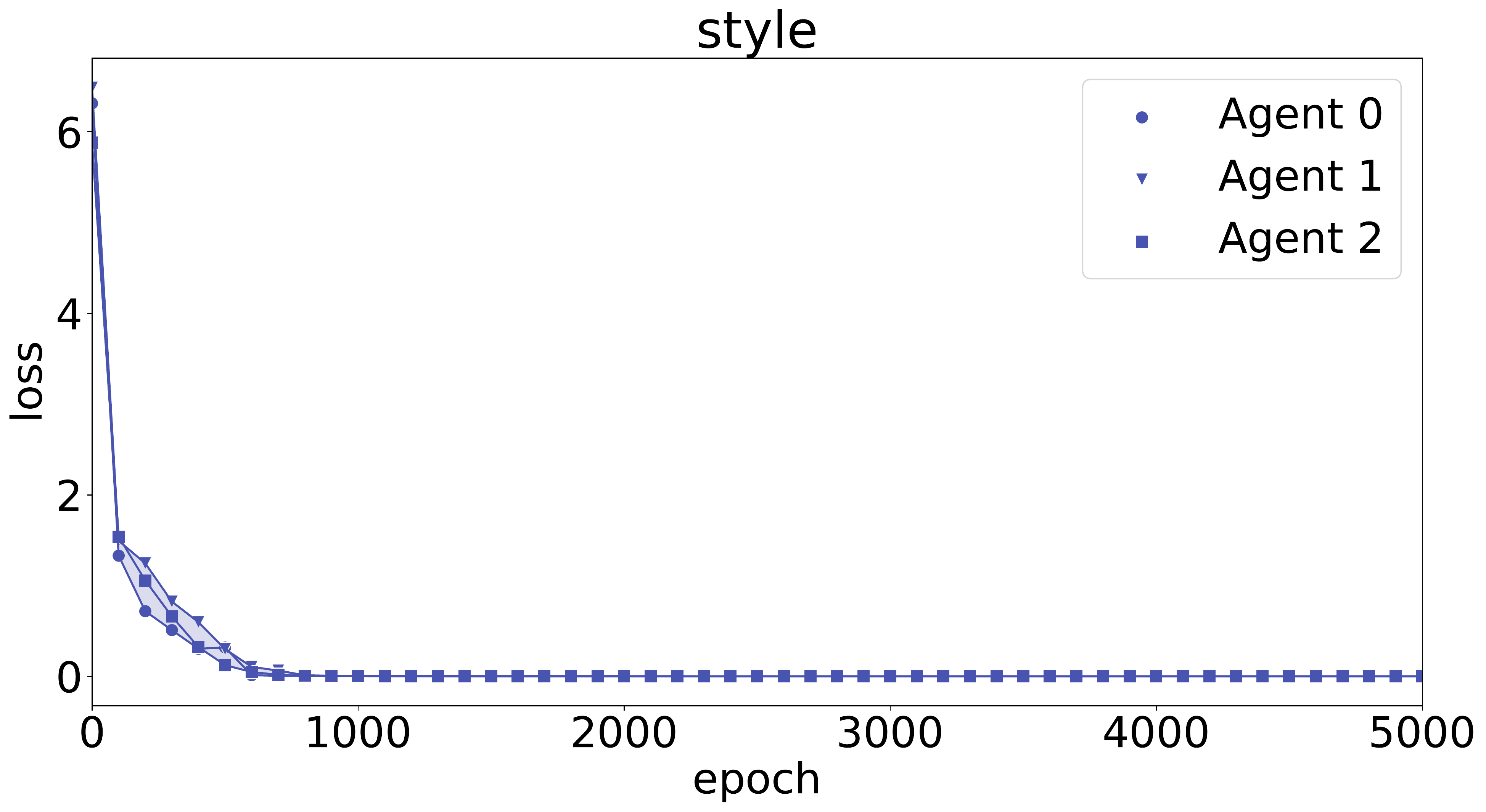}
    \includegraphics[width=0.245\textwidth]{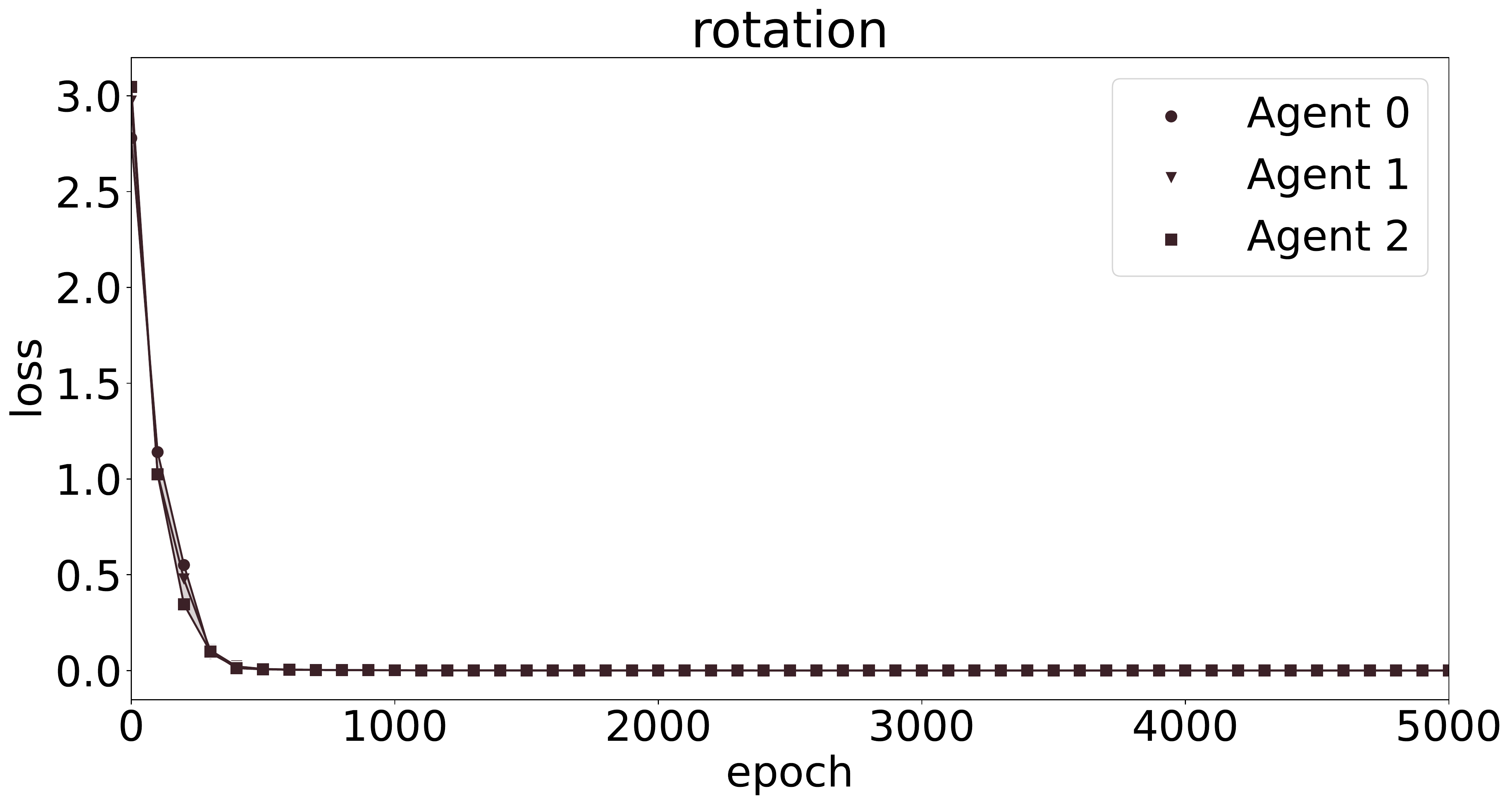}
    \includegraphics[width=0.245\textwidth]{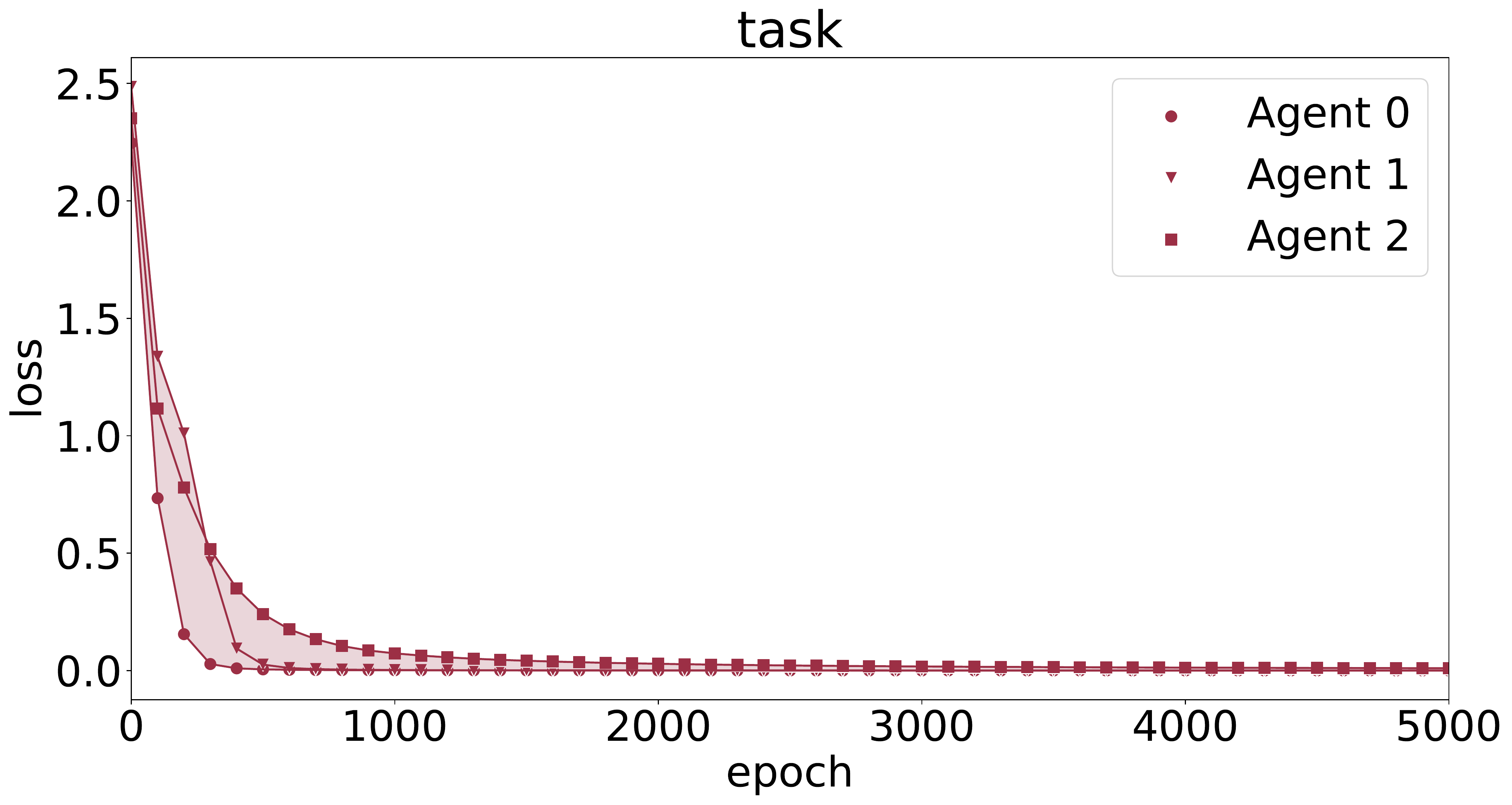}
    \includegraphics[width=0.245\textwidth]{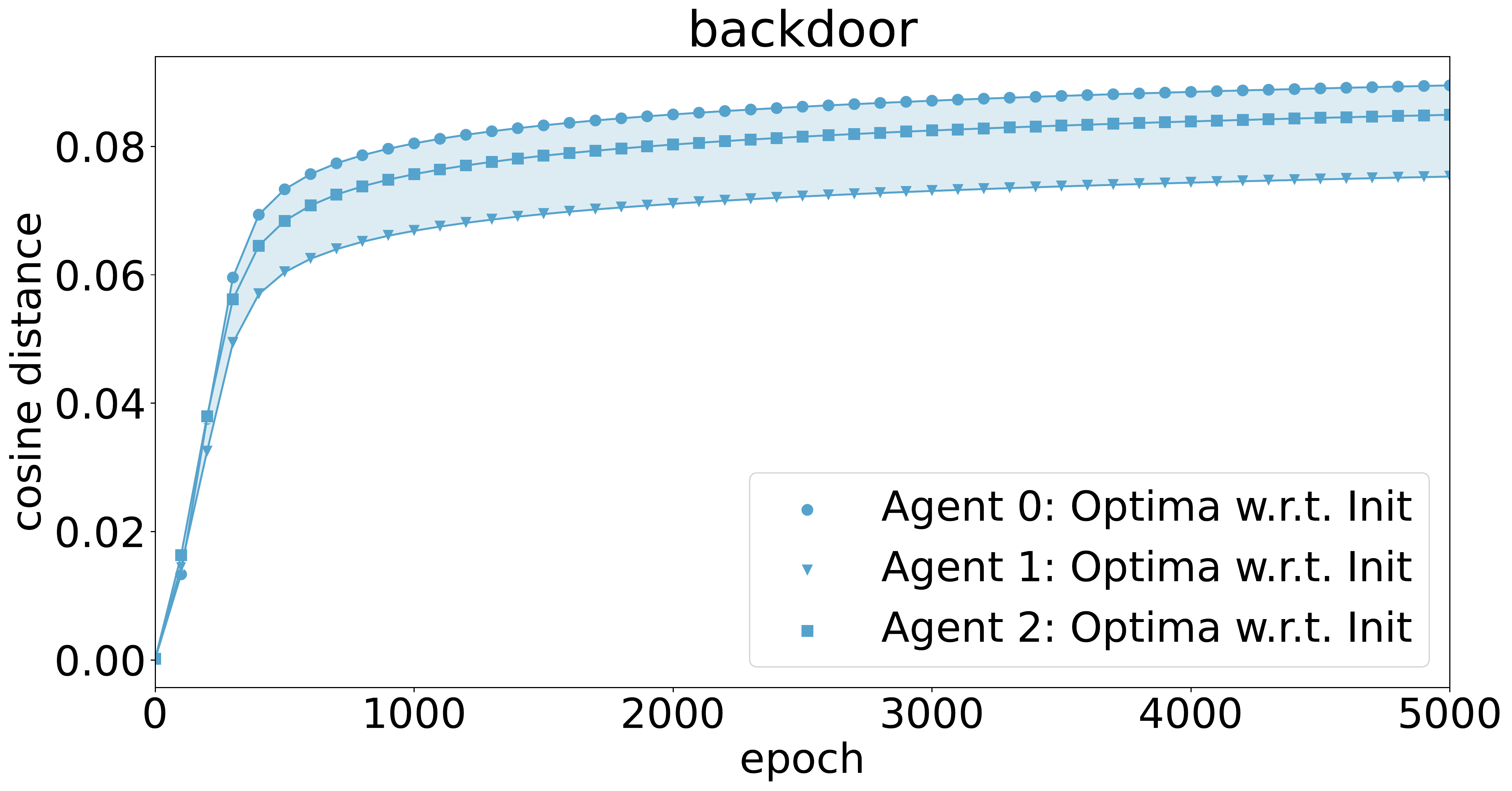}
    \includegraphics[width=0.245\textwidth]{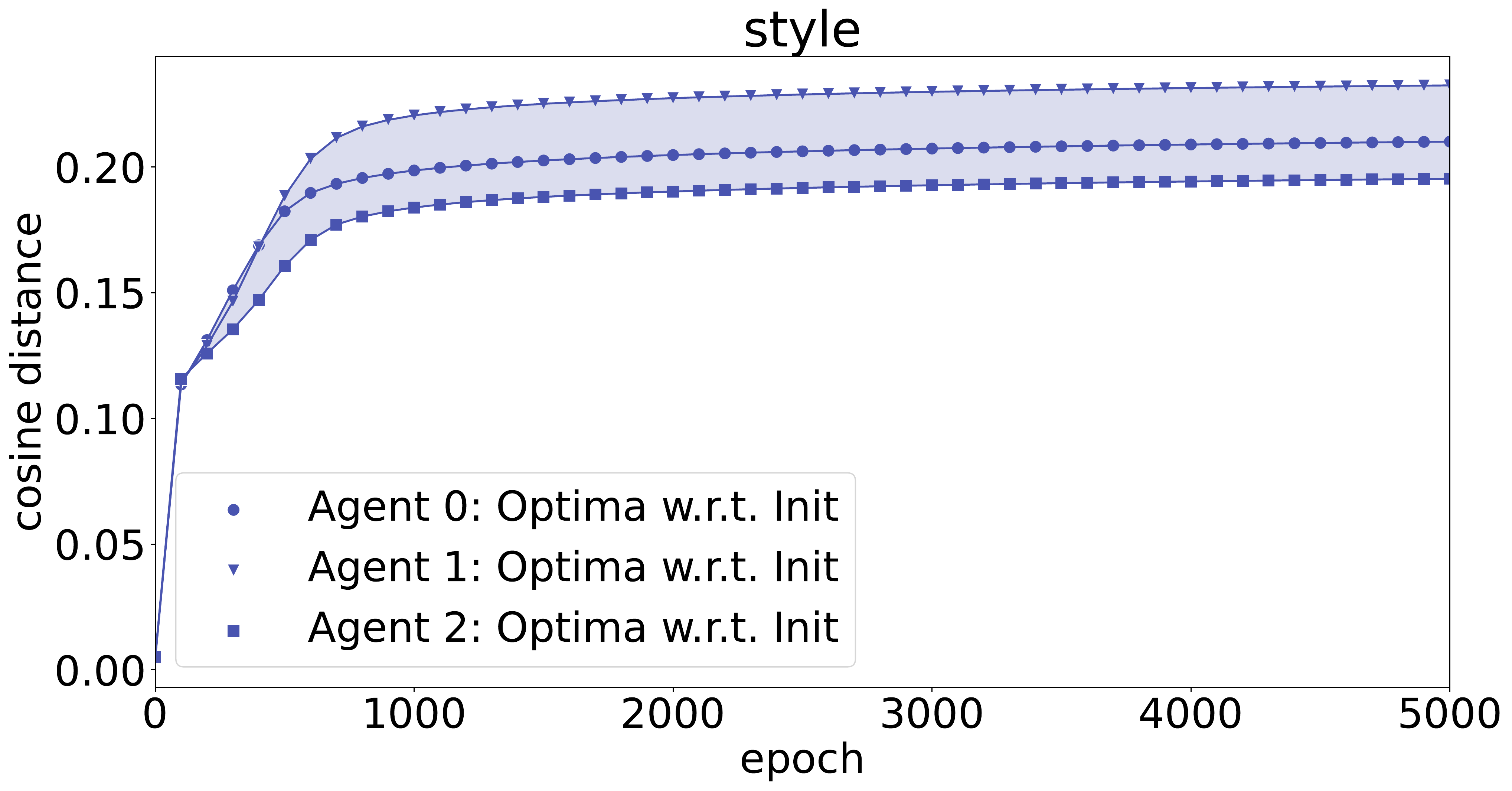}
    \includegraphics[width=0.245\textwidth]{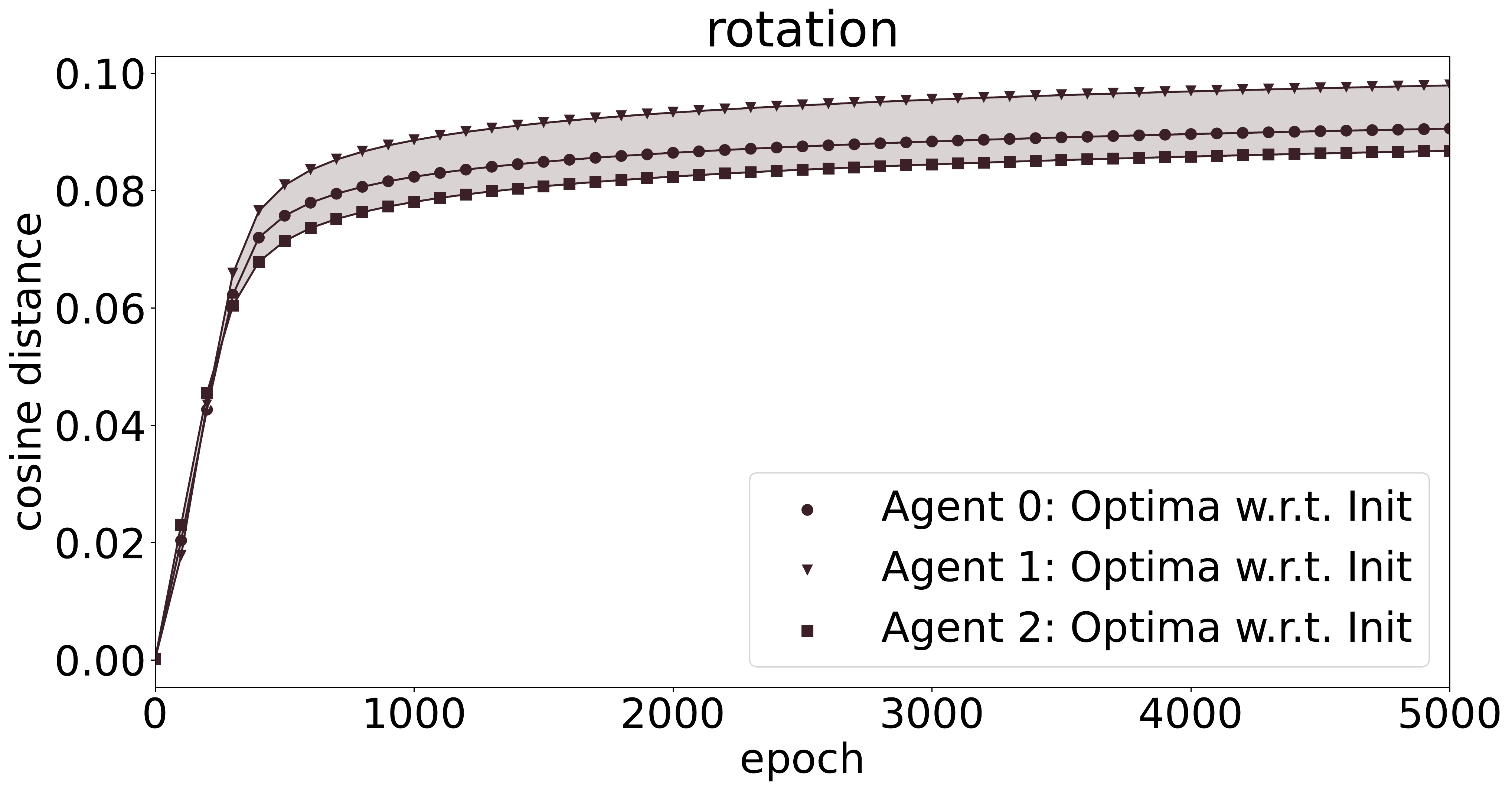}
    \includegraphics[width=0.245\textwidth]{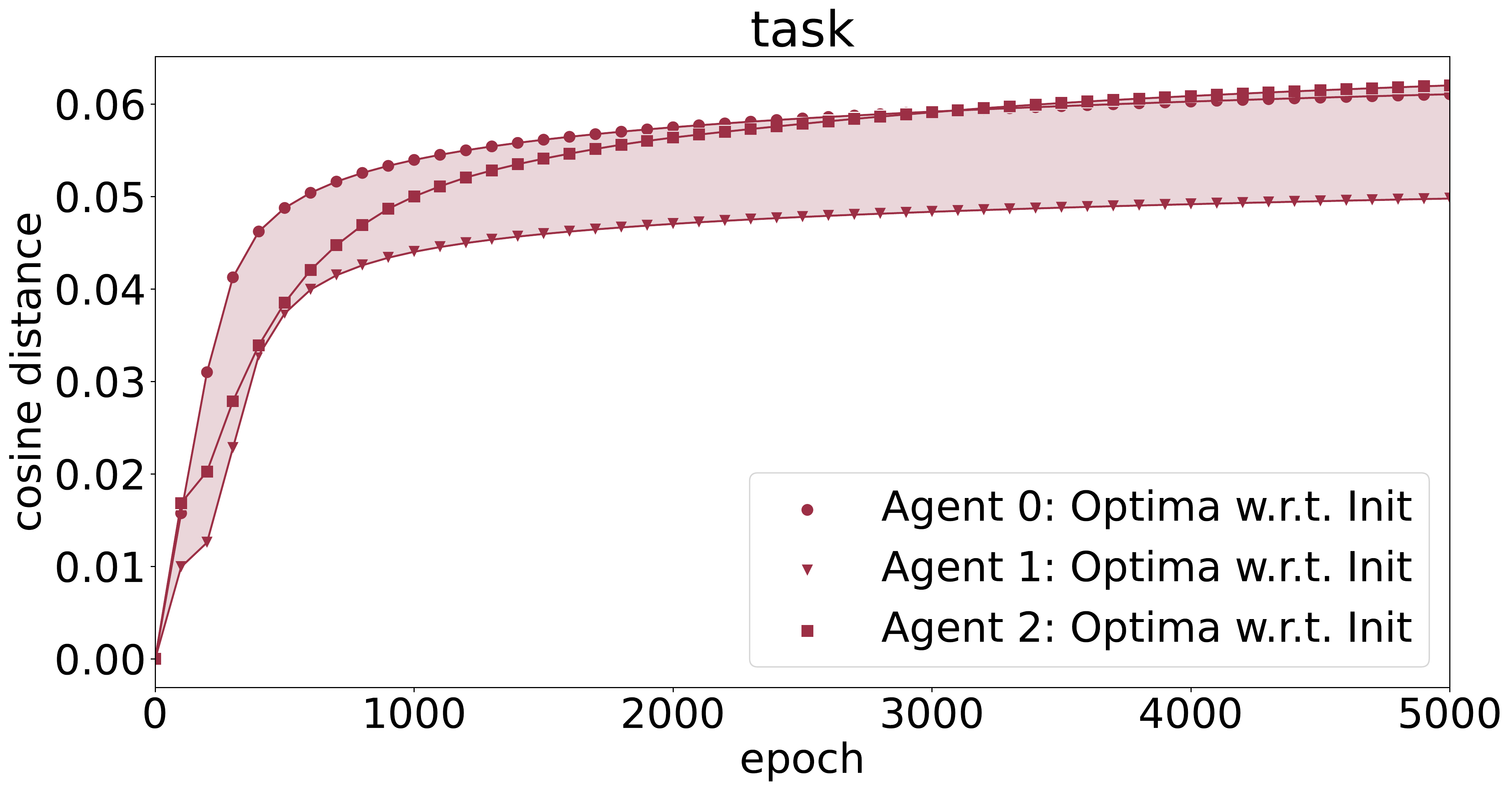}
    \includegraphics[width=0.245\textwidth]{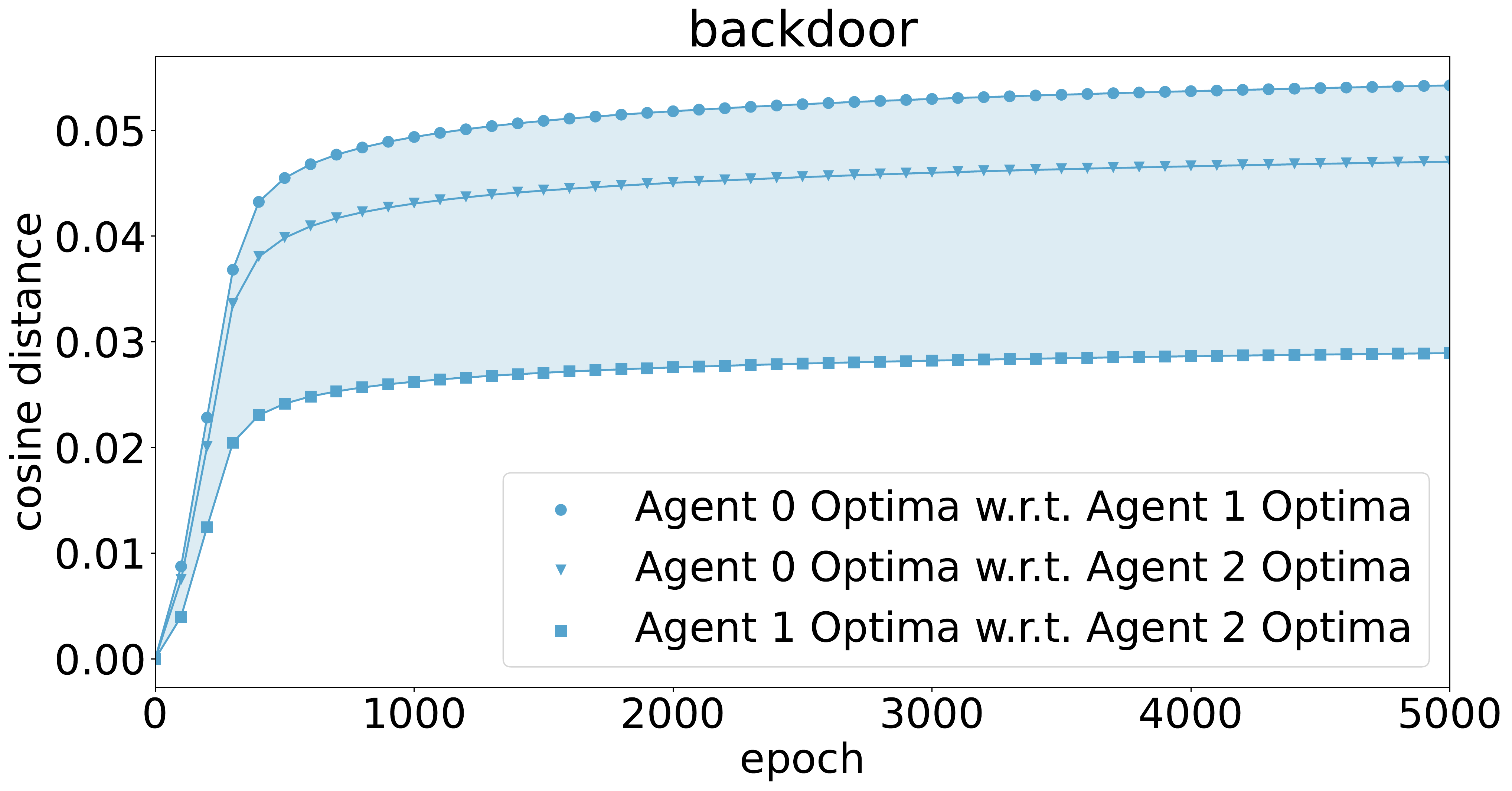}
    \includegraphics[width=0.245\textwidth]{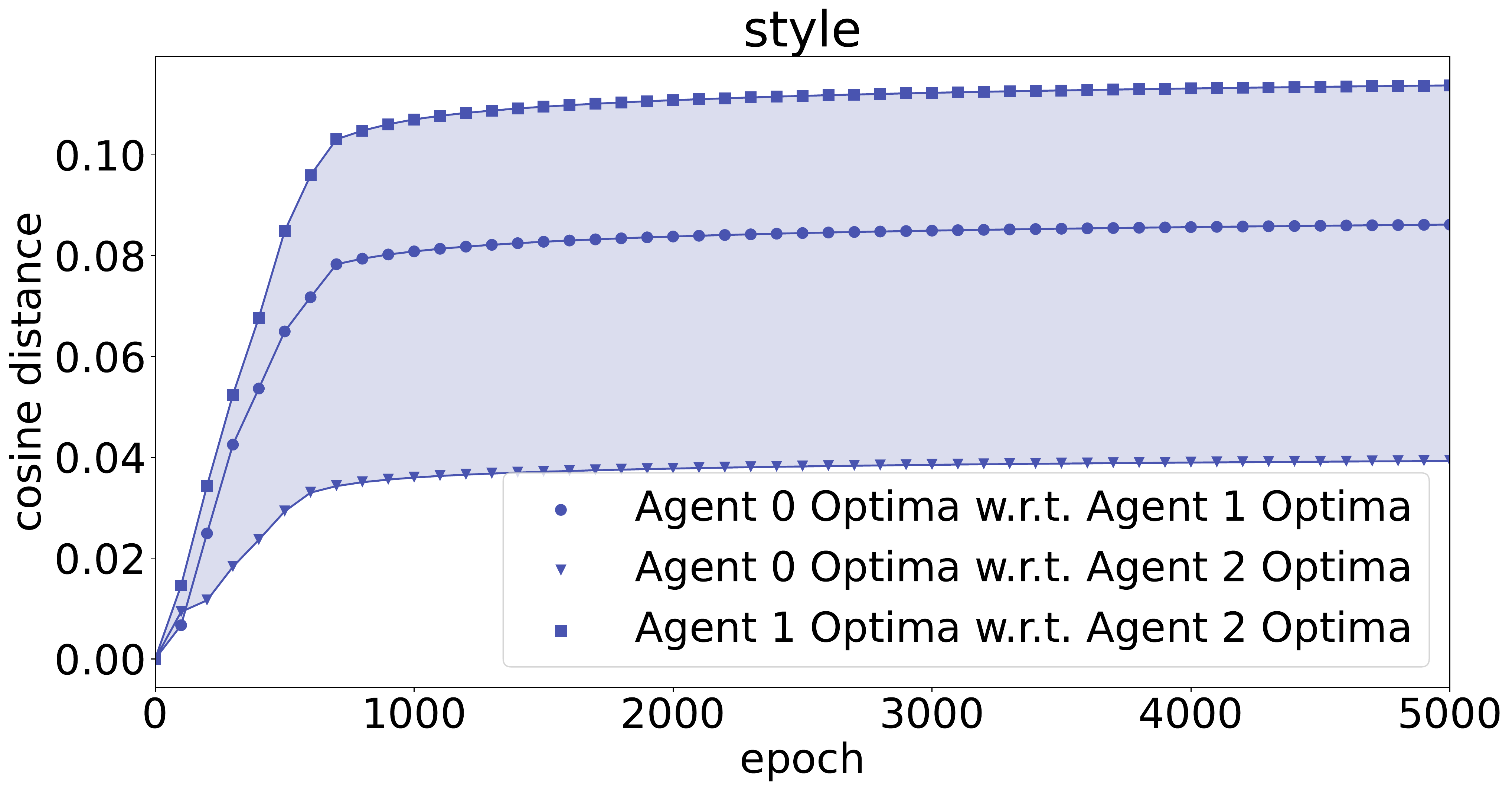}
    \includegraphics[width=0.245\textwidth]{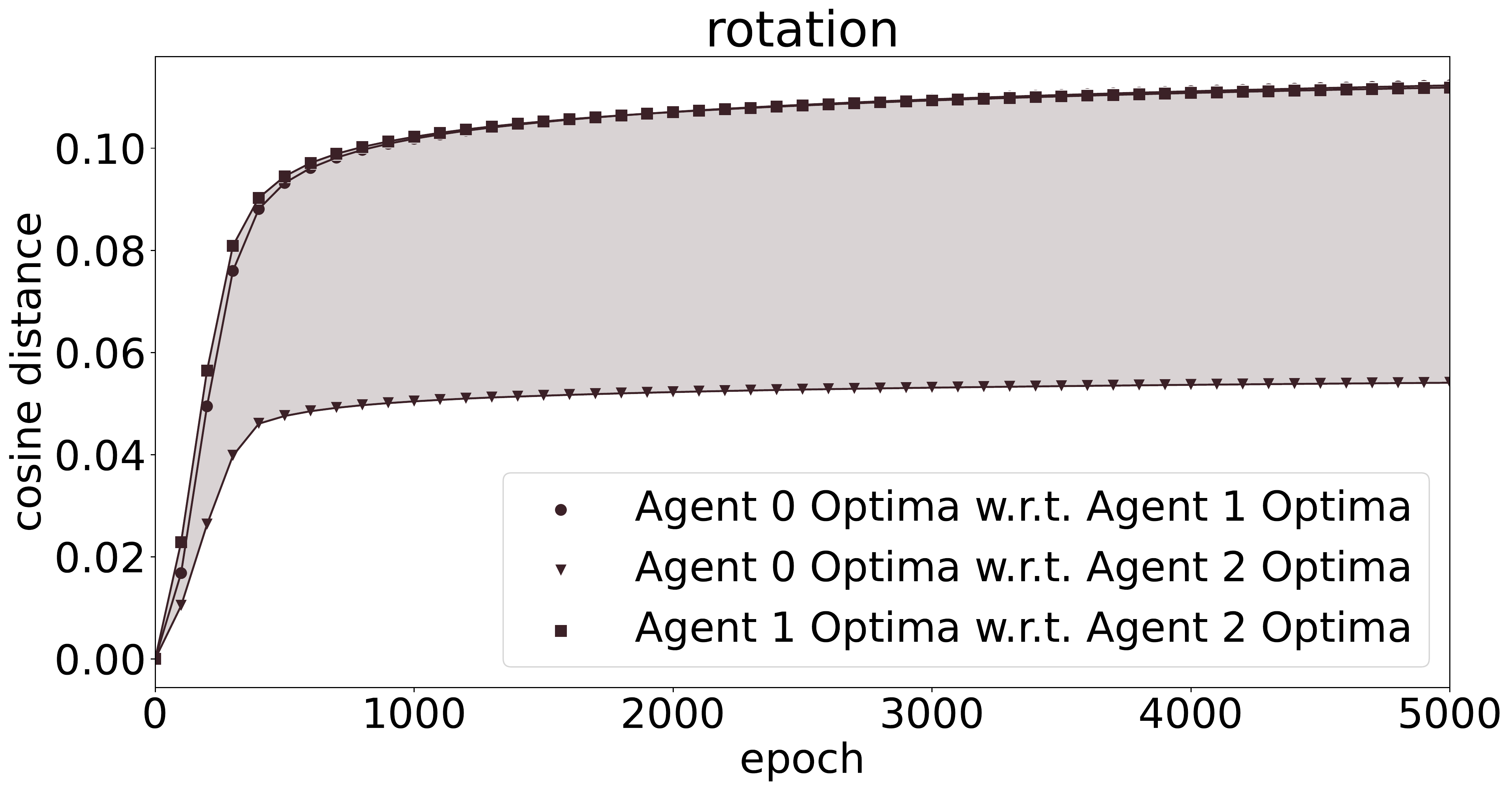}
    \includegraphics[width=0.245\textwidth]{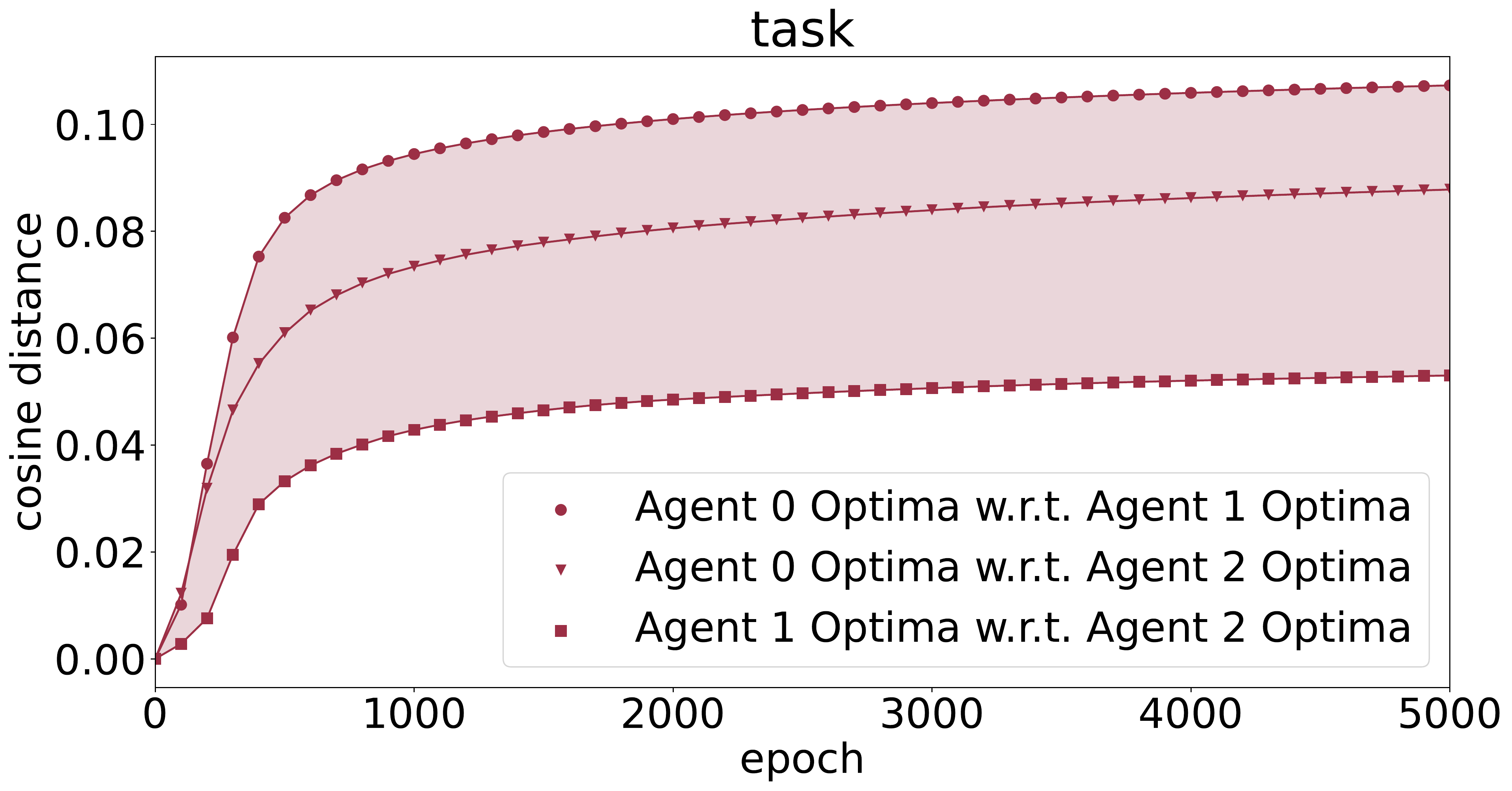}
    \includegraphics[width=0.245\textwidth]{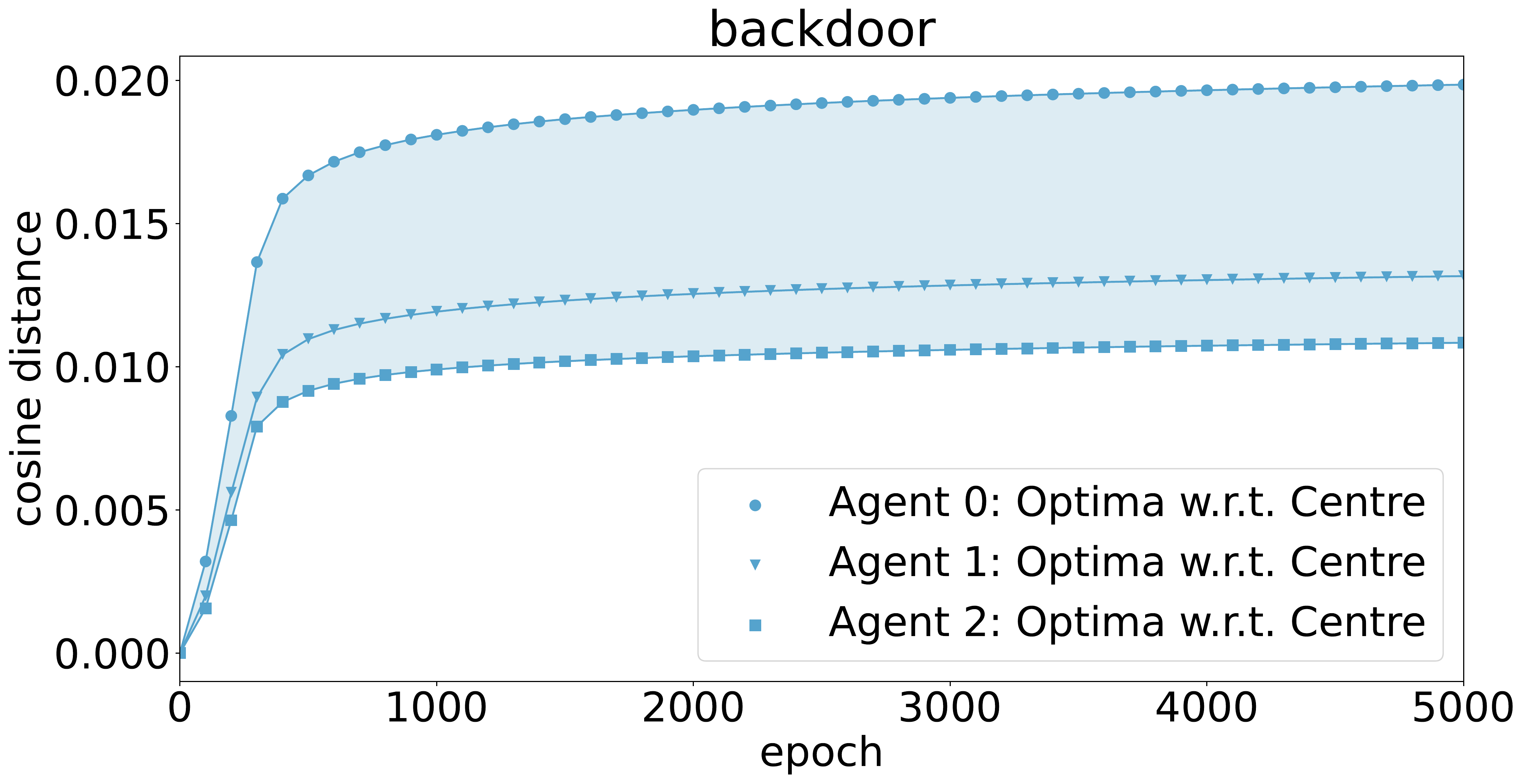}
    \includegraphics[width=0.245\textwidth]{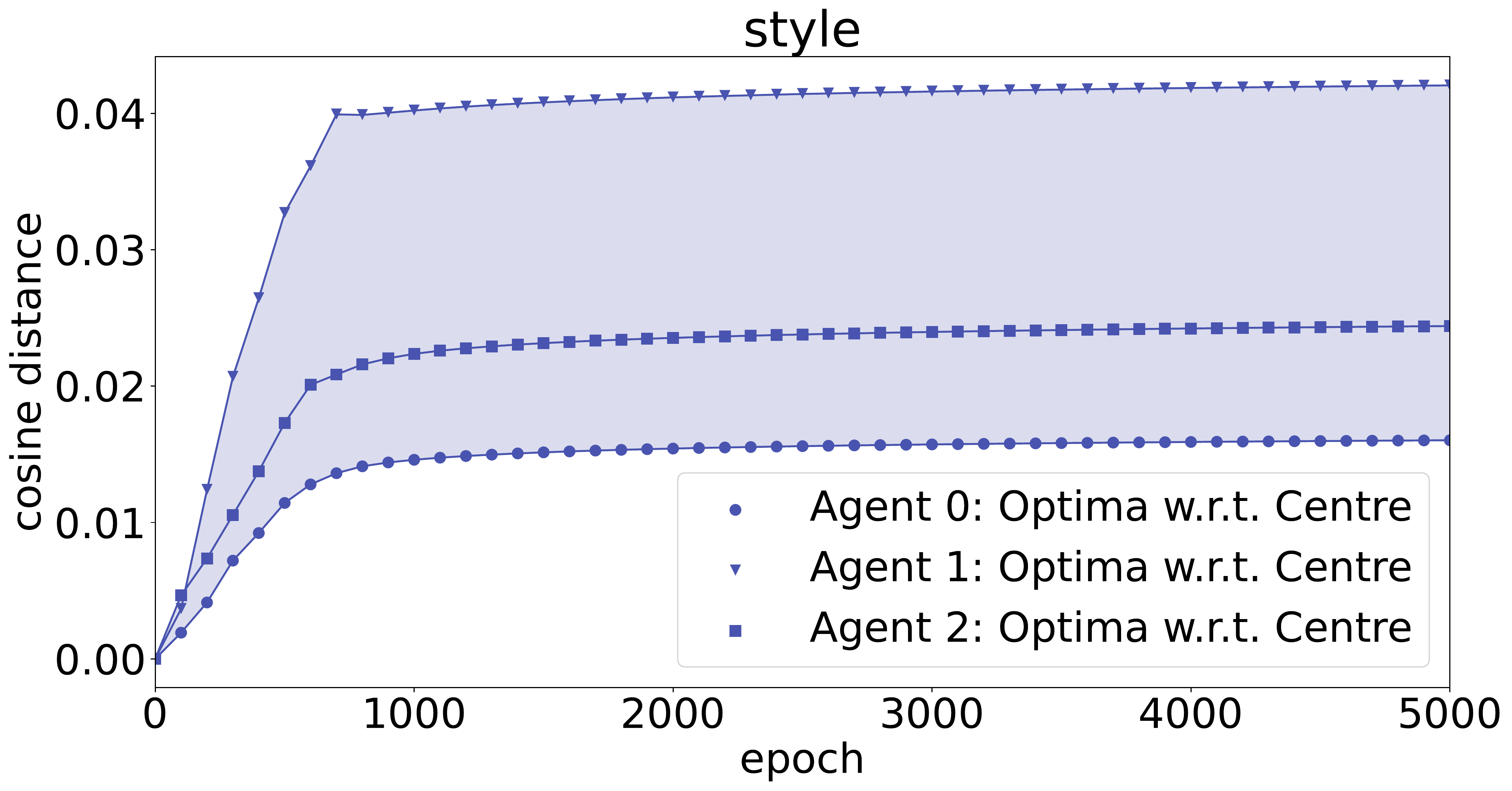}
    \includegraphics[width=0.245\textwidth]{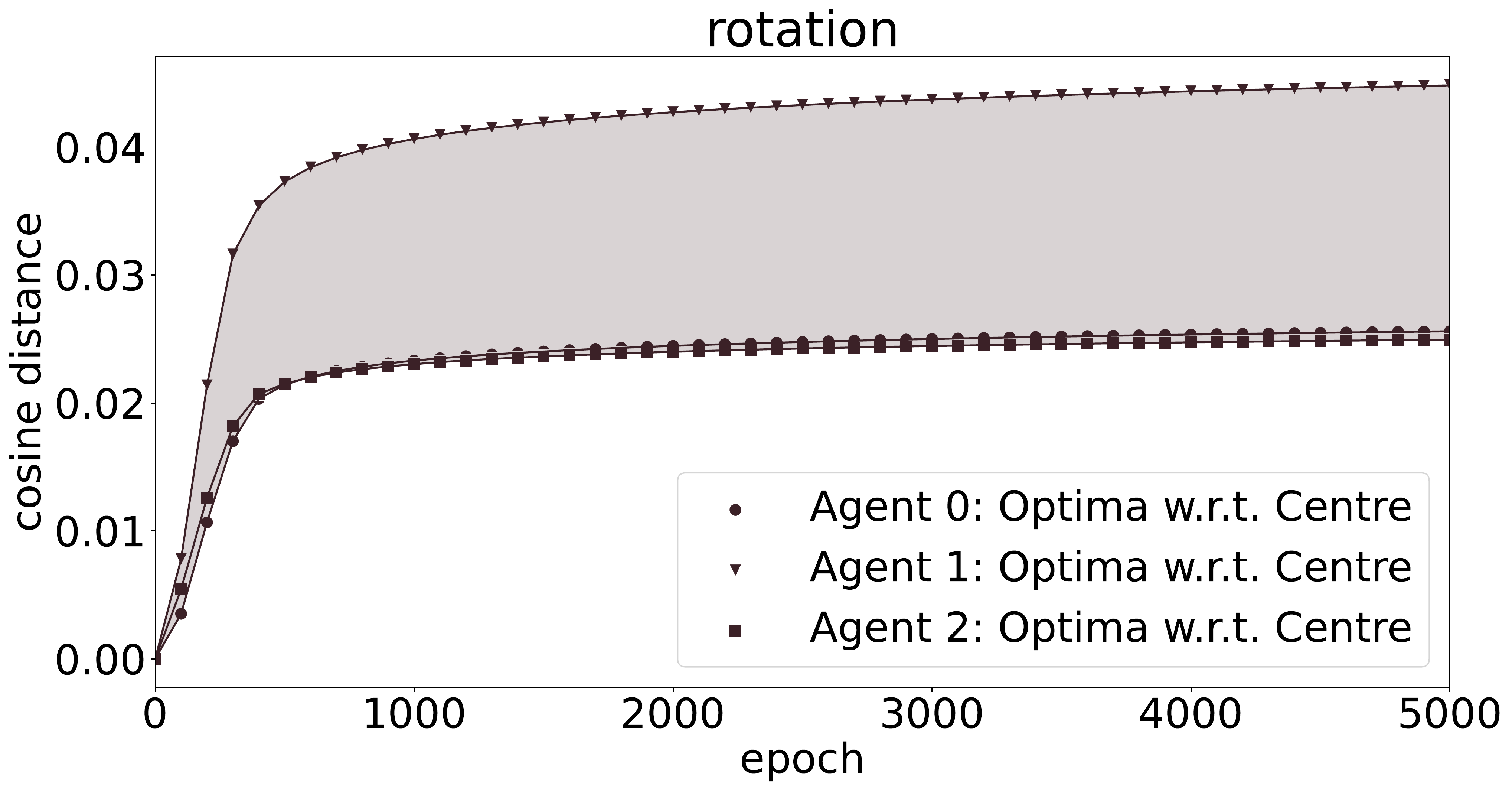}
    \includegraphics[width=0.245\textwidth]{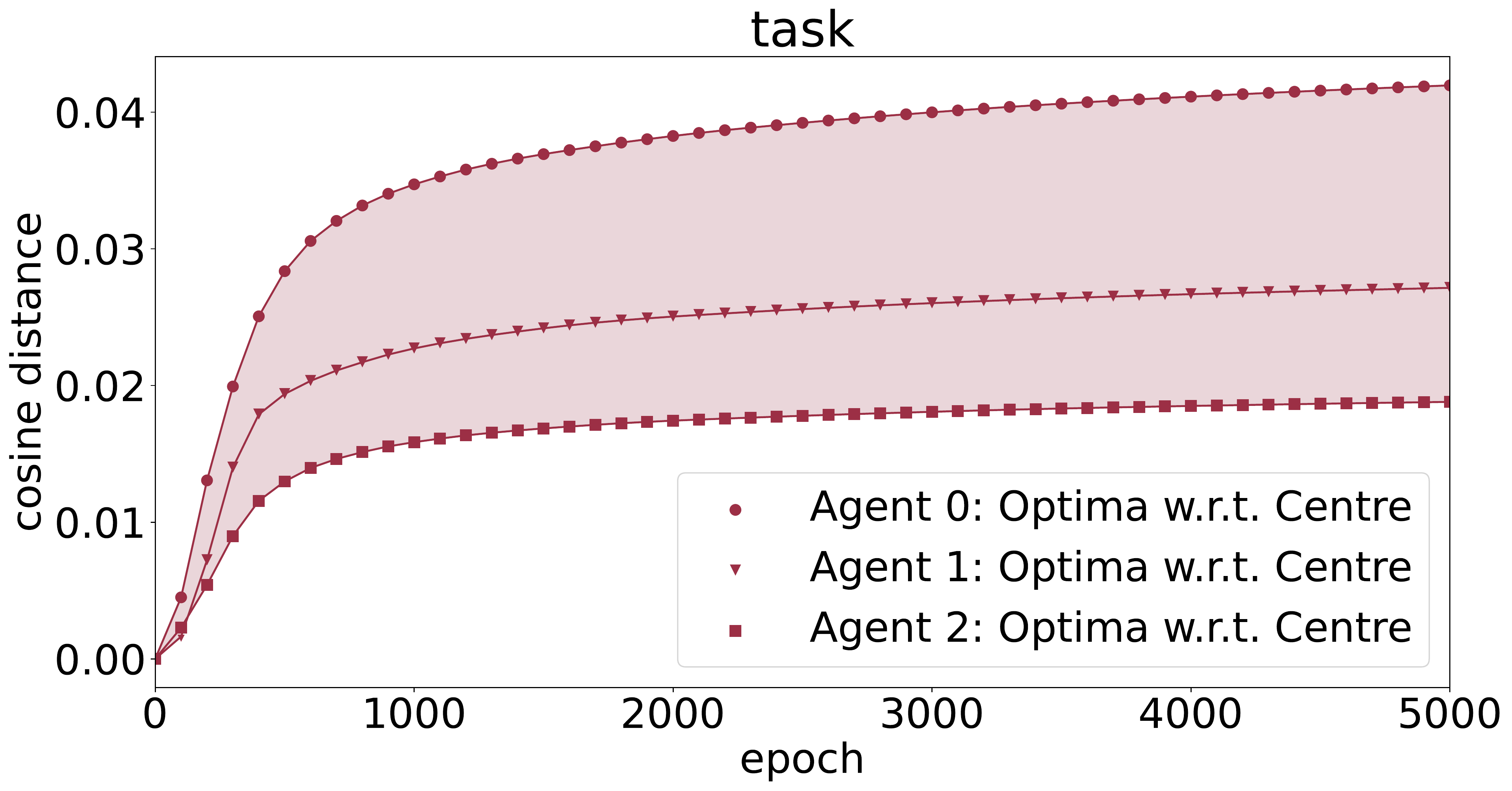}
    \caption{
    \textit{Change in parameter subspace dynamics: }
    Loss and cosine distance per epoch during training of CPS with 3 train-time distributions ($\beta=0.0$).
    Perturbation type of train-time subsets are
    backdoor (column 1), stylization (column 2), rotation (column 3), and task (column 4).
    }
\end{figure*}

\begin{figure*}[h]
\renewcommand{\fnum@table}{Appendix~A.3.\thetable}
\renewcommand{\fnum@figure}{Appendix~A.3.\thefigure}
\setcounter{figure}{1}
\setcounter{table}{1}
    \centering
    \includegraphics[width=0.245\textwidth]{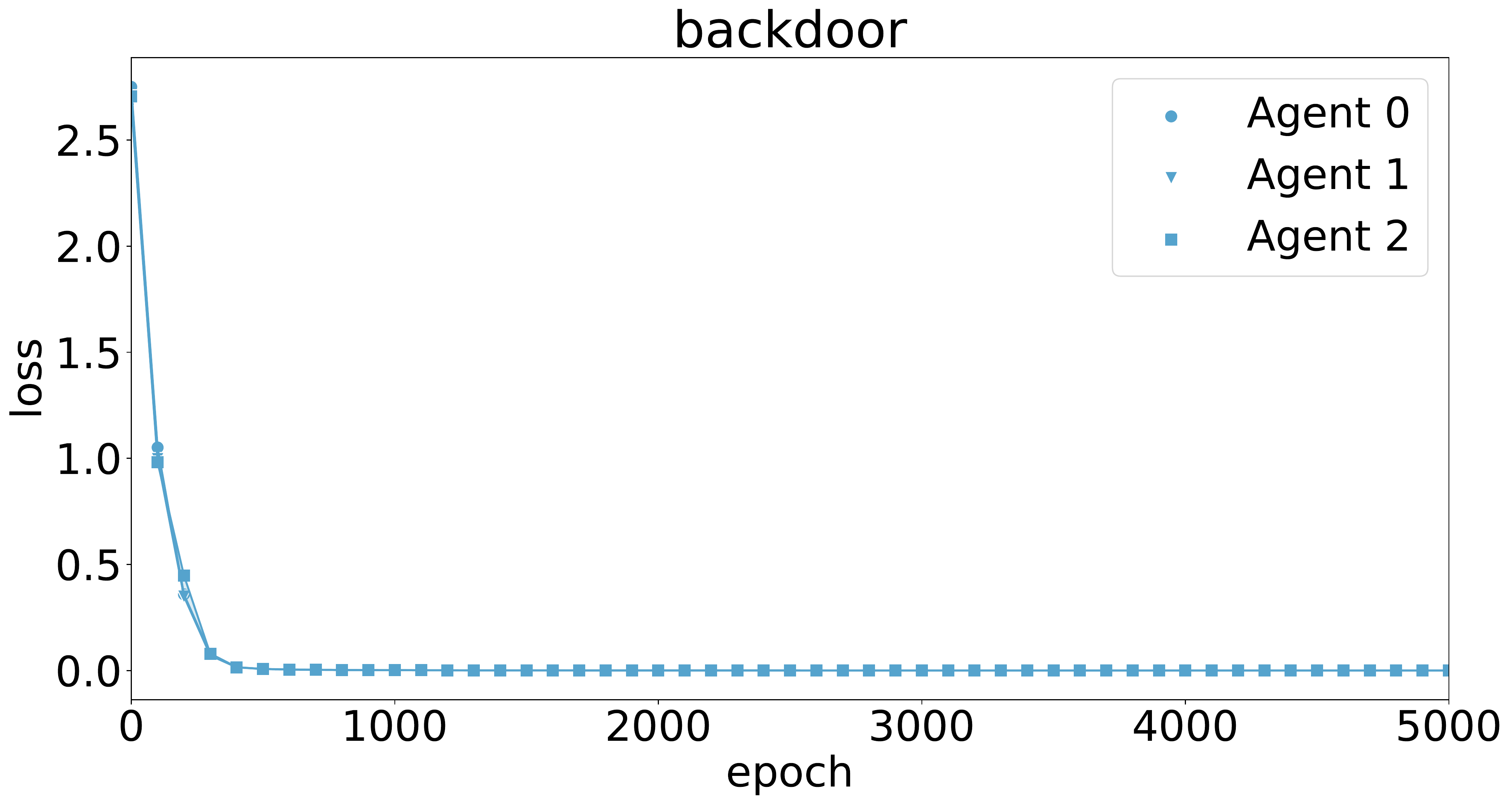}
    \includegraphics[width=0.245\textwidth]{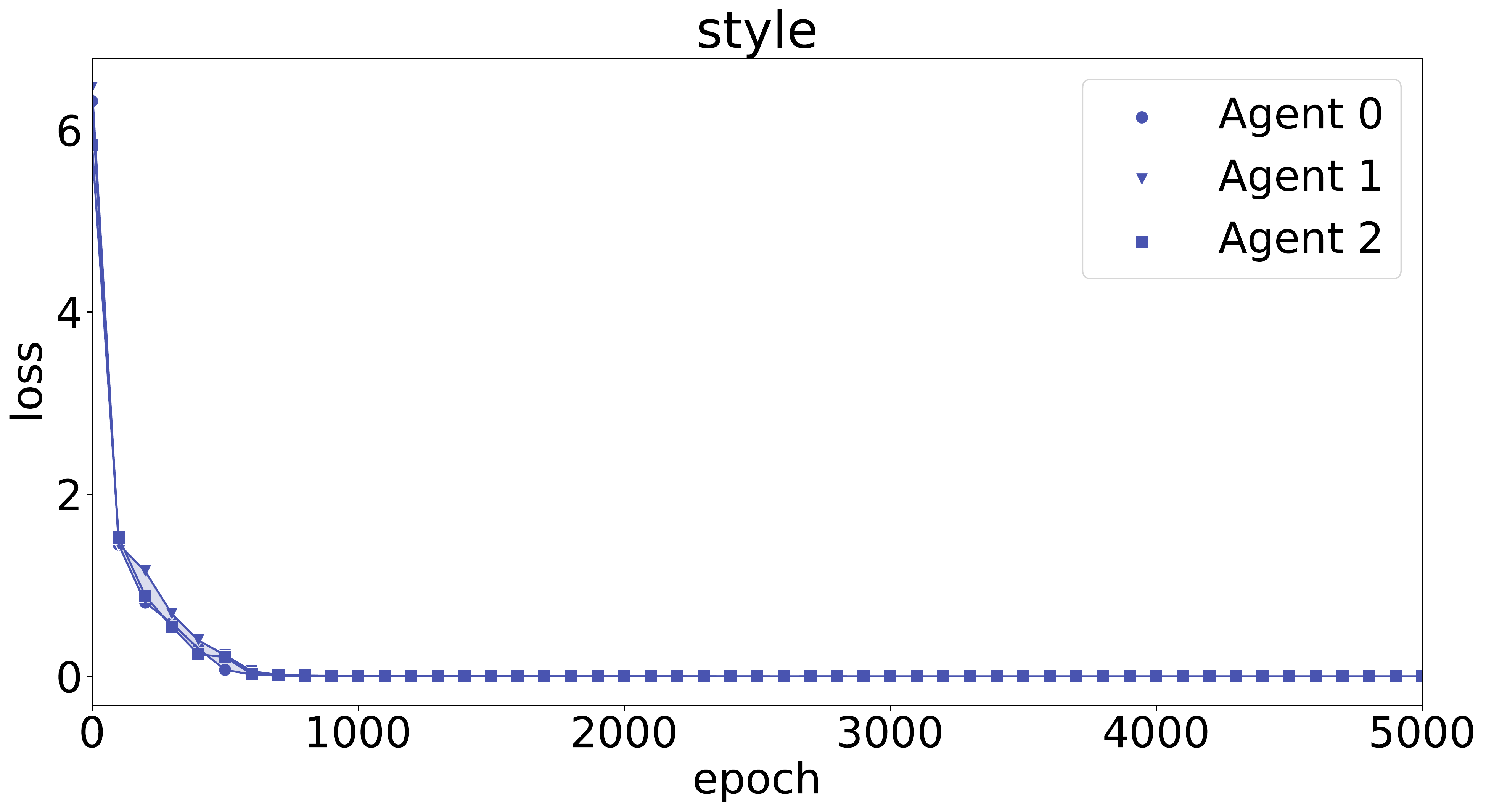}
    \includegraphics[width=0.245\textwidth]{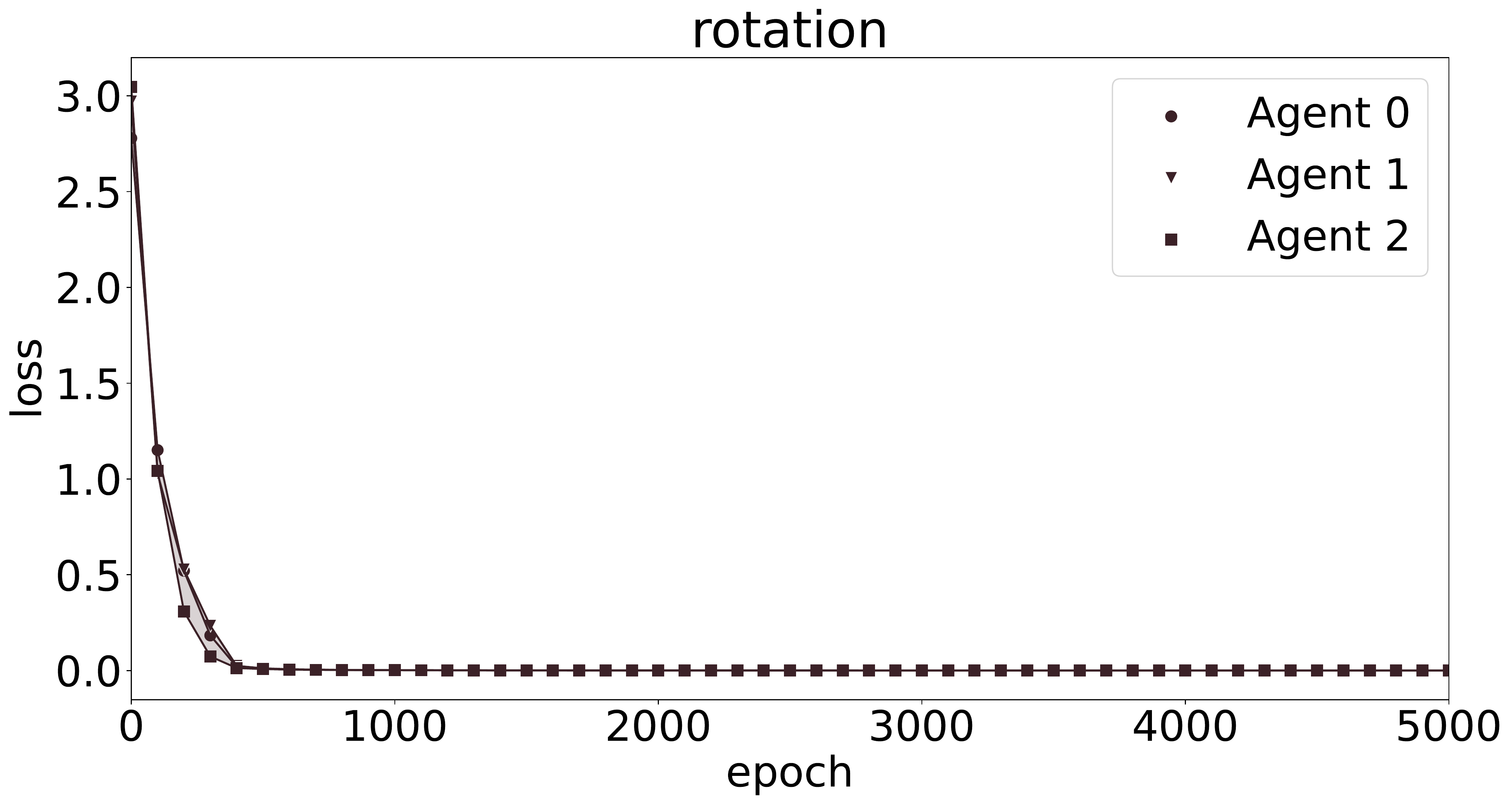}
    \includegraphics[width=0.245\textwidth]{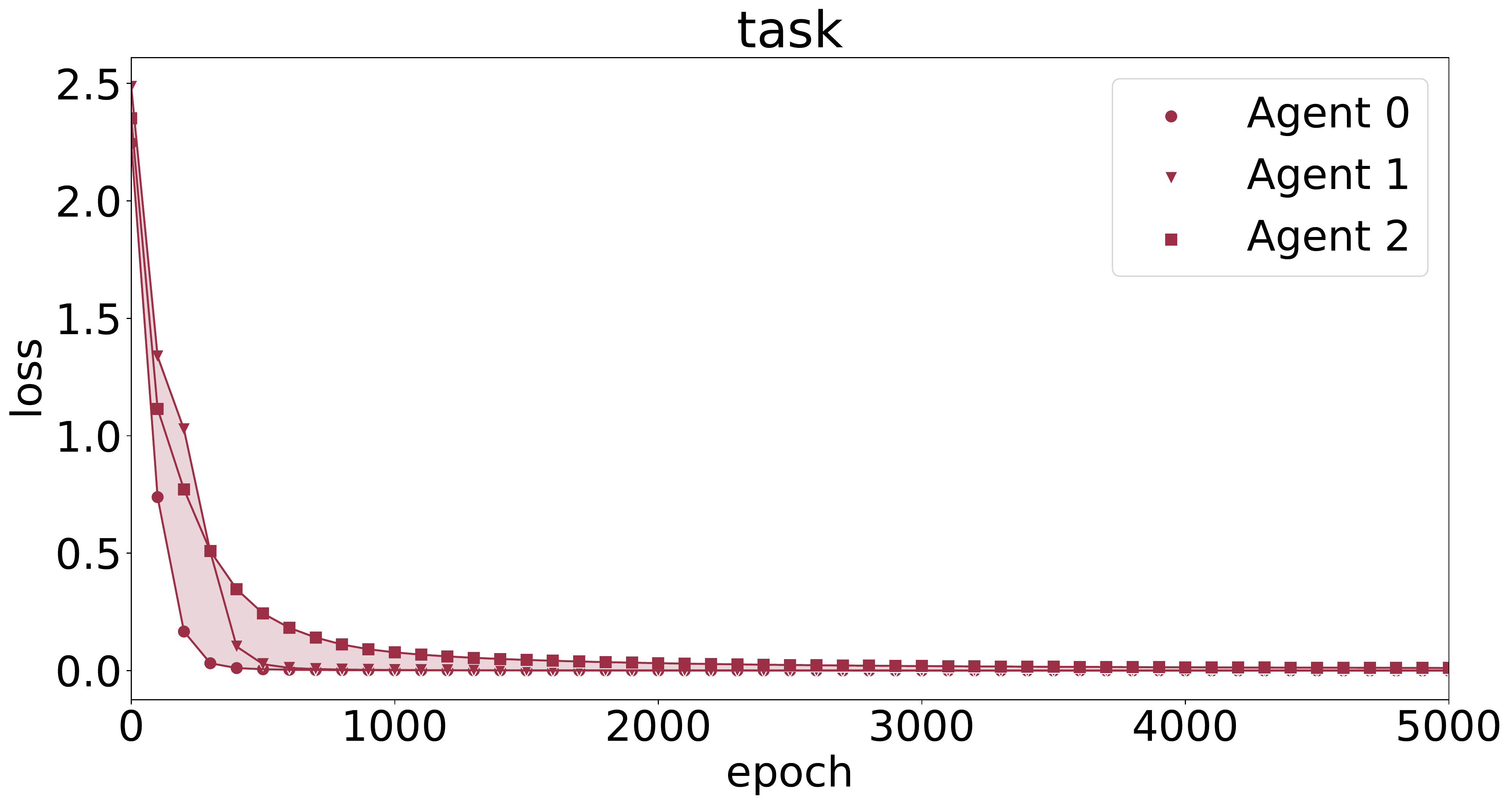}
    \includegraphics[width=0.245\textwidth]{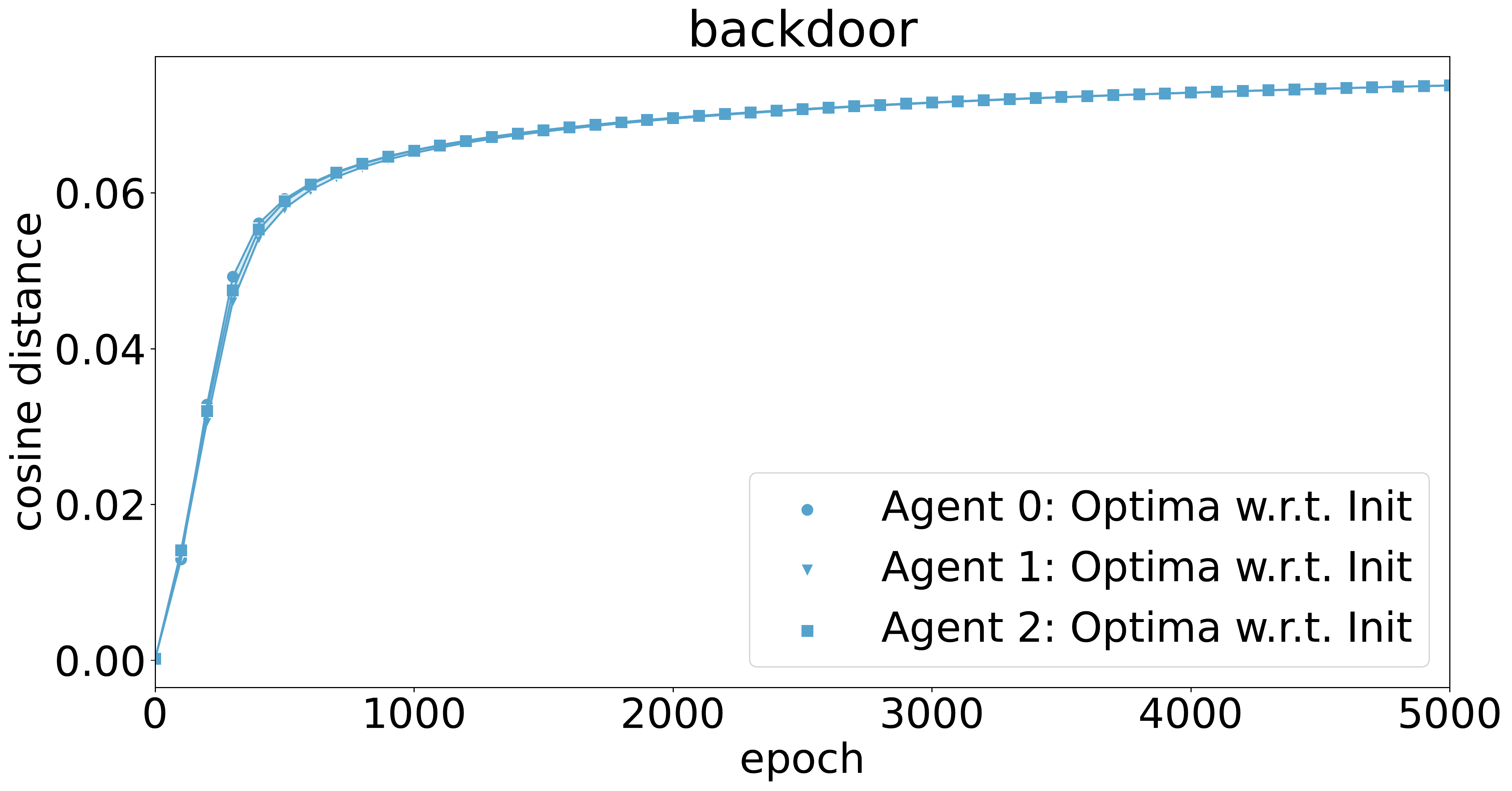}
    \includegraphics[width=0.245\textwidth]{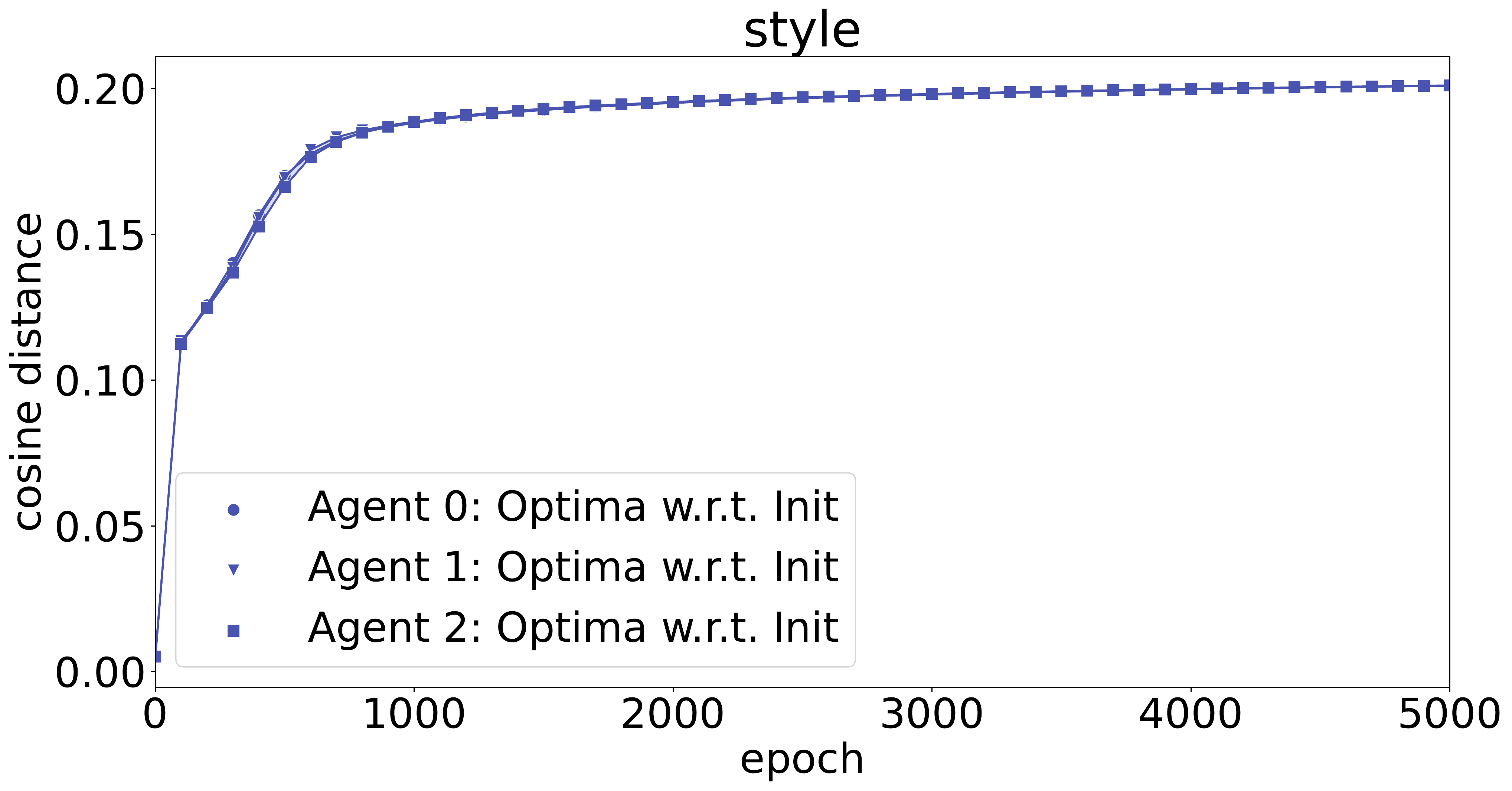}
    \includegraphics[width=0.245\textwidth]{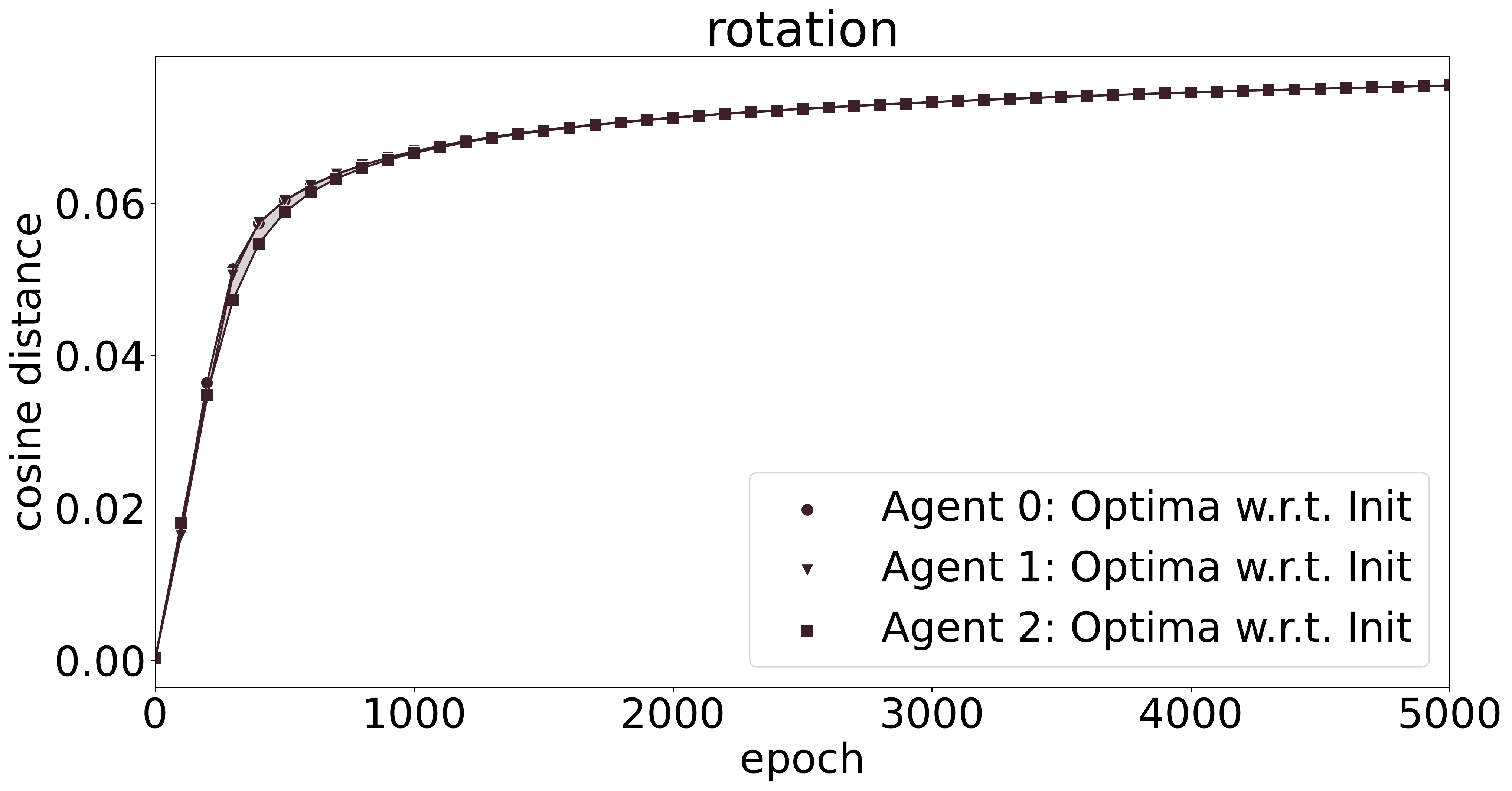}
    \includegraphics[width=0.245\textwidth]{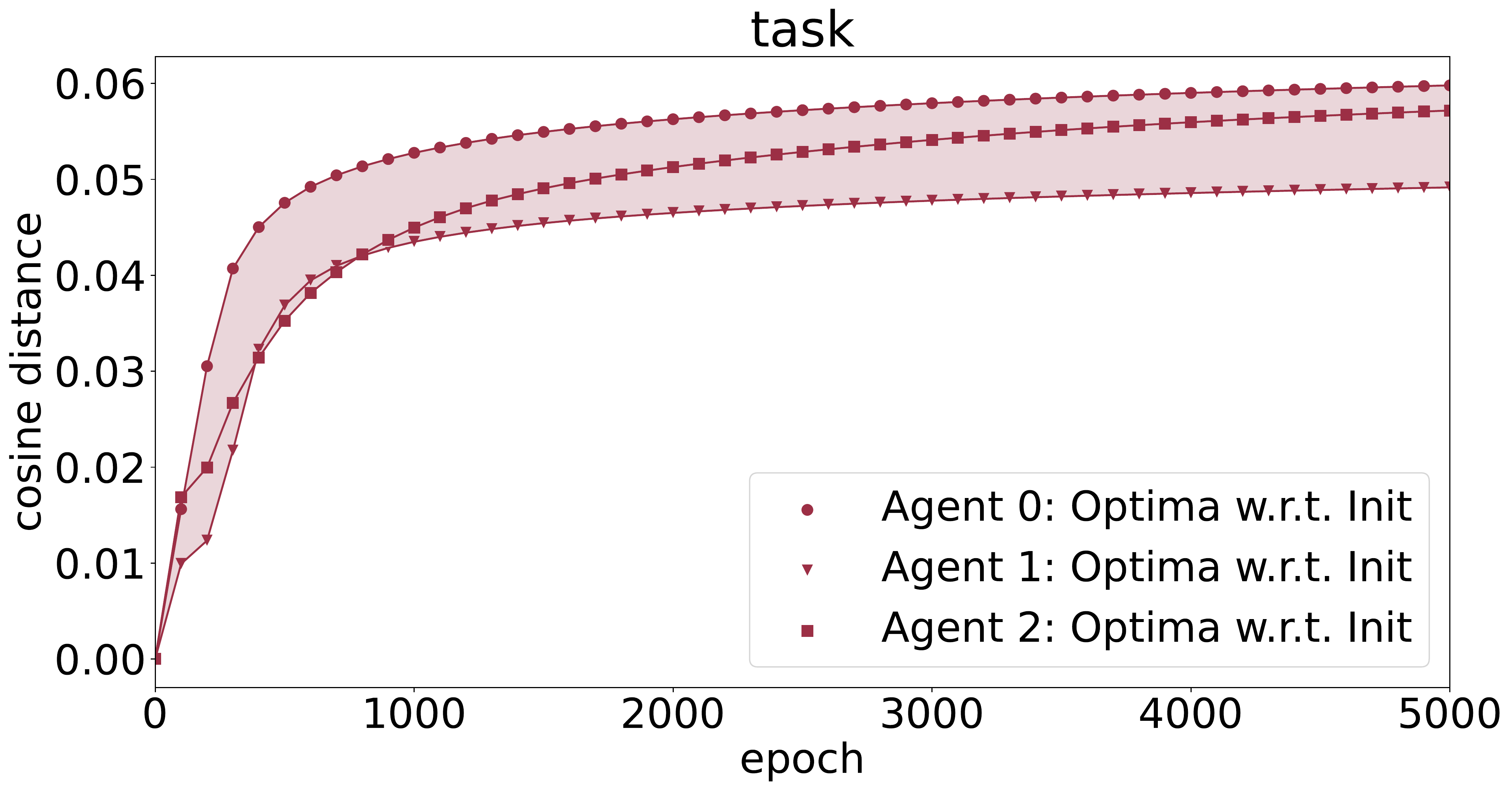}
    \includegraphics[width=0.245\textwidth]{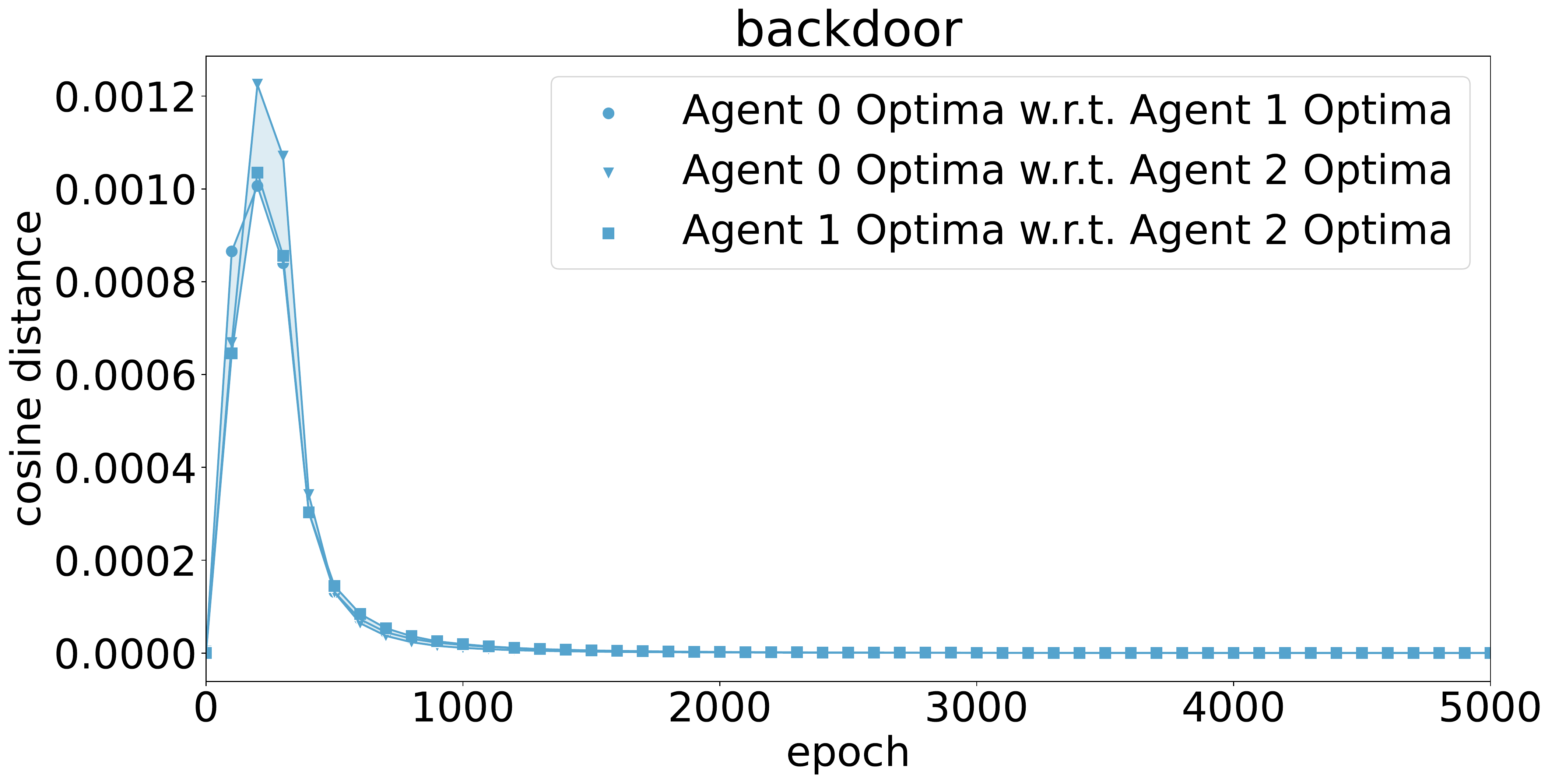}
    \includegraphics[width=0.245\textwidth]{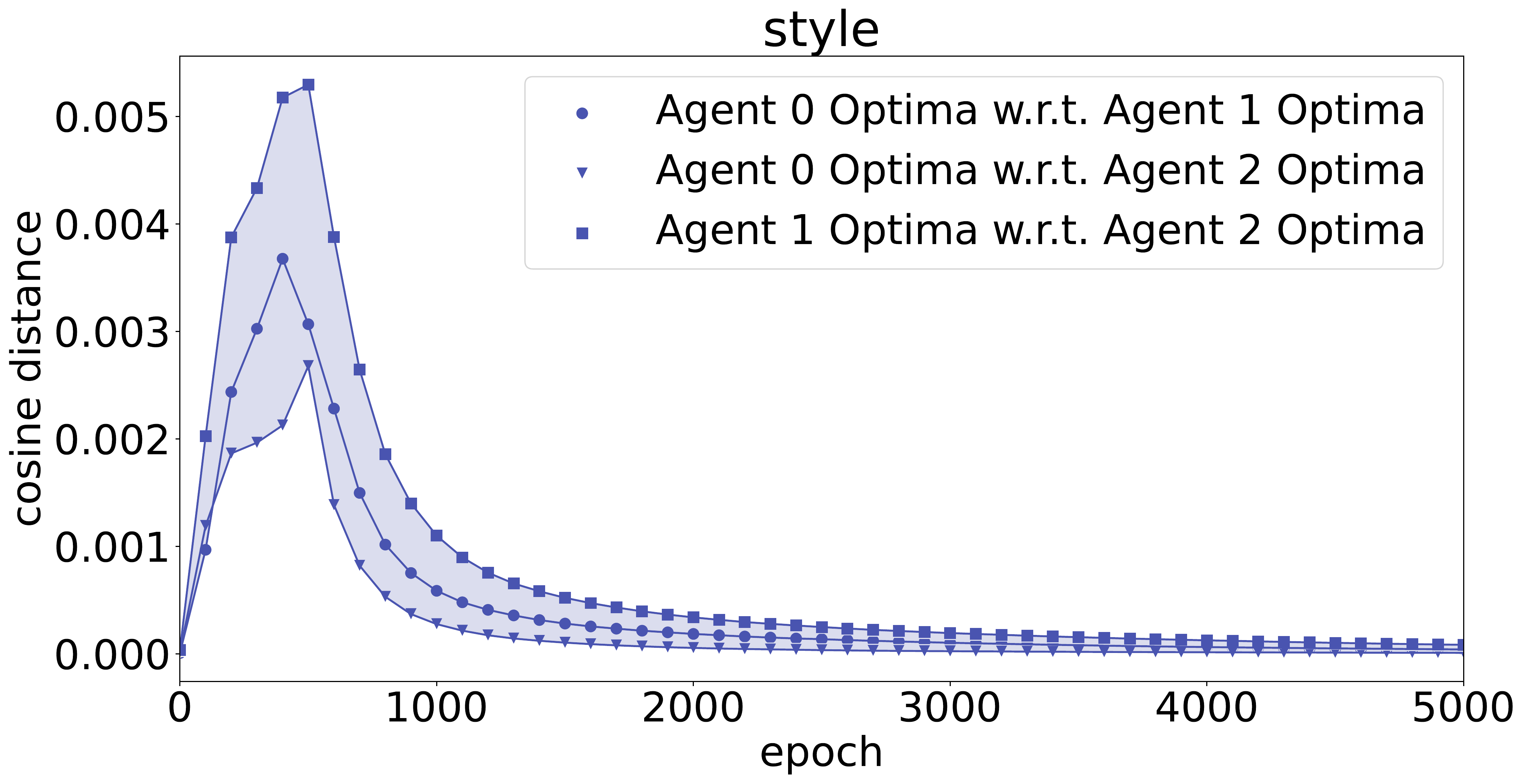}
    \includegraphics[width=0.245\textwidth]{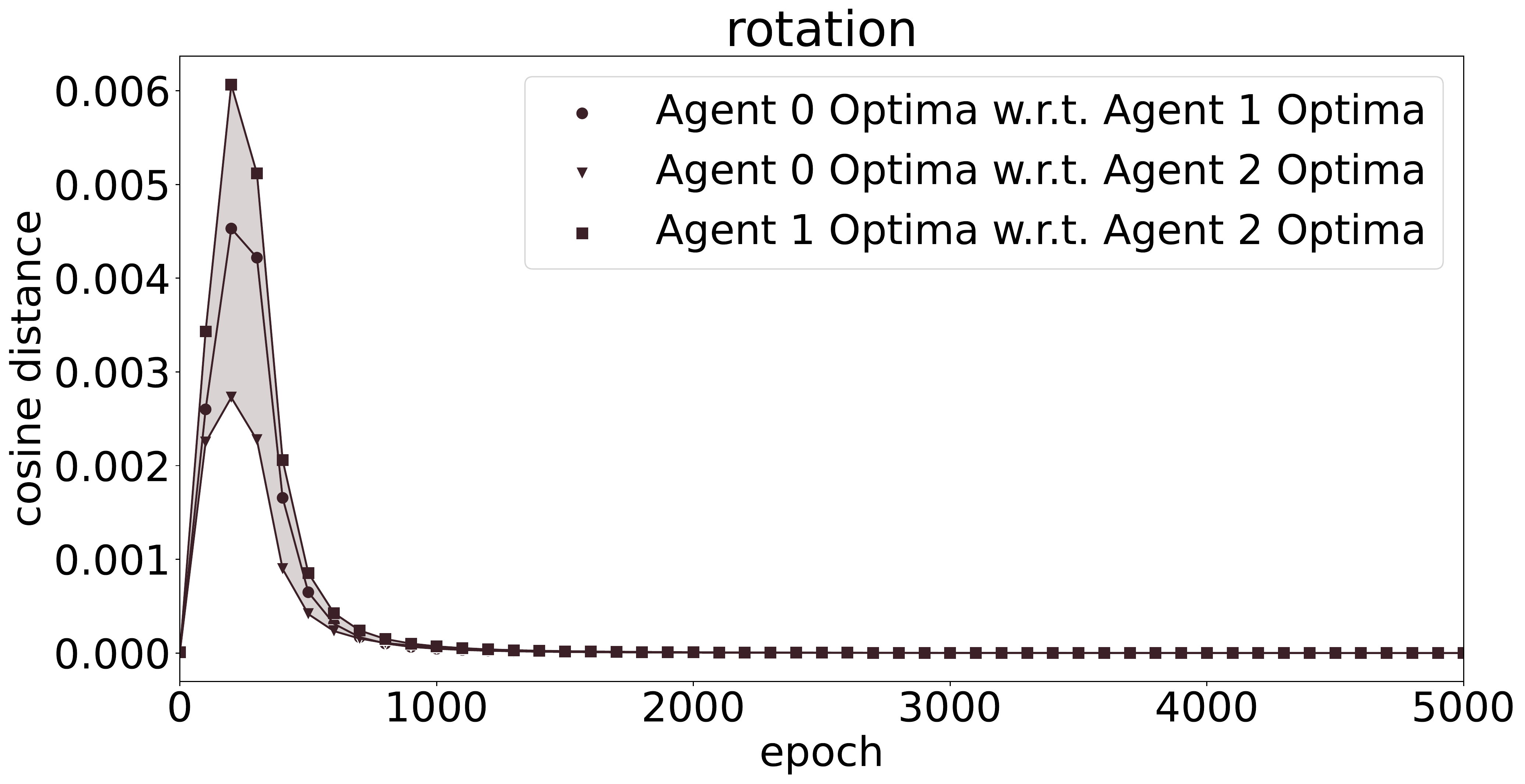}
    \includegraphics[width=0.245\textwidth]{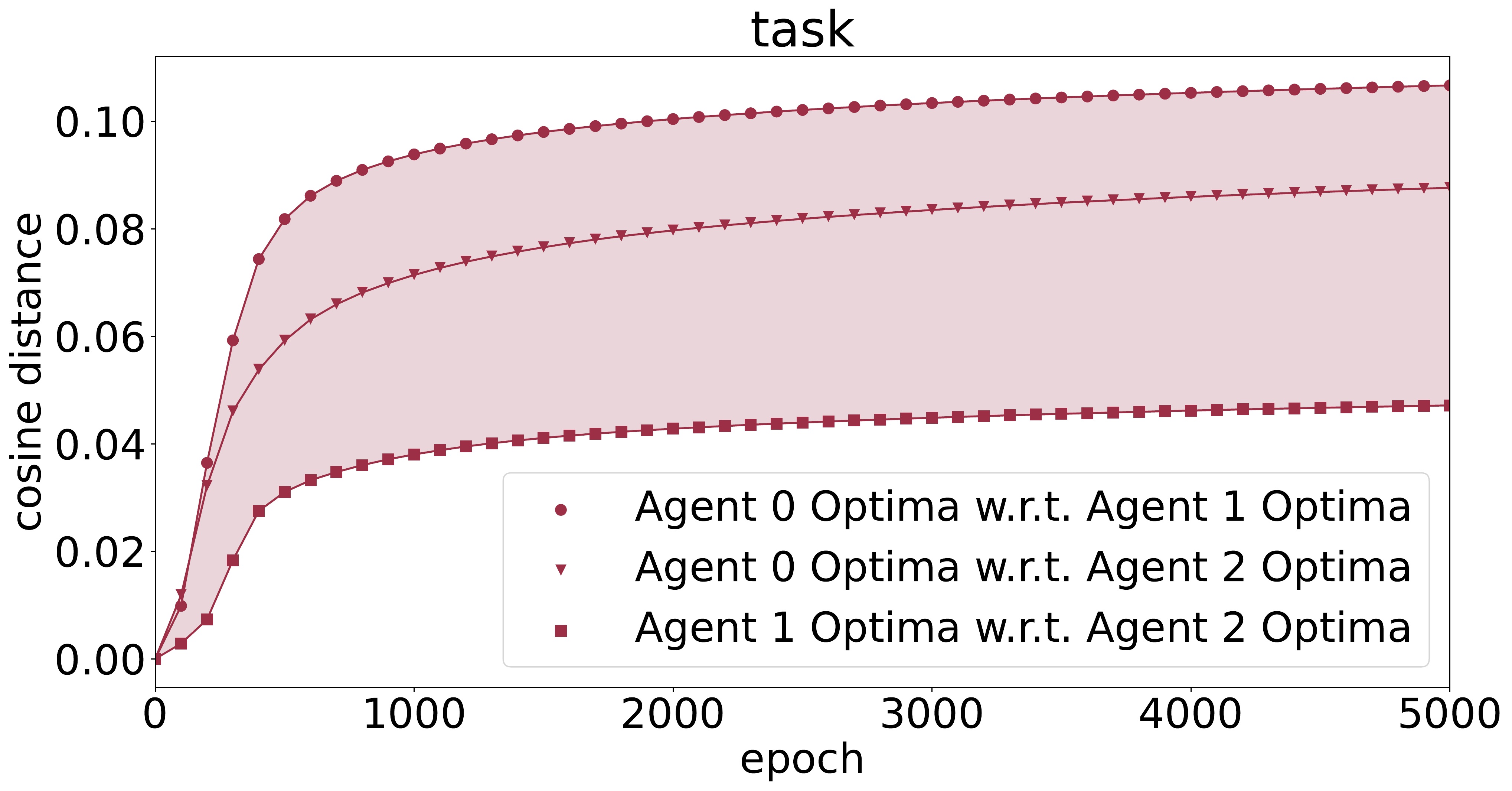}
    \includegraphics[width=0.245\textwidth]{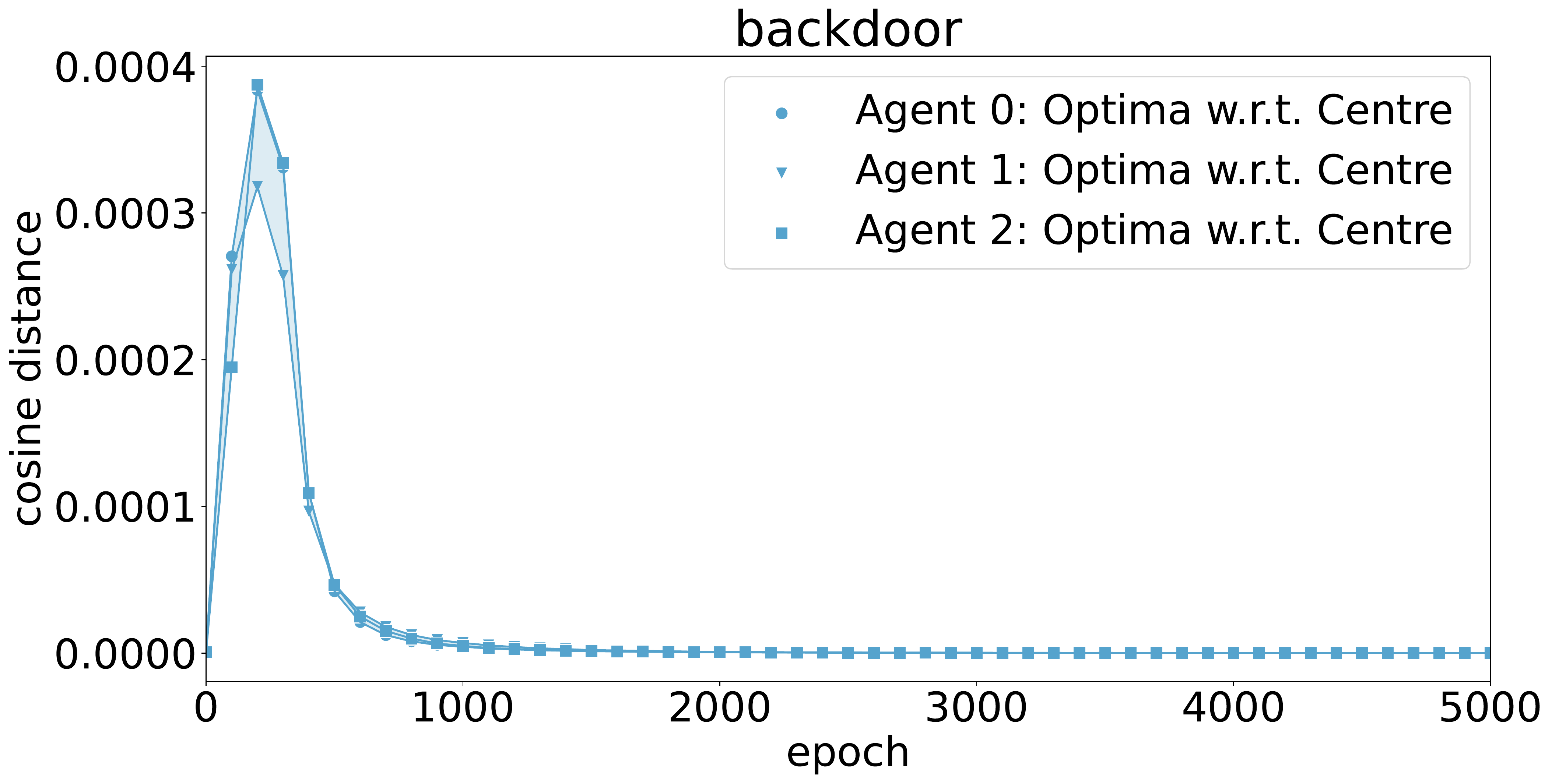}
    \includegraphics[width=0.245\textwidth]{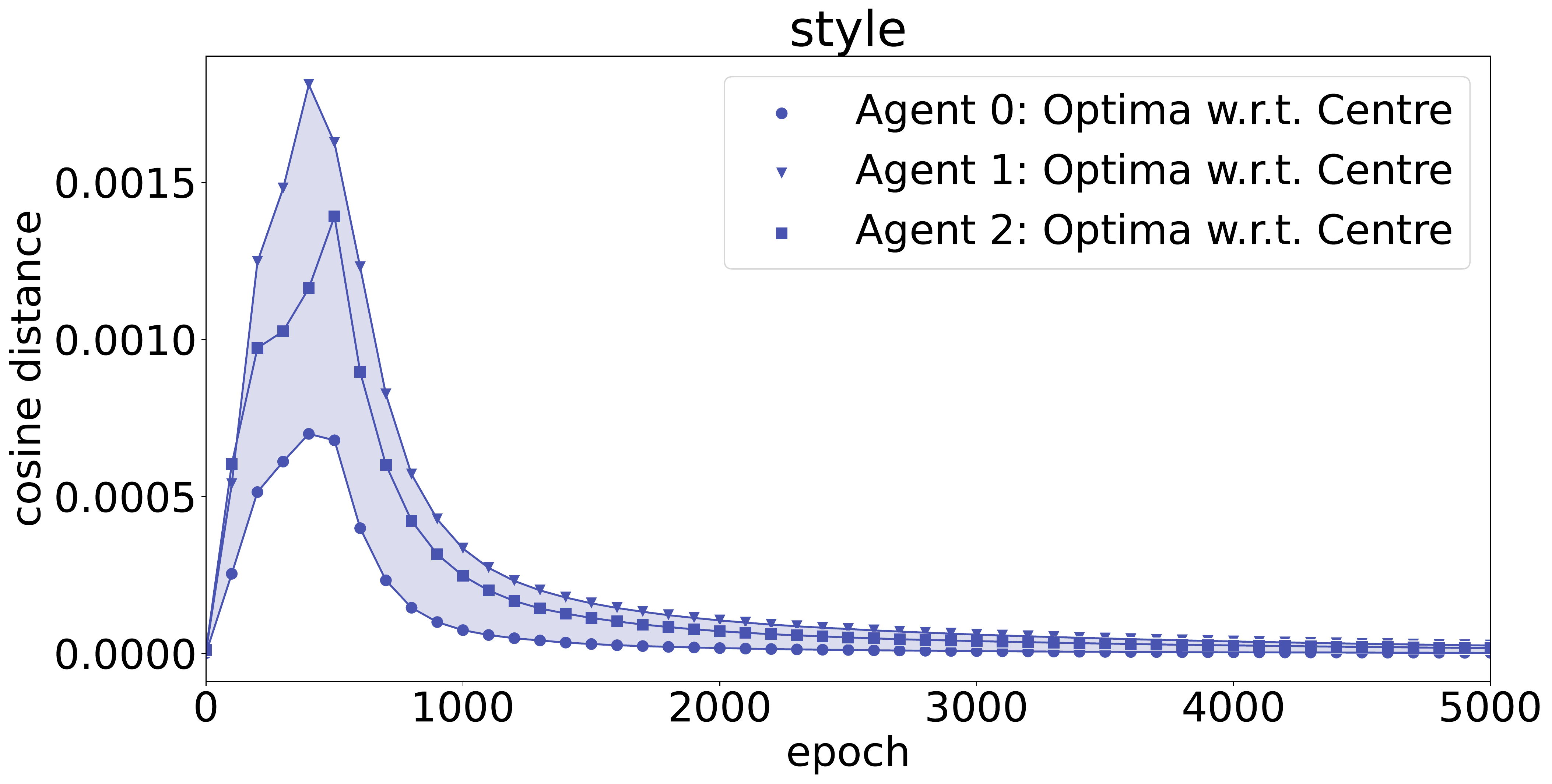}
    \includegraphics[width=0.245\textwidth]{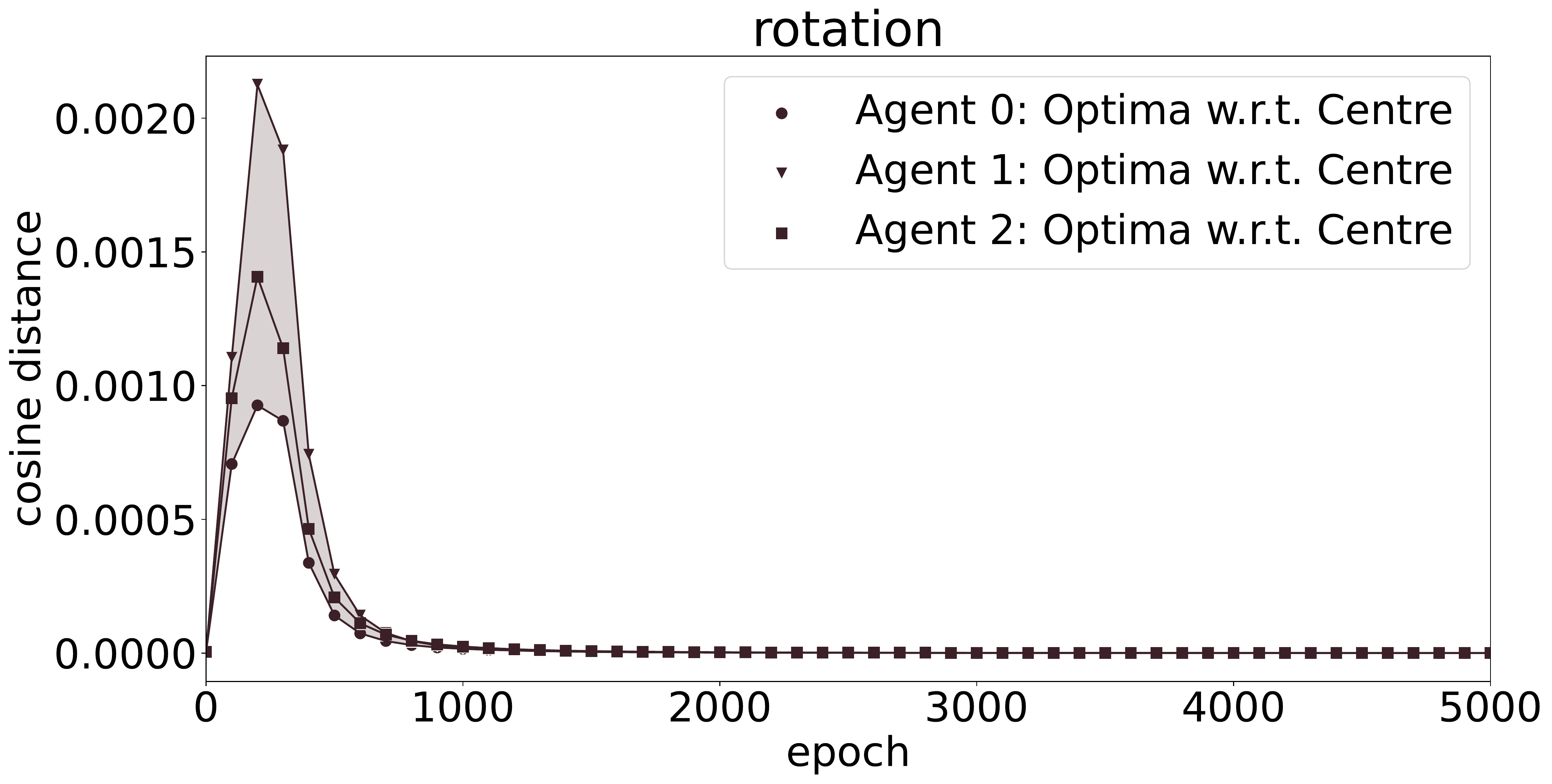}
    \includegraphics[width=0.245\textwidth]{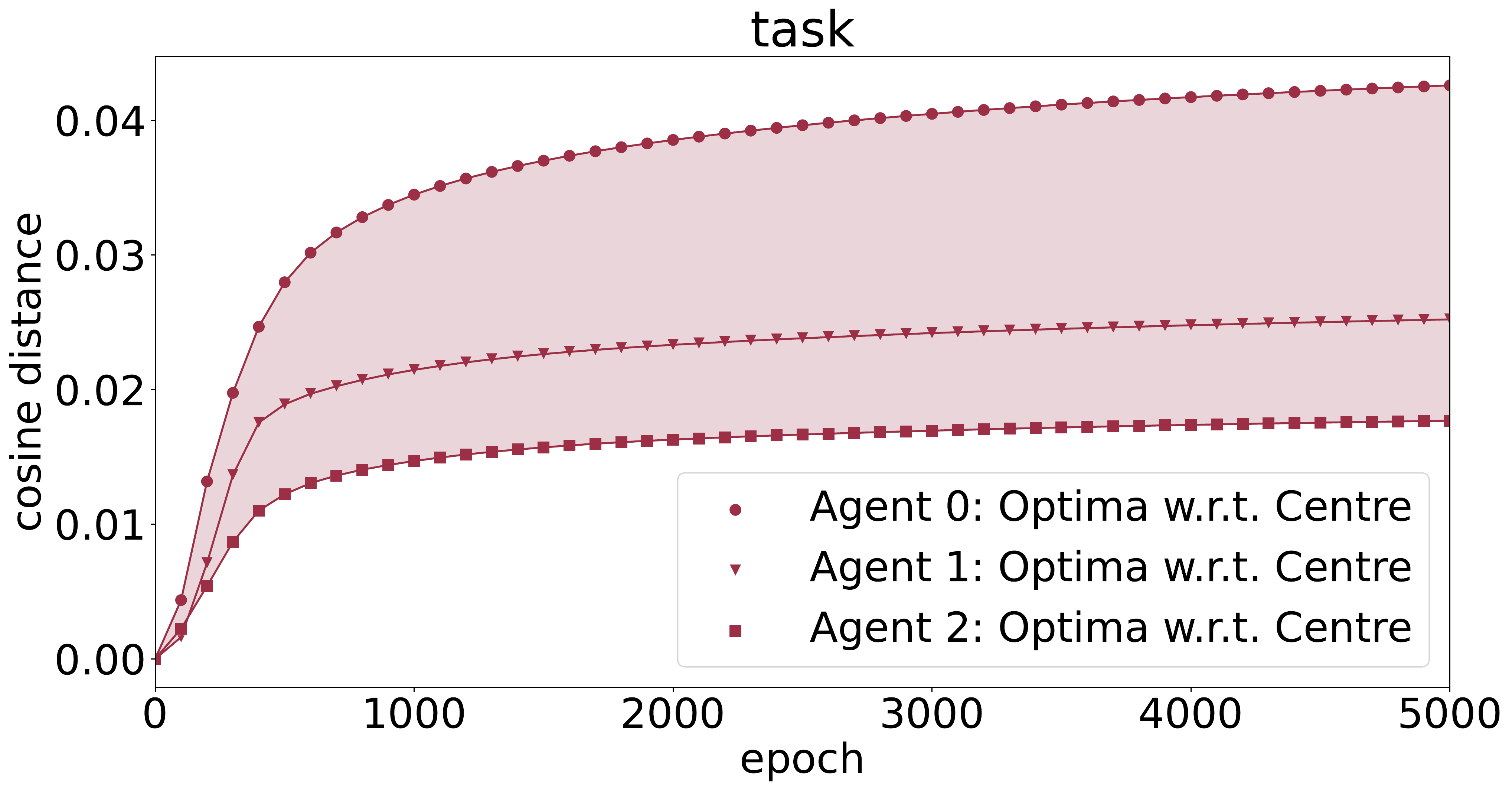}
    \caption{
    \textit{Change in parameter subspace dynamics: }
    Loss and cosine distance per epoch during training of CPS with 3 train-time distributions ($\beta=1.0$).
    Perturbation type of train-time subsets are
    backdoor (column 1), stylization (column 2), rotation (column 3), and task (column 4).
    }
    \label{fig:beta1_timeseries}
\end{figure*}

\begin{table*}[t]
\renewcommand{\fnum@table}{Appendix~A.3.\thetable}
\renewcommand{\fnum@figure}{Appendix~A.3.\thefigure}
\setcounter{figure}{3}
\setcounter{table}{3}
\centering
	\begin{minipage}{0.48\textwidth}
	\centering
    \resizebox{\textwidth}{!}{
        \begin{tabular}{|lrcccc}
        \multicolumn{6}{c}{3-layer CNN} \\ \hline
        \multicolumn{2}{|c}{\textit{Type / \# sets}} & 2 & 3 & 5 & \multicolumn{1}{c|}{10} \\ \hline
        \multicolumn{1}{|r}{\multirow{3}{*}{CPS (backdoor)}} 
        & centre & \multicolumn{1}{|c|}{$29.6 \pm 7.8$} & \multicolumn{1}{c|}{$55.3 \pm 15.8$} & \multicolumn{1}{c|}{$49.9 \pm 14.2$} & \multicolumn{1}{c|}{$49.9 \pm 13.6$} \\ 
        & ens. (mean) & \multicolumn{1}{|c|}{$33.7 \pm 8.7$} & \multicolumn{1}{c|}{$55.5 \pm 15.7$} & \multicolumn{1}{c|}{$49.7 \pm 14.4$} & \multicolumn{1}{c|}{$49.9 \pm 13.7$} \\
        & intp. (max) & \multicolumn{1}{|c|}{$55.1 \pm 14.9$} & \multicolumn{1}{c|}{$57.5 \pm 15.5$} & \multicolumn{1}{c|}{$52.7 \pm 15.6$} & \multicolumn{1}{c|}{$55.7 \pm 15.5$} \\
        \hline
        \multicolumn{1}{|r}{\multirow{3}{*}{CPS (stylization)}} 
        & centre & \multicolumn{1}{|c|}{$13.6 \pm 2.0$} & \multicolumn{1}{c|}{$57.5 \pm 17.8$} & \multicolumn{1}{c|}{$39.3 \pm 1.8$} & \multicolumn{1}{c|}{$34.4 \pm 4.9$} \\
        & ens. (mean) & \multicolumn{1}{|c|}{$20.9 \pm 3.8$} & \multicolumn{1}{c|}{$57.3 \pm 18.3$} & \multicolumn{1}{c|}{$39.8 \pm 2.0$} & \multicolumn{1}{c|}{$33.8 \pm 4.3$} \\
        & intp. (max) & \multicolumn{1}{|c|}{$64.0 \pm 20.2$} & \multicolumn{1}{c|}{$59.7 \pm 18.2$} & \multicolumn{1}{c|}{$52.5 \pm 2.8$} & \multicolumn{1}{c|}{$52.9 \pm 5.0$} \\ 
        \hline
        \multicolumn{1}{|r}{\multirow{3}{*}{CPS (rotation)}} 
        & centre & \multicolumn{1}{|c|}{$19.4 \pm 0.6$} & \multicolumn{1}{c|}{$19.0 \pm 1.8$} & \multicolumn{1}{c|}{$17.2 \pm 2.7$} & \multicolumn{1}{c|}{$9.9 \pm 3.0$} \\  
        & ens. (mean) & \multicolumn{1}{|c|}{$19.6 \pm 0.4$} & \multicolumn{1}{c|}{$19.2 \pm 1.8$} & \multicolumn{1}{c|}{$17.1 \pm 2.2$} & \multicolumn{1}{c|}{$15.7 \pm 2.7$} \\
        & intp. (max) & \multicolumn{1}{|c|}{$21.7 \pm 0.5$} & \multicolumn{1}{c|}{$59.6 \pm 17.2$} & \multicolumn{1}{c|}{$49.7 \pm 1.6$} & \multicolumn{1}{c|}{$50.5 \pm 5.6$} \\
        \hline
        \end{tabular}
        }
	\end{minipage}
	\begin{minipage}{0.463\textwidth}
	\centering
    \resizebox{\textwidth}{!}{
        \begin{tabular}{|lrcccc}
        \multicolumn{6}{c}{6-layer CNN} \\ \hline
        \multicolumn{2}{|c}{\textit{Type / \# sets}} & 2 & 3 & 5 & \multicolumn{1}{c|}{10} \\ \hline
        \multicolumn{1}{|r}{\multirow{3}{*}{CPS (backdoor)}} 
        & centre  & \multicolumn{1}{|c|}{$60.2 \pm 13.0$} & \multicolumn{1}{c|}{$48.4 \pm 2.4$} & \multicolumn{1}{c|}{$50.7 \pm 2.1$} & \multicolumn{1}{c|}{$52.7 \pm 4.7$} \\
        & ens. (mean) & \multicolumn{1}{|c|}{$59.3 \pm 12.8$} & \multicolumn{1}{c|}{$48.6 \pm 2.9$} & \multicolumn{1}{c|}{$49.3 \pm 1.7$} & \multicolumn{1}{c|}{$52.2 \pm 4.7$} \\ 
        & intp. (max) & \multicolumn{1}{|c|}{$61.9 \pm 12.9$} & \multicolumn{1}{c|}{$51.8 \pm 2.4$} & \multicolumn{1}{c|}{$52.8 \pm 1.7$} & \multicolumn{1}{c|}{$56.7 \pm 4.6$} \\
        \hline
        \multicolumn{1}{|r}{\multirow{3}{*}{CPS (stylization)}} 
        & centre & \multicolumn{1}{|c|}{$22.6 \pm 0.8$} & \multicolumn{1}{c|}{$47.9 \pm 1.1$} & \multicolumn{1}{c|}{$48.2 \pm 2.2$} & \multicolumn{1}{c|}{$52.6 \pm 3.9$} \\
        & ens. (mean) & \multicolumn{1}{|c|}{$29.6 \pm 1.1$} & \multicolumn{1}{c|}{$47.4 \pm 1.0$} & \multicolumn{1}{c|}{$47.7 \pm 2.5$} & \multicolumn{1}{c|}{$51.8 \pm 3.5$} \\
        & intp. (max)  & \multicolumn{1}{|c|}{$53.4 \pm 1.4$} & \multicolumn{1}{c|}{$50.0 \pm 0.8$} & \multicolumn{1}{c|}{$51.1 \pm 2.9$} & \multicolumn{1}{c|}{$55.9 \pm 3.5$} \\ 
        \hline
        \multicolumn{1}{|r}{\multirow{3}{*}{CPS (rotation)}} 
        & centre   & \multicolumn{1}{|c|}{$15.9 \pm 0.1$} & \multicolumn{1}{c|}{$19.4 \pm 1.2$} & \multicolumn{1}{c|}{$21.2 \pm 3.1$} & \multicolumn{1}{c|}{$24.8 \pm 3.4$} \\
        & ens. (mean) & \multicolumn{1}{|c|}{$17.3 \pm 0.5$} & \multicolumn{1}{c|}{$19.8 \pm 1.2$} & \multicolumn{1}{c|}{$20.8 \pm 2.3$} & \multicolumn{1}{c|}{$24.2 \pm 3.0$} \\
        & intp. (max) & \multicolumn{1}{|c|}{$48.1 \pm 2.5$} & \multicolumn{1}{c|}{$51.4 \pm 1.5$} & \multicolumn{1}{c|}{$49.5 \pm 4.0$} & \multicolumn{1}{c|}{$51.3 \pm 3.7$} \\
        \hline
        \end{tabular}
        }
	\end{minipage}
	\caption{
	\textit{CPS vs single test-time distributions: }
	The relationship between the type of trained perturbations (backdoor / stylization / rotation), number of perturbed sets, and model capacity.
	}
	\label{tab:source_hypertuning}
\end{table*}

\begin{table*}[t]
\renewcommand{\fnum@table}{Appendix~A.3.\thetable}
\renewcommand{\fnum@figure}{Appendix~A.3.\thefigure}
\setcounter{figure}{4}
\setcounter{table}{4}
\centering
	\begin{minipage}[t]{0.48\textwidth}
	\begin{minipage}[t]{\textwidth}
	\centering
    \resizebox{\textwidth}{!}{
        \begin{tabular}{|lr|ccc|}
        \hline
        && \multicolumn{3}{c|}{layers $\ell$} \\
        && 3 & 6 & 9 \\ \hline
        \multirow{3}{*}{
        \thead{2 tasks (2 $\times$ 5-label-set per task, \\ 2 $\times$ same coarse label per task, \\ 0 $\times$ distinctly different coarse label per task)}}
        & centre & $48.0 \pm 12.1$ & $20.0 \pm 1.2$ & $30.1 \pm 3.2$ \\ 
        & ens. (mean) & $68.0 \pm 32.0$ & $24.0 \pm 2.4$ & $35.4 \pm 0.0$ \\ 
        & intp. (max) & $86.6 \pm 9.1$ & $65.7 \pm 34.2$ & $91.0 \pm 9.0$ \\
        \hline
        \multirow{3}{*}{
        \thead{3 tasks (2 $\times$ 5-label-set per task, \\ 2 $\times$ same coarse label per task, \\ 1 $\times$ distinctly different coarse label per task)}}
        & centre & $29.5 \pm 4.8$ & $20.6 \pm 2.0$ & $20.6 \pm 1.1$ \\ 
        & ens. (mean) & $29.3 \pm 5.3$ & $23.6 \pm 1.7$ & $19.9 \pm 0.8$ \\ 
        & intp. (max) & $83.4 \pm 12.0$ & $74.1 \pm 36.3$ & $53.8 \pm 32.7$\\
        \hline
        \multirow{3}{*}{
        \thead{4 tasks (2 $\times$ 5-label-set per task, \\ 2 $\times$ same coarse label per task, \\ 2 $\times$ distinctly different coarse label per task)}}
        & centre & $17.7 \pm 1.9$ & $19.2 \pm 2.1$ & $20.1 \pm 0.6$ \\ 
        & ens. (mean) & $19.6 \pm 0.6$ & $19.4 \pm 0.8$ & $21.8 \pm 1.5$\\ 
        & intp. (max) & $38.5 \pm 27.5$ & $64.0 \pm 36.2$ & $49.3 \pm 26.7$\\
        \hline
        \end{tabular}
        }
    \label{tab:task_hypertuning_1}
    \caption{
    \textit{CPS vs multiple test-time distributions: }
    Varying model depth against distinct coarse label sets in train-time distributions.
    }
	\end{minipage}
	\begin{minipage}[t]{\textwidth}
	\centering
	\tiny
    \resizebox{\textwidth}{!}{
        \begin{tabular}{|crr|c|c|}
        \hline
        \multicolumn{3}{|c}{} & \multicolumn{2}{c|}{6-layer CNN}\\ 
        \multicolumn{3}{|c}{} & \multicolumn{1}{c}{narrow}  & \multicolumn{1}{c|}{wide} \\\cline{4-5}
        \multirow{6}{*}{\rotatebox[origin=c]{90}{2 tasks}} 
        & \multirow{3}{*}{same} 
        & centre & $17.5 \pm 2.1$ & $37.9 \pm 4.6$ \\ 
        & 
        & ens. (mean) & $24.2 \pm 5.6$ & $43.2 \pm 2.2$ \\ 
        &
        & intp. (max) & $62.1 \pm 37.9$ & $82.7 \pm 3.5$ \\ \cline{4-5}
        & \multirow{3}{*}{different} 
        & centre & $23.5 \pm 6.9$ & $21.7 \pm 0.8$ \\ 
        &
        & ens. (mean) & $30.9 \pm 10.8$ & $31.0 \pm 11.4$ \\ 
        &
        & intp. (max) & $62.9 \pm 37.1$ & $60.8 \pm 39.2$ \\ 
        \cline{1-5}
        \end{tabular}
        }
    \label{tab:task_hypertuning_2}
    \caption{
    \textit{CPS vs multiple test-time distributions: }
    $N=2$ tasks of either \textit{same} or \textit{different} coarse labels, against model width.
    }
	\end{minipage}
	\end{minipage}
	\hfill
    \begin{minipage}[t]{0.48\textwidth}
    \vspace{0.015cm}
	\centering
    \resizebox{\textwidth}{!}{
        \begin{tabular}{|lr|ccc|}
        \hline
        && \multicolumn{3}{c|}{layers $\ell$} \\
        && 3 & 6 & 9 \\ \hline
        \multirow{3}{*}{
        \thead{3 tasks (1 $\times$ 5-label-set per task, \\ 3 $\times$ same coarse label)}}
        & centre & $21.7 \pm 2.2$ & $31.0 \pm 8.4$ & $25.1 \pm 2.2$ \\ 
        & ens. (mean) & $23.3 \pm 4.6$ & $31.0 \pm 4.9$ & $26.8 \pm 1.3$ \\ 
        & intp. (max) & $89.7 \pm 7.0$ & $88.5 \pm 4.5$ & $93.9 \pm 7.8$ \\
        \hline
        \multirow{3}{*}{
        \thead{3 tasks (1 $\times$ 5-label-set per task, \\ 2 $\times$ same coarse label, \\ 1 $\times$ different coarse label)}}
        & centre & $20.7 \pm 5.2$ & $24.2 \pm 1.4$ & $22.5 \pm 2.7$ \\ 
        & ens. (mean) & $24.3 \pm 9.3$ & $26.9 \pm 3.0$ & $21.2 \pm 1.1$ \\ 
        & intp. (max) & $87.4 \pm 3.1$ & $83.9 \pm 4.5$ & $57.8 \pm 31.4$ \\
        \hline
        \multirow{3}{*}{
        \thead{3 tasks (1 $\times$ 5-label-set per task, \\ 3 $\times$ distinctly different coarse label)}}
        & centre & $19.0 \pm 6.5$ & $21.7 \pm 1.2$ & $21.9 \pm 3.8$ \\ 
        & ens. (mean) & $21.2 \pm 6.9$ & $25.0 \pm 3.5$ & $22.4 \pm 1.2$ \\ 
        & intp. (max) & $91.4 \pm 8.3$ & $89.0 \pm 5.6$ & $74.0 \pm 36.7$ \\
        \hline
        \multirow{3}{*}{
        \thead{5 tasks (1 $\times$ 5-label-set per task, \\ 5 $\times$ same coarse label)}}
        & centre & $20.2 \pm 1.3$ & $23.2 \pm 6.6$ & $20.2 \pm 1.3$ \\ 
        & ens. (mean) & $20.2 \pm 1.3$ & $26.3 \pm 3.6$ & $21.2 \pm 1.2$ \\ 
        & intp. (max) & $21.8 \pm 0.7$ & $87.7 \pm 24.7$ & $61.1 \pm 32.0$ \\
        \hline
        \multirow{3}{*}{
        \thead{5 tasks (1 $\times$ 5-label-set per task, \\ 2 $\times$ same coarse label, \\ 3 $\times$ distinctly different coarse label)}}
        & centre & $20.2 \pm 1.3$ & $19.9 \pm 1.5$ & $23.5 \pm 5.4$ \\ 
        & ens. (mean) & $20.2 \pm 1.3$ & $21.4 \pm 1.5$ & $21.6 \pm 2.4$ \\ 
        & intp. (max) & $21.6 \pm 0.5$ & $84.8 \pm 30.3$ & $55.8 \pm 36.3$ \\
        \hline
        \multirow{3}{*}{
        \thead{5 tasks (1 $\times$ 5-label-set per task, \\ 2 $\times$ same coarse label, \\ 3 $\times$ different coarse label of same set)}}
        & centre & $20.2 \pm 1.3$ & $23.7 \pm 3.5$ & $20.0 \pm 1.9$ \\ 
        & ens. (mean) & $20.2 \pm 1.3$ & $23.2 \pm 1.9$ & $21.0 \pm 2.1$ \\ 
        & intp. (max) & $21.6 \pm 0.5$ & $73.7 \pm 32.3$ & $60.6 \pm 33.0$ \\
        \hline
        \end{tabular}
        }
    \label{tab:task_hypertuning_3}
    \caption{
    \textit{CPS vs multiple test-time distributions: }
    Varying model depth against varying task label set diversity (different fine labels, different coarse labels).
    }
	\end{minipage}
\end{table*}

\begin{table*}[t]
\renewcommand{\fnum@table}{Appendix~A.3.\thetable}
\renewcommand{\fnum@figure}{Appendix~A.3.\thefigure}
\setcounter{figure}{7}
\setcounter{table}{7}
\centering
\parbox{\textwidth}{
\centering
\resizebox{\textwidth}{!}{
\begin{tabular}{|lr|p{4cm}p{4cm}p{4cm}p{4cm}|}
\hline
&& CPS: 3 tasks, 3 $\times$ same coarse label, 6-layer CNN, unique-task solution 
& CPS: 3 tasks, 3 $\times$ same coarse label, 6-layer CNN, multi-task solution
& CPS: 2 tasks, 2 $\times$ same coarse label, 6-layer-wide CNN, unique-task solution 
& CPS: 2 tasks, 2 $\times$ same coarse label, 6-layer-wide CNN, multi-task solution
\\ \hline \hline
\multicolumn{6}{|c|}{-} \\ \hline 
\multirow{3}{*}{
\thead{
Seen tasks (3 tasks, 3 $\times$ same coarse label, \\ train-time task set)
}} 
& centre & \multicolumn{1}{c}{$33.9 \pm 5.7$} & \multicolumn{1}{c}{$27.2 \pm 0.0$} & \multicolumn{1}{c}{$38.9 \pm 4.8$} & \multicolumn{1}{c|}{$37.2 \pm 0.0$} \\ 
& ens. (mean) & \multicolumn{1}{c}{$31.0 \pm 4.8$} & \multicolumn{1}{c}{$28.6 \pm 0.0$} & \multicolumn{1}{c}{$36.3 \pm 3.2$} & \multicolumn{1}{c|}{$35.9 \pm 0.0$} \\
& intp. (max) & \multicolumn{1}{c}{$59.4 \pm 9.1$} & \multicolumn{1}{c}{$43.3 \pm 0.0$} & \multicolumn{1}{c}{$57.2 \pm 8.6$} & \multicolumn{1}{c|}{$40.6 \pm 0.0$} \\
\hline
\multirow{3}{*}{
\thead{
Unseen tasks (2 tasks, 2 $\times$ same coarse label \\ but unseen fine label)
}} 
& centre & \multicolumn{1}{c}{$34.8 \pm 2.7$} & \multicolumn{1}{c}{$33.8 \pm 0.0$} & \multicolumn{1}{c}{$31.0 \pm 0.2$} & \multicolumn{1}{c|}{$30.0 \pm 0.0$} \\ 
& ens. (mean) & \multicolumn{1}{c}{$29.4 \pm 0.2$} & \multicolumn{1}{c}{$29.3 \pm 0.0$} & \multicolumn{1}{c}{$32.2 \pm 3.8$} & \multicolumn{1}{c|}{$31.3 \pm 0.0$} \\
& intp. (max) & \multicolumn{1}{c}{$42.9 \pm 5.8$} & \multicolumn{1}{c}{$41.7 \pm 0.0$} & \multicolumn{1}{c}{$40.4 \pm 5.4$} & \multicolumn{1}{c|}{$35.2 \pm 0.0$} \\
\hline
\multirow{3}{*}{
\thead{
Unseen tasks (5 tasks, 1 $\times$ 5-label-set per task, \\ 3 $\times$ same coarse label but unseen fine label, \\ 2 $\times$ distinctly different coarse label)
}} 
& centre & \multicolumn{1}{c}{$18.4 \pm 3.5$} & \multicolumn{1}{c}{$18.0 \pm 0.0$} & \multicolumn{1}{c}{$23.3 \pm 4.3$} & \multicolumn{1}{c|}{$23.0 \pm 0.0$} \\ 
& ens. (mean) & \multicolumn{1}{c}{$18.8 \pm 2.4$} & \multicolumn{1}{c}{$18.9 \pm 0.0$} & \multicolumn{1}{c}{$23.1 \pm 3.1$} & \multicolumn{1}{c|}{$22.2 \pm 0.0$} \\
& intp. (max) & \multicolumn{1}{c}{$30.9 \pm 1.1$} & \multicolumn{1}{c}{$28.0 \pm 0.0$} & \multicolumn{1}{c}{$30.3 \pm 2.2$} & \multicolumn{1}{c|}{$32.0 \pm 0.0$} \\
\hline \hline 
\multicolumn{6}{|c|}{+ Adversarial Attack} \\ \hline
\multirow{3}{*}{
\thead{
Seen tasks (3 tasks, 3 $\times$ same coarse label, \\ train-time task set)
}} 
& centre & \multicolumn{1}{c}{$29.4 \pm 6.1$} & \multicolumn{1}{c}{$25.0 \pm 0.0$} & \multicolumn{1}{c}{$36.7 \pm 3.6$} & \multicolumn{1}{c|}{$31.7 \pm 0.0$} \\ 
& ens. (mean) & \multicolumn{1}{c}{$30.0 \pm 4.7$} & \multicolumn{1}{c}{$27.9 \pm 0.0$} & \multicolumn{1}{c}{$36.5 \pm 4.1$} & \multicolumn{1}{c|}{$32.2 \pm 0.0$} \\
& intp. (max) & \multicolumn{1}{c}{$55.6 \pm 5.2$} & \multicolumn{1}{c}{$42.8 \pm 0.0$} & \multicolumn{1}{c}{$52.8 \pm 5.7$} & \multicolumn{1}{c|}{$37.8 \pm 0.0$} \\
\hline
\multirow{3}{*}{
\thead{
Unseen tasks (2 tasks, 2 $\times$ same coarse label \\ but unseen fine label)
}} 
& centre & \multicolumn{1}{c}{$33.3 \pm 11.7$} & \multicolumn{1}{c}{$32.5 \pm 0.0$} & \multicolumn{1}{c}{$31.7 \pm 5.0$} & \multicolumn{1}{c|}{$38.3 \pm 0.0$} \\ 
& ens. (mean) & \multicolumn{1}{c}{$30.9 \pm 6.1$} & \multicolumn{1}{c}{$27.1 \pm 0.0$} & \multicolumn{1}{c}{$31.8 \pm 0.5$} & \multicolumn{1}{c|}{$34.4 \pm 0.0$} \\
& intp. (max) & \multicolumn{1}{c}{$42.5 \pm 2.5$} & \multicolumn{1}{c}{$37.5 \pm 0.0$} & \multicolumn{1}{c}{$38.3 \pm 3.3$} & \multicolumn{1}{c|}{$42.5 \pm 0.0$} \\
\hline
\multirow{3}{*}{
\thead{
Unseen tasks (5 tasks, 1 $\times$ 5-label-set per task, \\ 3 $\times$ same coarse label but unseen fine label, \\ 2 $\times$ distinctly different coarse label)
}} 
& centre & \multicolumn{1}{c}{$15.0 \pm 3.8$} & \multicolumn{1}{c}{$19.0 \pm 0.0$} & \multicolumn{1}{c}{$21.0 \pm 4.7$} & \multicolumn{1}{c|}{$20.0 \pm 0.0$} \\ 
& ens. (mean) & \multicolumn{1}{c}{$18.6 \pm 3.1$} & \multicolumn{1}{c}{$18.9 \pm 0.0$} & \multicolumn{1}{c}{$23.9 \pm 3.1$} & \multicolumn{1}{c|}{$22.3 \pm 0.0$} \\
& intp. (max) & \multicolumn{1}{c}{$25.3 \pm 2.9$} & \multicolumn{1}{c}{$22.3 \pm 0.0$} & \multicolumn{1}{c}{$33.0 \pm 6.9$} & \multicolumn{1}{c|}{$29.0 \pm 0.0$} \\
\hline \hline 
\multicolumn{6}{|c|}{+ Random Permutations} \\ \hline
\multirow{3}{*}{
\thead{
Seen tasks (3 tasks, 3 $\times$ same coarse label, \\ train-time task set)
}} 
& centre & \multicolumn{1}{c}{$33.3 \pm 4.9$} & \multicolumn{1}{c}{$27.8 \pm 0.0$} & \multicolumn{1}{c}{$39.4 \pm 5.5$} & \multicolumn{1}{c|}{$37.8 \pm 0.0$} \\ 
& ens. (mean) & \multicolumn{1}{c}{$31.6 \pm 4.6$} & \multicolumn{1}{c}{$28.7 \pm 0.0$} & \multicolumn{1}{c}{$37.2 \pm 3.8$} & \multicolumn{1}{c|}{$35.6 \pm 0.0$} \\
& intp. (max) & \multicolumn{1}{c}{$60.8 \pm 21.7$} & \multicolumn{1}{c}{$43.3 \pm 0.0$} & \multicolumn{1}{c}{$58.9 \pm 7.5$} & \multicolumn{1}{c|}{$40.6 \pm 0.0$} \\
\hline
\multirow{3}{*}{
\thead{
Unseen tasks (2 tasks, 2 $\times$ same coarse label \\ but unseen fine label)
}} 
& centre & \multicolumn{1}{c}{$34.8 \pm 2.7$} & \multicolumn{1}{c}{$33.1 \pm 0.0$} & \multicolumn{1}{c}{$31.3 \pm 0.0$} & \multicolumn{1}{c|}{$29.6 \pm 0.0$} \\ 
& ens. (mean) & \multicolumn{1}{c}{$29.4 \pm 0.0$} & \multicolumn{1}{c}{$28.8 \pm 0.0$} & \multicolumn{1}{c}{$32.7 \pm 3.1$} & \multicolumn{1}{c|}{$31.1 \pm 0.0$} \\
& intp. (max) & \multicolumn{1}{c}{$42.7 \pm 6.0$} & \multicolumn{1}{c}{$41.9 \pm 0.0$} & \multicolumn{1}{c}{$41.5 \pm 4.4$} & \multicolumn{1}{c|}{$35.6 \pm 0.0$} \\
\hline
\multirow{3}{*}{
\thead{
Unseen tasks (5 tasks, 1 $\times$ 5-label-set per task, \\ 3 $\times$ same coarse label but unseen fine label, \\ 2 $\times$ distinctly different coarse label)
}} 
& centre & \multicolumn{1}{c}{$18.5 \pm 3.7$} & \multicolumn{1}{c}{$18.3 \pm 0.0$} & \multicolumn{1}{c}{$22.9 \pm 4.4$} & \multicolumn{1}{c|}{$21.7 \pm 0.0$} \\ 
& ens. (mean) & \multicolumn{1}{c}{$18.5 \pm 2.2$} & \multicolumn{1}{c}{$19.0 \pm 0.0$} & \multicolumn{1}{c}{$23.1 \pm 2.7$} & \multicolumn{1}{c|}{$22.7 \pm 0.0$} \\
& intp. (max) & \multicolumn{1}{c}{$30.8 \pm 1.1$} & \multicolumn{1}{c}{$28.0 \pm 0.0$} & \multicolumn{1}{c}{$30.3 \pm 2.1$} & \multicolumn{1}{c|}{$29.3 \pm 0.0$} \\
\hline \hline 
\multicolumn{6}{|c|}{+ Stylization} \\ \hline
\multirow{3}{*}{
\thead{
Seen tasks (3 tasks, 3 $\times$ same coarse label, \\ train-time task set)
}} 
& centre & \multicolumn{1}{c}{$28.3 \pm 11.6$} & \multicolumn{1}{c}{$25.0 \pm 0.0$} & \multicolumn{1}{c}{$27.2 \pm 0.8$} & \multicolumn{1}{c|}{$31.7 \pm 0.0$} \\ 
& ens. (mean) & \multicolumn{1}{c}{$27.0 \pm 6.1$} & \multicolumn{1}{c}{$25.8 \pm 0.0$} & \multicolumn{1}{c}{$27.5 \pm 2.0$} & \multicolumn{1}{c|}{$30.9 \pm 0.0$} \\
& intp. (max) & \multicolumn{1}{c}{$48.3 \pm 5.9$} & \multicolumn{1}{c}{$36.7 \pm 0.0$} & \multicolumn{1}{c}{$36.7 \pm 6.2$} & \multicolumn{1}{c|}{$35.6 \pm 0.0$} \\
\hline
\multirow{3}{*}{
\thead{
Unseen tasks (2 tasks, 2 $\times$ same coarse label \\ but unseen fine label)
}} 
& centre & \multicolumn{1}{c}{$33.1 \pm 1.5$} & \multicolumn{1}{c}{$38.5 \pm 0.0$} & \multicolumn{1}{c}{$32.1 \pm 4.2$} & \multicolumn{1}{c|}{$29.0 \pm 0.0$} \\ 
& ens. (mean) & \multicolumn{1}{c}{$29.5 \pm 0.2$} & \multicolumn{1}{c}{$30.1 \pm 0.0$} & \multicolumn{1}{c}{$30.3 \pm 2.0$} & \multicolumn{1}{c|}{$30.3 \pm 0.0$} \\
& intp. (max) & \multicolumn{1}{c}{$42.3 \pm 0.6$} & \multicolumn{1}{c}{$41.5 \pm 0.0$} & \multicolumn{1}{c}{$36.3 \pm 3.3$} & \multicolumn{1}{c|}{$35.6 \pm 0.0$} \\
\hline
\multirow{3}{*}{
\thead{
Unseen tasks (5 tasks, 1 $\times$ 5-label-set per task, \\ 3 $\times$ same coarse label but unseen fine label, \\ 2 $\times$ distinctly different coarse label)
}} 
& centre & \multicolumn{1}{c}{$19.4 \pm 3.9$} & \multicolumn{1}{c}{$16.0 \pm 0.0$} & \multicolumn{1}{c}{$23.9 \pm 4.6$} & \multicolumn{1}{c|}{$19.7 \pm 0.0$} \\ 
& ens. (mean) & \multicolumn{1}{c}{$20.5 \pm 1.5$} & \multicolumn{1}{c}{$18.2 \pm 0.0$} & \multicolumn{1}{c}{$25.3 \pm 3.2$} & \multicolumn{1}{c|}{$22.2 \pm 0.0$} \\
& intp. (max) & \multicolumn{1}{c}{$32.8 \pm 3.3$} & \multicolumn{1}{c}{$31.0 \pm 0.0$} & \multicolumn{1}{c}{$32.8 \pm 2.5$} & \multicolumn{1}{c|}{$33.3 \pm 0.0$} \\
\hline \hline 
\multicolumn{6}{|c|}{+ Rotation} \\ \hline 
\multirow{3}{*}{
\thead{
Seen tasks (3 tasks, 3 $\times$ same coarse label, \\ train-time task set)
}} 
& centre & \multicolumn{1}{c}{$33.3 \pm 6.2$} & \multicolumn{1}{c}{$26.1 \pm 0.0$} & \multicolumn{1}{c}{$32.2 \pm 5.5$} & \multicolumn{1}{c|}{$32.8 \pm 0.0$} \\ 
& ens. (mean) & \multicolumn{1}{c}{$30.7 \pm 5.1$} & \multicolumn{1}{c}{$27.3 \pm 0.0$} & \multicolumn{1}{c}{$32.8 \pm 3.9$} & \multicolumn{1}{c|}{$32.9 \pm 0.0$} \\
& intp. (max) & \multicolumn{1}{c}{$54.4 \pm 3.4$} & \multicolumn{1}{c}{$37.2 \pm 0.0$} & \multicolumn{1}{c}{$52.8 \pm 7.5$} & \multicolumn{1}{c|}{$35.6 \pm 0.0$} \\
\hline
\multirow{3}{*}{
\thead{
Unseen tasks (2 tasks, 2 $\times$ same coarse label \\ but unseen fine label)
}} 
& centre & \multicolumn{1}{c}{$33.5 \pm 4.0$} & \multicolumn{1}{c}{$34.2 \pm 0.0$} & \multicolumn{1}{c}{$30.4 \pm 0.8$} & \multicolumn{1}{c|}{$35.0 \pm 0.0$} \\ 
& ens. (mean) & \multicolumn{1}{c}{$28.5 \pm 1.0$} & \multicolumn{1}{c}{$29.1 \pm 0.0$} & \multicolumn{1}{c}{$31.5 \pm 4.9$} & \multicolumn{1}{c|}{$34.8 \pm 0.0$} \\
& intp. (max) & \multicolumn{1}{c}{$41.3 \pm 7.5$} & \multicolumn{1}{c}{$40.8 \pm 0.0$} & \multicolumn{1}{c}{$39.6 \pm 6.3$} & \multicolumn{1}{c|}{$37.9 \pm 0.0$} \\
\hline
\multirow{3}{*}{
\thead{
Unseen tasks (5 tasks, 1 $\times$ 5-label-set per task, \\ 3 $\times$ same coarse label but unseen fine label, \\ 2 $\times$ distinctly different coarse label)
}} 
& centre & \multicolumn{1}{c}{$17.8 \pm 3.2$} & \multicolumn{1}{c}{$19.7 \pm 0.0$} & \multicolumn{1}{c}{$21.7 \pm 3.1$} & \multicolumn{1}{c|}{$18.7 \pm 0.0$} \\ 
& ens. (mean) & \multicolumn{1}{c}{$18.4 \pm 2.6$} & \multicolumn{1}{c}{$17.9 \pm 0.0$} & \multicolumn{1}{c}{$21.5 \pm 1.1$} & \multicolumn{1}{c|}{$19.6 \pm 0.0$} \\
& intp. (max) & \multicolumn{1}{c}{$31.4 \pm 1.5$} & \multicolumn{1}{c}{$25.3 \pm 0.0$} & \multicolumn{1}{c}{$28.6 \pm 3.3$} & \multicolumn{1}{c|}{$30.3 \pm 0.0$} \\
\hline
\end{tabular}
}
}
\caption{
\textit{CPS vs multiple test-time distributions: }
Evaluating seen/unseen tasks (varying perturbation types)
with CPS of varying over-parameterization (number of tasks and model width).
}
\label{tab:task_baseline}
\end{table*}

\end{document}